# Reasoning over Ontologies with Hidden Content: The Import-by-Query Approach

**Bernardo Cuenca Grau**                                    BERNARDO.CUENCA.GRAU@CS.OX.AC.UK
**Boris Motik**                                                       BORIS.MOTIK@CS.OX.AC.UK
*Department of Computer Science, Oxford University*
*Wolfson Building, Parks Road, Oxford OX1 3QD UK*

## Abstract

There is currently a growing interest in techniques for hiding parts of the signature of an ontology $\mathcal{K}_h$ that is being reused by another ontology $\mathcal{K}_v$. Towards this goal, in this paper we propose the *import-by-query framework*, which makes the content of $\mathcal{K}_h$ accessible through a limited query interface. If $\mathcal{K}_v$ reuses the symbols from $\mathcal{K}_h$ in a certain restricted way, one can reason over $\mathcal{K}_v \cup \mathcal{K}_h$ by accessing only $\mathcal{K}_v$ and the query interface. We map out the landscape of the import-by-query problem. In particular, we outline the limitations of our framework and prove that certain restrictions on the expressivity of $\mathcal{K}_h$ and the way in which $\mathcal{K}_v$ reuses symbols from $\mathcal{K}_h$ are strictly necessary to enable reasoning in our setting. We also identify cases in which reasoning is possible and we present suitable import-by-query reasoning algorithms.

## 1. Introduction

Ontologies—formal conceptualizations of a domain of interest—have become increasingly important in computer science. They play a central role in many applications, such as the Semantic Web and biomedical information systems. The most widely used ontology languages are the Web Ontology Language (OWL) (Horrocks, Patel-Schneider, & van Harmelen, 2003) and its revision OWL 2 (Cuenca Grau, Horrocks, Motik, Parsia, Patel-Schneider, & Sattler, 2008), which have been standardized by the World Wide Web Consortium (W3C). The formal underpinning of the OWL family of languages is provided by description logics (DLs) (Baader, Calvanese, McGuinness, Nardi, & Patel-Schneider, 2007)—knowledge representation formalisms with well-understood computational properties.

Constructing ontologies is a labor-intensive task, so *reusing* (parts of) well-established ontologies is seen as key to reducing ontology development cost. Consequently, the problem of *ontology reuse* has recently received significant attention (Stuckenschmidt, Parent, & Spaccapietra, 2009; Lutz & Wolter, 2010; Lutz, Walther, & Wolter, 2007; Cuenca Grau, Horrocks, Kazakov, & Sattler, 2008, 2007; Doran, Tamma, & Iannone, 2007; Jiménez-Ruiz, Cuenca Grau, Sattler, Schneider, & Berlanga Llavori, 2008).

We discuss the problems of ontology reuse by means of an example from the health-care domain. In particular, ontologies are currently being used in several countries to describe electronic patient records (EPR). The representation of patients' data typically involves ontological descriptions of human anatomy, medical conditions, drugs and treatments, and so on. The latter domains have already been described in well-established reference ontologies, such as SNOMED-CT, GALEN, or the Foundational Model of Anatomy (FMA). In order to save resources, increase interoperability between applications, and rely on experts' knowledge, these and other reference ontologies should be reused whenever possible. For example, assume that some reference ontology $\mathcal{K}_h$ describes concepts such as the "ventricular septum defect"; then, one might reuse the terms from $\mathcal{K}_h$ in order to define an ontology $\mathcal{K}_v$ of concepts such as "patients having a ventricular septum defect," which might then be embedded in an EPR application.

To enable ontology reuse, OWL provides an *importing* mechanism: an ontology $\mathcal{K}_v$ can *import* another ontology $\mathcal{K}_h$, and the result is logically equivalent to $\mathcal{K}_v \cup \mathcal{K}_h$. OWL reasoners deal with imports by loading both ontologies and merging their contents, thus requiring physical access to the axioms of $\mathcal{K}_h$. The vendor of $\mathcal{K}_h$, however, may be reluctant to distribute (parts of) the contents of $\mathcal{K}_h$, as doing so might allow competitors to plagiarize $\mathcal{K}_h$. Moreover, $\mathcal{K}_h$ may contain information that is sensitive from a privacy point of view. Finally, one may want to impose a varying cost on the reuse of different parts of $\mathcal{K}_h$.

Rather than publishing the entire ontology, the vendor of $\mathcal{K}_h$ might want to freely distribute the symbols that describe organs and medical conditions, but without distributing the axioms describing these symbols. Furthermore, the vendor might want to completely hide the sensitive information from $\mathcal{K}_h$, such as the information about treatments. It should, however, be possible to reuse the published part of $\mathcal{K}_h$ without affecting the ontology's consequences; that is, if a part of $\mathcal{K}_h$ is used to construct an ontology $\mathcal{K}_v$, then any query $q$ mentioning only symbols from $\mathcal{K}_v$ should be answered over $\mathcal{K}_v$ and the respective part of $\mathcal{K}_h$ in the same way as this would be done over $\mathcal{K}_v \cup \mathcal{K}_h$. To stipulate that $\mathcal{K}_h$ should not be publicly available, we call the ontology $\mathcal{K}_h$ *hidden* and, by analogy, we call $\mathcal{K}_v$ *visible*.

Motivated by such scenarios, several approaches to hiding a subset $\Upsilon$ of the signature of $\mathcal{K}_h$ have been developed. For example, one possible approach is to publish an $\Upsilon$-*interpolant* of $\mathcal{K}_h$—an ontology that contains no symbols from $\Upsilon$ and that coincides with $\mathcal{K}_h$ on all logical consequences formed using the symbols not in $\Upsilon$ (Konev, Walther, & Wolter, 2009; Wang, Wang, Topor, Pan, & Antoniou, 2009; Wang, Wang, Topor, & Pan, 2008; Wang, Wang, Topor, & Zhang, 2010; Wang et al., 2008; Lutz & Wolter, 2011; Nikitina, 2011). Publishing an interpolant ensures that the sensitive information in $\mathcal{K}_h$ (i.e., the information about the symbols from $\mathcal{K}_h$ not mentioned in the interpolant) is not exposed in any way; furthermore, interpolants preserve all consequences of symbols not in $\Upsilon$ and have the additional advantage that the developers of $\mathcal{K}_v$ can reason over the union of $\mathcal{K}_v$ and the interpolant using off-the-shelf reasoners. The interpolation approach may, however, exhibit several drawbacks. First, an interpolant may exist only if $\mathcal{K}_h$ is expressed in a relatively weak ontology language and if it satisfies certain syntactic conditions (Konev et al., 2009). Second, although interpolants preserve logical consequences formed using symbols not in $\Upsilon$, they are not *robust under replacement* (Sattler, Schneider, & Zakharyaschev, 2009)—that is, the union of $\mathcal{K}_v$ and an $\Upsilon$-interpolant of $\mathcal{K}_h$ is not guaranteed to yield the same consequences as $\mathcal{K}_h \cup \mathcal{K}_v$ for a query $q$ involving only symbols from $\mathcal{K}_v$. Finally, an $\Upsilon$-interpolant of $\mathcal{K}_h$ can be exponentially larger than $\mathcal{K}_h$, and it may reveal more information than what is strictly needed. We refer the reader to Section 7 for a detailed discussion of the related work.

In this paper, we propose a novel approach to ontology reuse that addresses the problems outlined above by making $\mathcal{K}_h$ accessible via a limited query interface called an *oracle*. The oracle advertises a *public* subset $\Gamma$ of the signature of $\mathcal{K}_h$ (e.g., all symbols describing organs or medical conditions), and it can answer queries over $\mathcal{K}_h$ that are expressed in a particular query language and that use only the symbols from $\Gamma$. Under certain assumptions, a so-called *import-by-query* algorithm can reason over $\mathcal{K}_v \cup \mathcal{K}_h$ (e.g., determine the satisfiability of $\mathcal{K}_v \cup \mathcal{K}_h$) by posing queries to the oracle for $\mathcal{K}_h$, and without accessing any of the axioms from $\mathcal{K}_h$. Furthermore, reasoning can be performed without making the axioms of $\mathcal{K}_v$ available to $\mathcal{K}_h$, which is beneficial as $\mathcal{K}_v$ might also contain sensitive information from a privacy point of view. Finally, our framework can be applicable even in cases when the relevant interpolant for $\mathcal{K}_h$ does not exist.

In order to achieve these benefits, however, $\mathcal{K}_v$ must reuse the symbols from $\Gamma$ only in a syntactically restricted way, and the formal properties of import-by-query algorithms and the specific restrictions necessary for an import-by-query algorithm to exist depend on the oracle query language and the ontology languages used to express $\mathcal{K}_v$ and $\mathcal{K}_h$. In this paper, we explore the properties of import-by-query reasoning with languages ranging from the lightweight description logic $\mathcal{EL}$ (Baader, Brandt, & Lutz, 2005) to the expressive logic $\mathcal{ALCHOIQ}$ (Horrocks & Sattler, 2005), combined with the following types of oracles.

- Queries for *concept satisfiability oracles* are concepts constructed using the symbols in $\Gamma$ expressed in a particular DL; for each query, the oracle decides the satisfiability of the query concept w.r.t. $\mathcal{K}_h$.

- Queries for *ABox satisfiability oracles* are ABoxes constructed using the symbols in $\Gamma$; for each query, the oracle decides the satisfiability of the query ABox w.r.t. $\mathcal{K}_h$.

- Queries for *ABox entailment oracles* consist of an ABox and an assertion, both constructed using the symbols in $\Gamma$; for each query, the oracle determines whether the assertion is entailed by $\mathcal{K}_h$ and the query ABox.

Concept satisfiability, ABox satisfiability, and ABox entailment have been implemented in most state-of-the-art DL reasoners, so the above mentioned query languages seem like a natural foundation for practical implementations of our framework.

The main contributions of this paper are as follows:

1. We present the import-by-query framework, formalize the notions of an oracle and an import-by-query algorithm, and establish the connections between import-by-query algorithms based on different types of oracles.

2. We explore the limitations of our framework for a wide range of description logics and formulate precise conditions under which import-by-query algorithms fail to exist.

3. We identify sufficient conditions on the visible ontology $\mathcal{K}_v$ for which an import-by-query algorithm can be obtained.

4. We present a general hypertableau-based (Motik, Shearer, & Horrocks, 2009) import-by-query algorithm that relies on ABox satisfiability oracles and that is applicable to $\mathcal{K}_v$ and $\mathcal{K}_h$ given in the expressive description logic $\mathcal{ALCHIQ}$ (Horrocks & Sattler, 1999), provided that $\mathcal{K}_v$ satisfies our sufficient conditions.

5. Our general algorithm, however, is unlikely to be suitable for practice due to a high degree of nondeterminism. Therefore, we present a practical (goal-oriented) variant that is applicable whenever $\mathcal{K}_h$ is expressed in a Horn DL. This algorithm can be readily applied to ontologies expressed in the lightweight description logic $\mathcal{EL}$, but it is not guaranteed to be computationally optimal. Therefore, we also present a practical and computationally optimal algorithm that can be used if both $\mathcal{K}_v$ and $\mathcal{K}_h$ are expressed in $\mathcal{EL}$.

6. We establish the lower bounds on the size and the number of queries that an import-by-query algorithm may need to ask an oracle in order to solve a reasoning task.

Our results provide flexible and useful ways for ontology designers to ensure selective access to their ontologies, as well as a family of reasoning algorithms that provide a starting point for implementation and optimization. Furthermore, we believe our techniques can also be adapted to other settings, such as distributed ontology reasoning, or collaborative ontology development scenarios in which ontology developers have restricted access to the parts of the ontology developed by others.

## 2. Preliminaries

In this section, we recapitulate the description logic notation used in this paper, we present an overview of various hypertableau reasoning algorithms for description logics (Motik et al., 2009), and we recapitulate various notions of modular ontology reuse (Lutz, Walther, & Wolter, 2007; Cuenca Grau, Horrocks, Kazakov, & Sattler, 2008; Konev, Lutz, Walther, & Wolter, 2008).

### 2.1 Description Logics

The syntax of the description logic $\mathcal{ALCHOIQ}$ is defined w.r.t. pairwise-disjoint countably infinite sets of *atomic concepts* $N_C$, *atomic roles* $N_R$, and *named individuals* $N_I$. Set $N_C$ contains a distinguished infinite subset $N_O \subseteq N_C$ of *nominal concepts* (or simply *nominals*). A *role* is either an atomic role or an *inverse role* $R^-$ for $R$ an atomic role.

The set of *concepts* is the smallest set containing $\top$, $A$, $\neg C$, $C_1 \sqcap C_2$, $\exists R.C$ (existential restriction), and $\geq n\,R.C$ (cardinality restriction), for $A$ an atomic concept, $C$, $C_1$, and $C_2$ concepts, $R$ a role, and $n$ a nonnegative integer. Furthermore, $\bot$, $C_1 \sqcup C_2$, $\forall R.C$, and $\leq n\,R.C$ are abbreviations of $\neg\top$, $\neg(\neg C_1 \sqcap \neg C_2)$, and $\neg(\exists R.\neg C)$, and $\neg(\geq n{+}1\,R.C)$, respectively. We also often treat concepts of the form $\exists R.C$ as abbreviations of $\geq 1\,R.C$.

A *concept inclusion axiom* has the form $C_1 \sqsubseteq C_2$ for $C_1$ and $C_2$ concepts, a *concept equivalence* $C_1 \equiv C_2$ is an abbreviation for $C_1 \sqsubseteq C_2$ and $C_2 \sqsubseteq C_1$, and a *concept definition* is a concept equivalence of the form $A \equiv C$ with $A$ an atomic concept. A *role inclusion axiom* has the form $R_1 \sqsubseteq R_2$ for $R_1$ and $R_2$ roles. A *TBox axiom* is either a concept inclusion axiom or a role inclusion axiom. A *TBox* $\mathcal{T}$ is a finite set of TBox axioms. An *assertion* has the form $C(a)$, $R(a,b)$, $\neg R(a,b)$, $a \approx b$, or $a \not\approx b$, for $C$ a concept, $R$ a role, and $a$ and $b$ individuals. An *ABox* $\mathcal{A}$ is a finite set of assertions. An ABox is *normalized* if it contains only assertions of the form $A(a)$, $\neg A(a)$, $R(a,b)$, $\neg R(a,b)$, and $a \not\approx b$, where $A$ is an atomic concept and $R$ is an atomic role. An *axiom* is either a TBox axiom or an assertion. A *knowledge base* $\mathcal{K} = \mathcal{T} \cup \mathcal{A}$ consists of a TBox $\mathcal{T}$ and an ABox $\mathcal{A}$.

Table 1: Model-Theoretic Semantics of $\mathcal{ALCHOIQ}$

| Interpretation of Roles |
|---|
| $(R^-)^I = \{\langle y, x \rangle \mid \langle x, y \rangle \in R^I\}$ |

| Interpretation of Concepts |
|---|
| $\top^I = \triangle^I$ |
| $(\neg C)^I = \triangle^I \setminus C^I$ |
| $(C_1 \sqcap C_2)^I = C_1^I \cap C_2^I$ |
| $(\exists R.C)^I = \{x \mid \exists y : \langle x, y \rangle \in R^I \wedge y \in C^I\}$ |
| $(\geq n\,R.C)^I = \{x \mid \sharp\{y \mid \langle x, y \rangle \in R^I \wedge y \in C^I\} \geq n\}$ |

| Satisfaction of Axioms in an Interpretation | |
|---|---|
| $I \models C \sqsubseteq D$ | iff $C^I \subseteq D^I$ |
| $I \models R_1 \sqsubseteq R_2$ | iff $R_1^I \subseteq R_2^I$ |
| $I \models C(a)$ | iff $a^I \in C^I$ |
| $I \models R(a, b)$ | iff $\langle a^I, b^I \rangle \in R^I$ |
| $I \models \neg R(a, b)$ | iff $\langle a^I, b^I \rangle \notin R^I$ |
| $I \models a \approx b$ | iff $a^I = b^I$ |
| $I \models a \not\approx b$ | iff $a^I \neq b^I$ |

A *signature* is a set of atomic concepts and atomic roles. For $\alpha$ a concept, a role, an axiom, or a set of axioms, the *signature* of $\alpha$, written $\mathsf{sig}(\alpha)$, is the set of atomic concepts and atomic roles occurring in $\alpha$.[1]

The cardinality of a set $S$ is written $\sharp S$. An *interpretation* $I = (\triangle^I, \cdot^I)$ consists of a nonempty *domain* set $\triangle^I$ and a function $\cdot^I$ that assigns an object $a^I \in \triangle^I$ to each individual $a$, a set $A^I \subseteq \triangle^I$ to each atomic concept $A$ such that $A \in N_O$ implies $\sharp A^I = 1$, and a relation $R^I \subseteq \triangle^I \times \triangle^I$ to each atomic role $R$. Table 1 defines the extension of $\cdot^I$ to roles and concepts, as well as the satisfaction of axioms in $I$. An interpretation $I$ is a *model* of $\mathcal{K}$, written $I \models \mathcal{K}$, if $I$ satisfies all axioms in $\mathcal{K}$; if such $I$ exists, then $\mathcal{K}$ is *satisfiable*. A concept $C$ is *satisfiable* w.r.t. $\mathcal{K}$ if a model $I$ of $\mathcal{K}$ exists such that $C^I \neq \emptyset$.

Sometimes, nominal concepts are defined as having the form $\{a\}$ for $a$ an individual, and such a concept is interpreted as $(\{a\})^I = \{a^I\}$; that is, a nominal concept contains precisely the given individual. The drawback of such a definition is that it blurs the distinction between concepts and individuals at the syntactic level. Such a distinction is important for the import-by-query framework since our framework supports sharing concepts, but not individuals. In this paper we thus use the above given alternative definition, where nominals are "special" atomic concepts with a singleton interpretation. It is well known that these two definitions are equally expressive (Baader et al., 2007).

Some of our results use a general notion of a description logic. Formally, we define a *description logic* $\mathcal{DL}$ as a pair consisting of a set of concepts and a set of knowledge bases. We call the elements of the former set $\mathcal{DL}$-*concepts* and the elements of the latter set $\mathcal{DL}$-*knowledge bases*. Each concept in a $\mathcal{DL}$-knowledge base must be a $\mathcal{DL}$-concept. A $\mathcal{DL}$-*TBox* (resp. $\mathcal{DL}$-*ABox*) is a $\mathcal{DL}$-knowledge base containing no assertions (resp. no TBox axioms).

---

1. Note that we are treating nominals as special atomic concepts (and not as individuals); hence, $\mathsf{sig}(\alpha)$ includes the nominals, but not the individuals occurring in $\alpha$.

A $\mathcal{DL}$-*TBox axiom* (resp. $\mathcal{DL}$-*assertion*) is a TBox axiom (resp. assertion) that occurs in some $\mathcal{DL}$-knowledge base. A description logic $\mathcal{DL}_1$ is a *fragment* of $\mathcal{DL}_2$ (or, conversely, $\mathcal{DL}_2$ extends $\mathcal{DL}_1$) if each $\mathcal{DL}_1$-concept is a $\mathcal{DL}_2$-concept and each $\mathcal{DL}_1$-knowledge base is a $\mathcal{DL}_2$-knowledge base. Since the "unqualified" notions of a concept and knowledge base are defined for $\mathcal{ALCHOIQ}$, our definitions imply that each description logic considered in this paper is a fragment of $\mathcal{ALCHOIQ}$.

Let $\mathcal{DL}_1$ and $\mathcal{DL}_2$ be description logics. We say that $\mathcal{DL}_1$ *allows for* $\mathcal{DL}_2$-*definitions* if, for each $\mathcal{DL}_1$-knowledge base $\mathcal{K}$, each atomic concept $A$, and each $\mathcal{DL}_2$-concept $C$, we have that $\mathcal{K} \cup \{A \equiv C\}$ is a $\mathcal{DL}_1$-knowledge base. Furthermore, $\mathcal{DL}_1$ has the *finite model property* if each satisfiable $\mathcal{DL}_1$-knowledge base has a model with a finite domain.

The description logic $\mathcal{ALC}$ is obtained from $\mathcal{ALCHOIQ}$ by disallowing nominal concepts ($\mathcal{O}$), inverse roles ($\mathcal{I}$), role inclusion axioms ($\mathcal{H}$), and cardinality restrictions ($\mathcal{Q}$). The description logics between $\mathcal{ALC}$ and $\mathcal{ALCHOIQ}$ are named by appending combinations of letters $\mathcal{O}$, $\mathcal{H}$, $\mathcal{I}$, and $\mathcal{Q}$ to $\mathcal{ALC}$.

The DL $\mathcal{EL}$ (Baader et al., 2005) (resp. $\mathcal{FL}_0$, see Baader et al., 2007) is obtained from $\mathcal{ALC}$ by allowing only concepts of the form $\top$, $\bot$, $A$, $C_1 \sqcap C_2$, and $\exists R.C$ (resp. $\forall R.C$) for $A$ and $R$ atomic, and by allowing only assertions of the form $C(a)$ or $R(a,b)$, with $C$ an $\mathcal{EL}$ (resp. $\mathcal{FL}_0$) concept and $R$ an atomic role. In recent years, significant effort has been devoted to the development of DL languages with good computational properties, such as $\mathcal{EL}$, DL-Lite (Calvanese, De Giacomo, Lembo, Lenzerini, & Rosati, 2007), and Horn-$\mathcal{SHIQ}$ (Hustadt, Motik, & Sattler, 2005). An $\mathcal{ALCHIQ}$ knowledge base is *Horn* if it is expressed in the Horn-$\mathcal{SHIQ}$ fragment of $\mathcal{ALCHIQ}$.

For an ABox $\mathcal{A}$, with $\mathcal{G}(\mathcal{A})$ we denote the graph whose nodes are precisely the individuals occurring in $\mathcal{A}$, and that contains an undirected edge between individuals $a$ and $b$ if and only if $a = b$ or both $a$ and $b$ occur together in an assertion in $\mathcal{A}$. Individuals $a$ and $b$ are *connected* in $\mathcal{A}$ if $a$ and $b$ are connected in $\mathcal{G}(\mathcal{A})$; furthermore, $\mathcal{A}$ is *connected* if all pairs of individuals occurring in $\mathcal{A}$ are connected. An ABox $\mathcal{A}' \subseteq \mathcal{A}$ is a *connected component* of $\mathcal{A}$ if $\mathcal{G}(\mathcal{A}')$ is a connected component of $\mathcal{G}(\mathcal{A})$.

## 2.2 Hypertableau Reasoning Algorithm

The hypertableau calculus by Motik et al. (2009) decides the satisfiability of an $\mathcal{ALCHOIQ}$ knowledge base $\mathcal{K}$. As we show in Section 4.1, the presence of nominals precludes the existence of an import-by-query algorithm; hence, in this section we present an overview of a simplified version of the algorithm that is applicable if $\mathcal{K}$ is an $\mathcal{ALCHIQ}$ knowledge base.

The algorithm first preprocesses $\mathcal{K}$ into a set of rules $\mathcal{R}$—implications interpreted under first-order semantics—and a normalized ABox $\mathcal{A}$ such that $\mathcal{K}$ is equisatisfiable with $\mathcal{R} \cup \mathcal{A}$. Preprocessing consists of three steps. First, transitivity axioms are eliminated from $\mathcal{K}$ by encoding them using concept inclusions. Second, axioms are normalized and complex concepts are replaced with atomic ones in a way similar to the structural transformation for first-order logic. Third, the normalized axioms are translated into rules by using the correspondence between description and first-order logic. We omit the details of the preprocessing for the sake of brevity; Motik et al. (2009) present all the relevant details. Preprocessing produces so-called HT-rules—syntactically restricted rules on which the hypertableau calculus is guaranteed to terminate; the precise syntactic form of HT-rules is described in Section 2.2.1.

After preprocessing, the satisfiability of $\mathcal{R} \cup \mathcal{A}$ is decided using the hypertableau calculus, which is described in Section 2.2.2.

### 2.2.1 HT-Rules

Let $N_V$ be a set of variables disjoint with the set of individuals $N_I$. An *atom* is an expression of the form $C(s)$ (a *concept* atom), $R(s,t)$ (a *role* atom), or $s \approx t$ (an *equality* atom), where $s, t \in N_V \cup N_I$, $C$ is a concept, and $R$ is a role. A *rule* is an expression of the form

$$U_1 \wedge \ldots \wedge U_m \rightarrow V_1 \vee \ldots \vee V_n \tag{1}$$

where $U_i$ and $V_j$ are atoms, $m \geq 0$, and $n \geq 0$. Conjunction $U_1 \wedge \ldots \wedge U_m$ is called the *body*, and disjunction $V_1 \vee \ldots \vee V_n$ is called the *head* of the rule. The empty body and the empty head are written as $\top$ and $\bot$, respectively. Rules are interpreted as universally quantified FOL implications in the usual way. A rule is *Horn* if it contains at most one head atom.

An *HT-rule* is a rule of the form

$$\begin{aligned} \bigwedge A_i(x) \wedge \bigwedge R_{ij}(x, y_i) \wedge \bigwedge S_{ij}(y_i, x) \wedge \bigwedge B_{ij}(y_i) \rightarrow \\ \bigvee C_i(x) \vee \bigvee R'_{ij}(x, y_i) \vee \bigvee S'_{ij}(y_i, x) \vee \bigvee D_{ij}(y_i) \vee \bigvee y_i \approx y_j \end{aligned} \tag{2}$$

where $R_{ij}$, $S_{ij}$, $R'_{ij}$, and $S'_{ij}$ are atomic roles; $A_i$, $B_{ij}$, and $D_{ij}$ are atomic concepts; and $C_i$ are either atomic concepts or concepts of the form $\geq n\,R.A$ or $\geq n\,R.\neg A$. In addition, each variable $y_i$ occurring in an HT-rule is required to occur in a body atom of the form $R_{ij}(x, y_i)$ or $S_{ij}(y_i, x)$. Intuitively, the body and the head of HT-rules can be seen as being star-shaped: "center variable" $x$ represents the center of the star, and "branch variables" $y_i$ can be connected to the center only through role atoms. Such shape ensures that satisfiable HT-rules will always have a tree-like model—a property that can be used to explain the good computational properties of many DLs.

As Motik et al. (2009) have shown, the preprocessing of $\mathcal{K}$ produces an equisatisfiable set of HT-rules and a normalized ABox; furthermore, if $\mathcal{K}$ is Horn, then the resulting set contains only Horn HT-rules. Furthermore, if certain description logic constructors are not used in $\mathcal{K}$, then $\mathcal{R}$ satisfies certain syntactic restrictions as discussed next.

- If $\mathcal{K}$ does not use cardinality restrictions, then no HT-rule $\varrho \in \mathcal{R}$ contains an atom of the form $y_i \approx y_j$ in the head.

- If $\mathcal{K}$ does not use inverse roles, then no HT-rule $\varrho \in \mathcal{R}$ contains an atom of the form $S'_{ij}(y_i, x)$ in the head or an atom of the form $S_{ij}(y_i, x)$ in the body.

- If $\mathcal{K}$ does not use role hierarchies, then no HT-rule $\varrho \in \mathcal{R}$ contains a role atom in the head.

As an example, consider the following knowledge base $\mathcal{K}$ and the corresponding set of HT-rules $\mathcal{R}$ obtained from $\mathcal{K}$.

$$A \sqsubseteq \exists R.B \qquad \rightsquigarrow \qquad A(x) \rightarrow \exists R.B(x) \tag{3}$$

$$A \sqsubseteq \exists R.C \qquad \rightsquigarrow \qquad A(x) \rightarrow \exists R.C(x) \tag{4}$$

$$\top \sqsubseteq \,\leq 1\,R.\top \qquad\qquad \rightsquigarrow \qquad\qquad R(x,y_1) \wedge R(x,y_2) \rightarrow y_1 \approx y_2 \qquad (5)$$

$$B \sqcap C \sqsubseteq D \qquad\qquad \rightsquigarrow \qquad\qquad B(x) \wedge C(x) \rightarrow D(x) \qquad (6)$$

$$\exists R.D \sqsubseteq E \qquad\qquad \rightsquigarrow \qquad\qquad R(x,y) \wedge D(y) \rightarrow E(x) \qquad (7)$$

Note that $\mathcal{R}$ is a set of Horn HT-rules. Note also that $\mathcal{K}$ uses a cardinality restriction $\leq 1\,R.\top$, so $\mathcal{R}$ contains a rule with an equality atom in the head. Furthermore, $\mathcal{K}$ does not use role hierarchies, so no rule in $\mathcal{R}$ contains a role atom in the head. Finally, $\mathcal{K}$ does not use inverse roles, so each role atom occurring in the body of a rule in $\mathcal{R}$ contains the center variable $x$ in the first position and a branch variable $y_i$ in the second position.

When applied to an $\mathcal{EL}$ knowledge base, the transformation by Motik et al. (2009) produces *$\mathcal{EL}$-rules*—HT-rules of the form (8) in which $C$ is either an atomic concept or a concept of the form $\exists R.A$ with $A$ an atomic concept.

$$\bigwedge_{i=1}^{k} A_i(x) \wedge \bigwedge_{i=1}^{m} \left[ R_i(x,y_i) \wedge \bigwedge_{j=1}^{m_i} B_{ij}(y_i) \right] \rightarrow C(x) \qquad (8)$$

Note that all the rules in our previous example except for the third one (which uses equality in the head) are $\mathcal{EL}$-rules.

### 2.2.2 HYPERTABLEAU CALCULUS FOR HT-RULES

Given an arbitrary set of HT-rules $\mathcal{R}$ and a normalized ABox $\mathcal{A}$, satisfiability of $\mathcal{R} \cup \mathcal{A}$ can be decided using the calculus described in Definition 1.

**Definition 1. *Individuals.*** *For a set of named individuals $N_I$, the set of all individuals $N_X$ is inductively defined as the smallest set such that $N_I \subseteq N_X$ and, if $x \in N_X$, then $x.i \in N_X$ for each integer $i$. The individuals in $N_X \setminus N_I$ are unnamed. An individual $x.i$ is a* successor *of $x$, and $x$ is a* predecessor *of $x.i$;* descendant *and* ancestor *are the transitive closures of successor and predecessor, respectively.*

***Pairwise Anywhere Blocking.*** *The label $\mathcal{L}_{\mathcal{A}}(s)$ of an individual $s$ and the label $\mathcal{L}_{\mathcal{A}}(s,t)$ of an individual pair $\langle s,t \rangle$ in an ABox $\mathcal{A}$ are defined as follows:*

$$\mathcal{L}_{\mathcal{A}}(s) \;=\; \{A \mid A(s) \in \mathcal{A} \text{ and } A \text{ is atomic}\}$$
$$\mathcal{L}_{\mathcal{A}}(s,t) \;=\; \{R \mid R(s,t) \in \mathcal{A}\}$$

*Let $\prec$ be a strict ordering on $N_X$ containing the ancestor relation. By induction on $\prec$, we assign to each individual $s$ in $\mathcal{A}$ a blocking status as follows.*

- *Individual $s$ is directly blocked by individual $t$ iff the following conditions hold, for $s'$ and $t'$ the predecessors of $s$ and $t$, respectively:*

  - *$s$ and $t$ are unnamed, $t$ is not blocked, and $t \prec s$;[2]*
  - *$\mathcal{L}_{\mathcal{A}}(s) = \mathcal{L}_{\mathcal{A}}(t)$ and $\mathcal{L}_{\mathcal{A}}(s') = \mathcal{L}_{\mathcal{A}}(t')$; and*
  - *$\mathcal{L}_{\mathcal{A}}(s,s') = \mathcal{L}_{\mathcal{A}}(t,t')$ and $\mathcal{L}_{\mathcal{A}}(s',s) = \mathcal{L}_{\mathcal{A}}(t',t)$.*

---

2. When blocking is used with $\mathcal{ALCHOIQ}$ knowledge bases, individuals $s'$ and $t'$ are also required to be unnamed; however, this restriction is not needed for $\mathcal{ALCHIQ}$ knowledge bases.

Table 2: Hypertableau Derivation Rules

| | Derivation Rules for HT-rules |
|---|---|
| *Hyp*-rule | If 1. $\varrho \in \mathcal{R}$ of the form (1) and<br>2. a mapping $\sigma$ from the variables in $\varrho$ to the individuals in $\mathcal{A}$ exists where<br>2.1 $\sigma(x)$ is not indirectly blocked for each variable $x$ in $\varrho$,<br>2.2 $\sigma(U_i) \in \mathcal{A}$ for each $1 \le i \le m$, and<br>2.3 $\sigma(V_j) \notin \mathcal{A}$ for each $1 \le j \le n$,<br>then $\mathcal{A}_1 := \mathcal{A} \cup \{\bot\}$ if $n = 0$;<br>$\mathcal{A}_j := \mathcal{A} \cup \{\sigma(V_j)\}$ for $1 \le j \le n$ otherwise. |
| $\ge$-rule | If 1. $\ge n\, R.C(s) \in \mathcal{A}$ such that $s$ is not blocked in $\mathcal{A}$ and<br>2. no individuals $u_1, \ldots, u_n$ in $\mathcal{A}$ exist such that<br>$\quad \{\mathsf{ar}(R, s, u_i), C(u_i) \mid 1 \le i \le n\} \cup \{u_i \not\approx u_j \mid 1 \le i < j \le n\} \subseteq \mathcal{A}$,<br>then $\mathcal{A}_1 := \mathcal{A} \cup \{\mathsf{ar}(R, s, t_i),\, C(t_i) \mid 1 \le i \le n\} \cup \{t_i \not\approx t_j \mid 1 \le i < j \le n\}$<br>$\quad$ where $t_1, \ldots, t_n$ are fresh successors of $s$. |
| $\approx$-rule | If 1. $s \approx t \in \mathcal{A}$ such that $s \ne t$ and neither $s$ not $t$ is indirectly blocked<br>then $\mathcal{A}_1 := \mathsf{merge}_{\mathcal{A}}(s \to t)$ if $t$ is named or $s$ is a descendant of $t$, and<br>$\quad \mathcal{A}_1 := \mathsf{merge}_{\mathcal{A}}(t \to s)$ otherwise. |
| $\bot$-rule | If 1. $s \not\approx s \in \mathcal{A}$ or $\{A(s), \neg A(s)\} \subseteq \mathcal{A}$ or $\{R(s,t), \neg R(s,t)\} \subseteq \mathcal{A}$<br>$\quad$ such that neither $s$ nor $t$ is indirectly blocked and<br>2. $\bot \notin \mathcal{A}$<br>then $\mathcal{A}_1 := \mathcal{A} \cup \{\bot\}$. |

The $\exists$-rule for $\mathcal{EL}$-rules

| | |
|---|---|
| $\exists$-rule | If $\quad \exists R.A(s) \in \mathcal{A}$ and $\{R(s, a_A), A(a_A)\} \nsubseteq \mathcal{A}$<br>then $\mathcal{A}_1 := \mathcal{A} \cup \{R(s, a_A), A(a_A)\}$ |

- *Individual $s$ is* indirectly blocked *iff its predecessor is blocked.*

- *Individual $s$ is* blocked *iff it is either directly or indirectly blocked.*

**Pruning and Merging.** *The ABox $\mathsf{prune}_{\mathcal{A}}(s)$ is obtained from $\mathcal{A}$ by removing all assertions containing a descendant of $s$. The ABox $\mathsf{merge}_{\mathcal{A}}(s \to t)$ is obtained from $\mathsf{prune}_{\mathcal{A}}(s)$ by replacing $s$ with $t$ in all assertions.*

**Clash.** *An ABox $\mathcal{A}$ contains a* clash *if $\bot \in \mathcal{A}$; otherwise, $\mathcal{A}$ is clash-free.*

**Derivation Rules.** *The derivation rules consist of the Hyp-, $\ge$-, $\approx$-, and $\bot$-rule from Table 2, which, given $\mathcal{R}$ and a clash-free ABox $\mathcal{A}$, derive the ABoxes $\langle \mathcal{A}_1, \ldots, \mathcal{A}_n \rangle$. In the Hyp-rule, $\sigma(U)$ is obtained from $U$ by replacing each variable $x$ with $\sigma(x)$. For a role $R$ and individuals $s$ and $t$, function $\mathsf{ar}(R, s, t)$ returns assertion $R(s, t)$ if $R$ is atomic, or assertion $S(t, s)$ if $R$ is an inverse role and $R = S^-$.*

**Derivation.** *A* derivation *for $\mathcal{R}$ and $\mathcal{A}$ is a pair $(T, \rho)$ where $T$ is a finitely branching tree and $\rho$ labels the nodes of $T$ with ABoxes such that (i) $\rho(\epsilon) = \mathcal{A}$ for $\epsilon$ the root, and (ii) for each node $t$, if a derivation rule is applicable to $\mathcal{R}$ and $\rho(t)$, then $t$ has children $t_1, \ldots, t_n$ such that $\langle \rho(t_1), \ldots, \rho(t_n) \rangle$ are the result of applying one derivation rule to $\mathcal{R}$ and $\rho(t)$. The algorithm returns $\mathsf{t}$ if some derivation for $\mathcal{R}$ and $\mathcal{A}$ has a leaf node labeled with a clash-free ABox, and $\mathsf{f}$ otherwise.*

The *Hyp*-rule is similar to the one of the hypertableau calculus for first-order logic: given an HT-rule of he form (1) and an ABox $\mathcal{A}$, the *Hyp*-rule tries to unify the atoms $U_1, \ldots, U_m$ with a subset of the assertions in $\mathcal{A}$; if a unifier $\sigma$ is found, the rule nondeterministically derives $\sigma(V_j)$ for some $1 \leq j \leq n$. For example, given the rule $A(x) \rightarrow \exists R.C(x) \vee D(x)$ and an assertion $A(a)$, the *Hyp*-rule derives either $\exists R.C(a)$ or $D(a)$. The $\geq$-rule deals with existential quantifiers; for example, given $\exists R.C(a)$, the rule introduces a fresh individual $t$ and derives $R(a, t)$ and $C(t)$. The $\approx$-rule deals with equality; for example, given $a \approx b$, the rule replaces the individual $a$ in all assertions with the individual $b$. Finally, the $\perp$-rule detects obvious contradictions such as $A(a)$ and $\neg A(a)$, $R(a, b)$ and $\neg R(a, b)$, or $a \not\approx a$.

Since $\mathcal{ALCHIQ}$ allows for cyclic concept inclusions of the form $C \sqsubseteq \exists R.C$, termination of the hypertableau calculus requires a *blocking* mechanism to prevent the $\geq$-rule from generating infinite sequences of successors. When an individual $s$ is directly blocked by another individual $t$, the $\geq$-rule is no longer applicable to $s$, which prevents the introduction of fresh successors of $s$. Furthermore, all descendants of $s$ are then indirectly blocked, which prevents the application of any of the rules in Table 2 to the descendants of $s$.

If a derivation for $\mathcal{R}$ and $\mathcal{A}$ exists in which a leaf node is labeled with a clash-free ABox $\mathcal{A}'$, then a model of $\mathcal{R} \cup \mathcal{A}$ can be constructed from $\mathcal{A}'$ via a well-known technique called *unraveling*. Models of $\mathcal{R} \cup \mathcal{A}$ obtained in such a way are called *canonical forest models*, and Motik et al. (2009) discuss in depth the properties of such models.

Let $\mathcal{R}$ be the set of HT-rules (3)–(7) given in Section 2.2.1, and let $\mathcal{A} = \{A(a), \neg E(a)\}$; we next show how to demonstrate using the hypertableau algorithm that $\mathcal{R} \cup \mathcal{A}$ is unsatisfiable. By applying the *Hyp*-rule to $A(a)$, we derive $\exists R.B(a)$ and $\exists R.C(a)$. Next, by applying the $\geq$-rule to $\exists R.B(a)$ we derive $R(a, t_1)$ and $B(t_1)$; and by applying the $\geq$-rule to $\exists R.C(a)$ we derive $R(a, t_2)$ and $C(t_2)$. Individuals $t_1$ and $t_2$ are fresh successors of $s$ and are actually of the form $s.1$ and $s.2$; however, for clarity we write them simply as $t_1$ and $t_2$. By applying the *Hyp*-rule to $R(a, t_1)$ and $R(a, t_2)$, we derive $t_1 \approx t_2$. Furthermore, to apply the $\approx$-rule to $t_1 \approx t_2$, we must replace $t_1$ with $t_2$ in all assertions; thus, we replace $R(a, t_1)$ and $B(t_1)$ with $R(a, t_2)$ and $B(t_2)$, respectively. Next, by applying the *Hyp*-rule to $B(t_2)$ and $C(t_2)$ we derive $D(t_2)$. Next, by applying the *Hyp*-rule to $R(a, t_2)$ and $D(t_2)$ we derive $E(a)$. Finally, by applying the the $\perp$-rule to $E(a)$ and $\neg E(a)$ we derive $\perp$. We have thus constructed a derivation for $\mathcal{R}$ and $\mathcal{A}$ whose (only) leaf contains a clash, and so $\mathcal{R} \cup \mathcal{A}$ is unsatisfiable.

### 2.2.3 Hypertableau Algorithm for $\mathcal{EL}$-rules

Since any $\mathcal{EL}$ knowledge base is an $\mathcal{ALCHIQ}$ knowledge base as well, the hypertableau algorithm can straightforwardly be applied to $\mathcal{EL}$ KBs. Motik and Horrocks (2008) showed, however, that a worst-case optimal algorithm can be obtained by modifying the $\geq$-rule. This modified algorithm works on a set $\mathcal{R}$ of $\mathcal{EL}$-*rules*.

The following algorithm checks satisfiability of $\mathcal{R} \cup \mathcal{A}$, for $\mathcal{R}$ a set of $\mathcal{EL}$-rules and $\mathcal{A}$ a normalized ABox.

**Definition 2.** *For each named individual $a \in N_I$ and each atomic concept $A \in N_C$, let $a_A$ be a fresh individual that is uniquely associated with $a$ and $A$. The hypertableau algorithm for $\mathcal{EL}$ is the same as the one described in Definition 1, but the derivation rules include the Hyp-, $\perp$-, and $\exists$-rule from Table 2.*

## 2.3 Modularity

Let $\mathcal{K}_v$ be a knowledge base that reuses a knowledge base $\mathcal{K}_h$, and let $\Gamma$ be the subset of $\mathsf{sig}(\mathcal{K}_h)$ that is being reused in $\mathcal{K}_v$—that is, $\Gamma = \mathsf{sig}(\mathcal{K}_h) \cap \mathsf{sig}(\mathcal{K}_v)$. It is often beneficial if $\mathcal{K}_v$ reuses $\mathcal{K}_h$ in a modular way; intuitively, this is the case if the knowledge base $\mathcal{K}_v$ does not "affect the meaning" of the symbols in $\Gamma$ (Lutz, Walther, & Wolter, 2007; Cuenca Grau, Horrocks, Kazakov, & Sattler, 2008; Konev, Lutz, Walther, & Wolter, 2008). Two different notions of modularity have been considered in literature, each providing a different formal account of what it means for $\mathcal{K}_v$ to "affect the meaning" of the symbols in $\Gamma$.

A knowledge base $\mathcal{K}_v$ is *deductively modular* w.r.t. a signature $\Gamma$ if, for all concepts $C$ and $D$ expressed in the same description logic as $\mathcal{K}_v$ such that $\mathsf{sig}(C) \subseteq \Gamma$ and $\mathsf{sig}(D) \subseteq \Gamma$, we have that $\mathcal{K}_v \models C \sqsubseteq D$ implies $\emptyset \models C \sqsubseteq D$. That is, the axioms of $\mathcal{K}_v$ must not give rise to nontrivial logical consequences that involve only the symbols from $\Gamma$.

A knowledge base $\mathcal{K}_v$ is *semantically modular* w.r.t. a signature $\Gamma$ if, for each interpretation $I = (\triangle^I, \cdot^I)$ for the symbols in $\Gamma$, there exists an interpretation $J = (\triangle^J, \cdot^J)$ such that $\triangle^I = \triangle^J$, $X^I = X^J$ for each $X \in \Gamma$, and $J \models \mathcal{K}_v$. That is, the axioms of $\mathcal{K}_v$ are not allowed to impose any constraints on the interpretation of the symbols from $\Gamma$.

Semantic modularity is stronger than the deductive one: if $\mathcal{K}_v$ is semantically modular w.r.t. $\Gamma$, then it is also deductively modular w.r.t. $\Gamma$; the converse does not hold necessarily. Deciding whether a knowledge base $\mathcal{K}_v$ is deductively or semantically modular w.r.t. a signature $\Gamma$ is a very hard computational problem for most DLs, and it is often undecidable (Lutz et al., 2007; Konev et al., 2008). Cuenca Grau, Horrocks, Kazakov, and Sattler (2008) have defined several practically useful sufficient syntactic conditions that guarantee semantic modularity.

## 3. The Import-by-Query Framework

In this section we introduce our framework. We first present a motivating example, after which we proceed with a formalization of the import-by-query problem.

Consider a medical research company (MRC) that has developed a knowledge base of human anatomy. This knowledge base contains concepts describing organs such as **Heart** and **TV** (tricuspid valve); medical conditions such as **CHD** (congenital heart defect), **VSD** (ventricular septum defect), and **AS** (aortic stenosis); and treatments such as Surgery. The roles **part**, **con**, and treatment relate organs with their parts, medical conditions, and treatments, respectively, and they are used to define concepts such as **VSD_Heart** (a heart with a ventricular septum defect) and **Sur_Heart** (a heart that requires surgical treatment). We focus on reusing schema knowledge, so we assume that the knowledge base consists only of a TBox $\mathcal{T}_h$, which is shown in Table 3. Assume that MRC wants to freely distribute information about organs and conditions, but hide the information about treatments. Thus, MRC identifies a set $\Gamma$ of *public* symbols of $\mathcal{T}_h$; we write these symbols in **bold**, and the remaining *private* symbols in sans serif. MRC does not want to distribute the axioms of $\mathcal{T}_h$, as this might allow competitors to copy parts of $\mathcal{T}_h$; therefore, we say that knowledge base $\mathcal{T}_h$ is *hidden*.

Consider also a health-care provider (HCP) that reuses $\mathcal{T}_h$ to describe types of patients such as *VSD_Patient* (patients with a ventricular septum defect), *HS_Patient* (patients requiring heart surgery), *AS_Patient* (patients with aortic stenosis), *EA_Patient* (patients with Ebstein's anomaly), and $TVD\_Patient$ (patients with a tricuspid valve defect). Since the TBox $\mathcal{T}_h$ does not describe Ebstein's anomaly, HCP defines $EA\_Heart$ as a heart with a ventricular septum defect and with a displaced tricuspid valve $Dis\_TV$; furthermore, it defines a displaced tricuspid valve as abnormal, and $Ab\_TV\_Heart$ as a heart with an abnormal tricuspid valve. In general, HCP's knowledge base could contain ABox assertions, so we denote the knowledge base with $\mathcal{K}_v$ and call it *visible*. The axioms of $\mathcal{K}_v$ are shown in Table 3, and the private symbols of $\mathcal{K}_v$ are written in *italic*. HCP can use the combined knowledge base $\mathcal{K}_v \cup \mathcal{T}_h$ to deduce that $VSD\_Patient \sqsubseteq HS\_Patient$ (patients with ventricular septum defect require heart surgery) and $EA\_Patient \sqsubseteq TVD\_Patient$ (patients with Ebstein's anomaly are a kind of patients with a tricuspid valve defect).

Table 3: Example Knowledge Bases

| Hidden Knowledge Base $\mathcal{T}_h$ | |
|---|---|
| $\gamma_1$ | $\mathbf{Heart} \sqsubseteq \mathbf{Organ} \sqcap \exists \mathbf{part}.\mathbf{TV}$ |
| $\gamma_2$ | $\mathbf{VSD} \sqsubseteq \mathbf{CHD}$ |
| $\gamma_3$ | $\mathbf{AS} \sqsubseteq \mathbf{CHD}$ |
| $\gamma_4$ | $\mathbf{VSD\_Heart} \equiv \mathbf{Heart} \sqcap \exists \mathbf{con}.\mathbf{VSD}$ |
| $\gamma_5$ | $\mathbf{VSD\_Heart} \sqsubseteq \exists \mathsf{treatment}.\mathsf{Surgery}$ |
| $\gamma_6$ | $\mathbf{Sur\_Heart} \equiv \mathbf{Heart} \sqcap \exists \mathsf{treatment}.\mathsf{Surgery}$ |

| Visible Knowledge Base $\mathcal{K}_v$ | |
|---|---|
| $\delta_1$ | $VSD\_Patient \equiv Patient \sqcap \exists hasOrg.\mathbf{VSD\_Heart}$ |
| $\delta_2$ | $HS\_Patient \equiv Patient \sqcap \exists hasOrg.\mathbf{Sur\_Heart}$ |
| $\delta_3$ | $AS\_Patient \equiv Patient \sqcap \exists hasOrg.(\mathbf{Heart} \sqcap \exists \mathbf{con}.\mathbf{AS})$ |
| $\delta_4$ | $Ab\_TV \sqsubseteq \mathbf{TV}$ |
| $\delta_5$ | $Dis\_TV \sqsubseteq Ab\_TV$ |
| $\delta_6$ | $EA\_Heart \equiv \mathbf{VSD\_Heart} \sqcap \exists \mathbf{part}.Dis\_TV$ |
| $\delta_7$ | $EA\_Patient \equiv Patient \sqcap \exists hasOrg.EA\_Heart$ |
| $\delta_8$ | $Ab\_TV\_Heart \equiv \mathbf{Heart} \sqcap \exists \mathbf{part}.Ab\_TV$ |
| $\delta_9$ | $TVD\_Patient \equiv Patient \sqcap \exists hasOrg.Ab\_TV\_Heart$ |

To support such scenarios, we propose the *import-by-query* framework. Instead of publishing (a subset of) the axioms of $\mathcal{T}_h$, MRC can publish an *oracle* for $\mathcal{T}_h$—a service that advertises a set $\Gamma$ of public symbols in $\mathcal{T}_h$ and a query language $\mathcal{L}$, and that can answer $\mathcal{L}$-queries over $\mathcal{T}_h$ provided that these queries use only symbols in $\Gamma$. A so-called *import-by-query algorithm* can then reason with $\mathcal{K}_v \cup \mathcal{T}_h$ (e.g., determine the satisfiability of $\mathcal{K}_v \cup \mathcal{T}_h$) without having physical access to the contents of $\mathcal{T}_h$, by just asking queries to the oracle. The existence of such an algorithm, however, depends on the oracle's query language, the DLs used to express $\mathcal{K}_v$ and $\mathcal{T}_h$, and the way in which the symbols from $\Gamma$ are reused in $\mathcal{K}_v$.

One of the most popular query languages in description logics is concept satisfiability, which is available in all DL reasoners known to us. It is thus natural to consider *concept satisfiability oracles*, which advertise a signature $\Gamma$ and check the satisfiability w.r.t. $\mathcal{T}_h$ of (not necessarily atomic) concepts formed using the symbols in $\Gamma$. Later on we show that import-by-query algorithms based on concept satisfiability oracles exist only if rather strong restrictions are imposed on the way $\mathcal{K}_v$ reuses the symbols from $\Gamma$; roughly speaking, it is not possible to mix roles from $\Gamma$ with concepts private to $\mathcal{K}_v$ in existential and universal restrictions. In our example, this means that axioms $\delta_6$ and $\delta_8$ from Table 3 would not be allowed in $\mathcal{K}_v$. To overcome the limitations of concept satisfiability oracles, we consider two additional types of (closely related) oracles that are more powerful than the oracles based on concept satisfiability. An *ABox satisfiability oracle* is given an ABox $\mathcal{A}$ with $\mathsf{sig}(\mathcal{A}) \subseteq \Gamma$, and it checks the satisfiability of $\mathcal{A} \cup \mathcal{T}_h$. An *ABox entailment oracle* is given an ABox $\mathcal{A}$ and an assertion $\alpha$ with $\mathsf{sig}(\mathcal{A}) \subseteq \Gamma$ and $\mathsf{sig}(\alpha) \subseteq \Gamma$, and it checks whether $\mathcal{A} \cup \mathcal{T}_h \models \alpha$. ABox satisfiability and entailment have been implemented in most state-of-the-art DL reasoners, so oracles based on such inferences seem natural.

In practice, it is natural to express oracle queries in the same DL as $\mathcal{T}_h$; however, for the sake of generality we allow queries to be expressed in an arbitrary description logic $\mathcal{L}$. Intuitively, this allows $\mathcal{K}_v$ to "learn more about the structure of the models of $\mathcal{T}_h$," which allows us to obtain more general results about nonexistence of import-by-query algorithms. Definition 3 formally introduces different types of oracles.

**Definition 3.** *Let $\mathcal{T}_h$ be a TBox, let $\Gamma$ be a signature, and let $\mathcal{L}$ be a description logic.*

*The* concept satisfiability oracle *for $\mathcal{T}_h$, $\Gamma$, and $\mathcal{L}$ is the Boolean function $\Omega^{\mathsf{c}}_{\mathcal{T}_h, \Gamma, \mathcal{L}}$ that, for each $\mathcal{L}$-concept $C$ with $\mathsf{sig}(C) \subseteq \Gamma$, returns $\mathsf{t}$ if and only if $C$ is satisfiable w.r.t. $\mathcal{T}_h$.*

*The* ABox satisfiability oracle *for $\mathcal{T}_h$, $\Gamma$, and $\mathcal{L}$ is the Boolean function $\Omega^{\mathsf{a}}_{\mathcal{T}_h, \Gamma, \mathcal{L}}$ that, for each connected $\mathcal{L}$-ABox $\mathcal{A}$ with $\mathsf{sig}(\mathcal{A}) \subseteq \Gamma$, returns $\mathsf{t}$ if and only if $\mathcal{T}_h \cup \mathcal{A}$ is satisfiable.*

*The* ABox entailment oracle *for $\mathcal{T}_h$, $\Gamma$, and $\mathcal{L}$ is the Boolean function $\Omega^{\mathsf{e}}_{\mathcal{T}_h, \Gamma, \mathcal{L}}$ that, for each connected $\mathcal{L}$-ABox $\mathcal{A}$ such that $\mathsf{sig}(\mathcal{A}) \subseteq \Gamma$ and each $\mathcal{L}$-assertion $\alpha$ that mentions only the individuals in $\mathcal{A}$ such that $\mathsf{sig}(\alpha) \subseteq \Gamma$, returns $\mathsf{t}$ if and only if $\mathcal{T}_h \cup \mathcal{A} \models \alpha$.*

We use the generic term *oracle* for either a concept satisfiability, an ABox satisfiability, or an ABox entailment oracle. Furthermore, if $\mathcal{L}$ is the same as the description logic of $\mathcal{T}_h$, we abbreviate $\Omega_{\mathcal{T}_h, \Gamma, \mathcal{L}}$ to $\Omega_{\mathcal{T}_h, \Gamma}$. Finally, we often refer to the oracle arguments (i.e., the concepts $C$, the ABoxes $\mathcal{A}$, and the pairs $\langle \mathcal{A}, \alpha \rangle$ in the case of concept satisfiability, ABox satisfiability, and ABox entailment oracles, respectively) as *oracle queries*.

We next formally define import-by-query algorithms using the well-known notion of an oracle Turing machine. A precise definition of the latter is given by Papadimitriou (1993); we next present just an informal overview of the main ideas. An oracle Turing machine $T$ has a separate *query* tape, on which it can write arbitrary strings over a given alphabet. At any point in time, $T$ can enter a special state $q_?$, upon which a black-box *oracle* $\Omega$ checks whether the string currently written on the query tape belongs to the language associated with $\Omega$; if that is the case, then $T$ enters a special state $q_{yes}$, and otherwise $T$ enters a special state $q_{no}$. This allows the oracle's answers to affect the computation of $T$. A combination of $T$ and $\Omega$ is usually written as $T^{\Omega}$. This definition assumes that the computation of $T$ depends only on the input and the oracle's answers; that is, if $\Omega_1$ and $\Omega_2$ are two distinct oracles, the computations of $T^{\Omega_1}$ will be indistinguishable from the computations of $T^{\Omega_2}$ if $\Omega_1$ and $\Omega_2$ return the same answers to queries encountered in computations. In the rest of this paper, we do not make any assumptions on the type of $T$: any "reasonable" Turing machine model can be used. We merely assume that $T$ is equipped with a suitable notion of a *run* which captures the computation of $T^{\Omega}$ on each input. A run can (but does not need to) accept or reject the input.

**Definition 4.** *A class of inputs $\mathcal{C}$ is a class of triples of the form $\langle \Gamma^{\mathcal{C}}, \mathcal{K}_v^{\mathcal{C}}, \mathcal{T}_h^{\mathcal{C}} \rangle$ where $\Gamma^{\mathcal{C}}$ is a signature, $\mathcal{K}_v^{\mathcal{C}}$ is a knowledge base, and $\mathcal{T}_h^{\mathcal{C}}$ is a TBox such that $\mathsf{sig}(\mathcal{K}_v^{\mathcal{C}}) \cap \mathsf{sig}(\mathcal{T}_h^{\mathcal{C}}) \subseteq \Gamma^{\mathcal{C}}$. Each triple in $\mathcal{C}$ is called an* input.

*An* import-by-query algorithm *for a description logic $\mathcal{L}$ and a class of inputs $\mathcal{C}$ based on oracles of type $x \in \{\mathsf{a}, \mathsf{e}, \mathsf{c}\}$ is an oracle Turing machine $\mathsf{ibq}^x$ that can be combined with an oracle of type $x$. For each input $\langle \Gamma, \mathcal{K}_v, \mathcal{T}_h \rangle \in \mathcal{C}$ the following properties must be satisfied, where $\mathsf{ibq}^x[\mathcal{T}_h, \Gamma, \mathcal{L}]$ is the combination of $\mathsf{ibq}^x$ and the oracle $\Omega_{\mathcal{T}_h, \Gamma, \mathcal{L}}^x$:*

1. *whenever $\mathsf{ibq}^x[\mathcal{T}_h, \Gamma, \mathcal{L}]$ enters the state $q_?$ in a run, the string on the query tape encodes a query accepted by $\Omega_{\mathcal{T}_h, \Gamma, \mathcal{L}}^x$;*

2. *$\mathsf{ibq}^x[\mathcal{T}_h, \Gamma, \mathcal{L}]$ has an accepting run on $\mathcal{K}_v$ if and only if $\mathcal{K}_v \cup \mathcal{T}_h$ is satisfiable; and*

3. *each run of $\mathsf{ibq}^x[\mathcal{T}_h, \Gamma, \mathcal{L}]$ on $\mathcal{K}_v$ is finite.*

Intuitively, the transition relation of $\mathsf{ibq}^x$ takes into account the possible answers of an oracle of type $x$, but $\mathsf{ibq}^x$ is not "executable" because the actual oracle is unknown. Thus, $\mathsf{ibq}^x$ can be seen as a computer program in which a particular subroutine is missing. Given an input $\langle \Gamma, \mathcal{K}_v, \mathcal{T}_h \rangle \in \mathcal{C}$, we can parameterize $\mathsf{ibq}^x$ by $\Omega_{\mathcal{T}_h, \Gamma, \mathcal{L}}^x$ to obtain $\mathsf{ibq}^x[\mathcal{T}_h, \Gamma, \mathcal{L}]$, and the latter Turing machine can be freely applied to $\mathcal{K}_v$.

In the rest of this paper, whenever the oracle type is not explicitly given, our discussion applies to all oracle types. We will consider various classes of inputs, each of which can be defined using the following formulation:

> $\mathcal{C}$ is the largest class of triples $\langle \Gamma^{\mathcal{C}}, \mathcal{K}_v^{\mathcal{C}}, \mathcal{T}_h^{\mathcal{C}} \rangle$ where $\mathsf{sig}(\mathcal{K}_v^{\mathcal{C}}) \cap \mathsf{sig}(\mathcal{T}_h^{\mathcal{C}}) \subseteq \Gamma^{\mathcal{C}}$ and $\Gamma^{\mathcal{C}}$, $\mathcal{K}_v^{\mathcal{C}}$, and $\mathcal{T}_h^{\mathcal{C}}$ satisfy some condition.

Usually, however, we abbreviate such formulations as follows:

> $\mathcal{C}[\Gamma^{\mathcal{C}}, \mathcal{K}_v^{\mathcal{C}}, \mathcal{T}_h^{\mathcal{C}}]$ is a class of inputs where $\Gamma^{\mathcal{C}}$, $\mathcal{K}_v^{\mathcal{C}}$, and $\mathcal{T}_h^{\mathcal{C}}$ satisfy some condition.

Definition 4 straightforwardly implies the following property, which essentially just reformulates the idea that the runs of a Turing machine are determined only by the oracles' answers, and not the oracles themselves.

**Proposition 1.** *Let $\mathsf{ibq}$ be an import-by-query algorithm for a description logic $\mathcal{L}$ and a class of inputs $\mathcal{C}$, let $\langle \Gamma, \mathcal{K}_v, \mathcal{T}_h^1 \rangle$ be an arbitrary input from $\mathcal{C}$, and let $Q_1, \ldots, Q_n$ be the oracle queries encountered in all possible runs of $\mathsf{ibq}[\mathcal{T}_h^1, \Gamma, \mathcal{L}]$ on $\mathcal{K}_v$. Then, for each $\mathcal{T}_h^2$ such that $\langle \Gamma, \mathcal{K}_v, \mathcal{T}_h^2 \rangle \in \mathcal{C}$ and $\Omega_{\mathcal{T}_h^1, \Gamma, \mathcal{L}}(Q_i) = \Omega_{\mathcal{T}_h^2, \Gamma, \mathcal{L}}(Q_i)$ for each $1 \leq i \leq n$, each run of $\mathsf{ibq}[\mathcal{T}_h^1, \Gamma, \mathcal{L}]$ on $\mathcal{K}_v$ is a run of $\mathsf{ibq}[\mathcal{T}_h^2, \Gamma, \mathcal{L}]$ on $\mathcal{K}_v$ and vice versa.*

In Section 4 we will identify DLs defining the oracle query language and classes of inputs for which no import-by-query algorithm based on oracles of a particular type exists. The following proposition shows that it suffices to prove nonexistence results for the most expressive DL and the smallest class of inputs; then, analogous results then hold for each weaker DL and each larger class of inputs.

**Proposition 2.** *Let $\mathcal{L}_1$ be a description logic and let $\mathcal{L}_2$ be a fragment of $\mathcal{L}_1$; let $\mathcal{C}_1$ and $\mathcal{C}_2$ be classes of inputs such that each triple in $\mathcal{C}_1$ also belongs to $\mathcal{C}_2$; and let $x \in \{\mathsf{a}, \mathsf{c}, \mathsf{e}\}$ be an oracle type. If there is no import-by-query algorithm for $\mathcal{L}_1$ and $\mathcal{C}_1$ based on oracles of type $x$, then there is also no import-by-query algorithm for $\mathcal{L}_2$ and $\mathcal{C}_2$ based on oracles of type $x$.*

*Proof.* We prove the contrapositive claim. Let $\mathsf{ibq}^x$ be an import-by-query algorithm for $\mathcal{L}_2$ and $\mathcal{C}_2$. Since each triple in $\mathcal{C}_1$ is also contained in $\mathcal{C}_2$, $\mathsf{ibq}^x$ is clearly an import-by-query algorithm for $\mathcal{L}_2$ and $\mathcal{C}_1$. Let $\langle \Gamma, \mathcal{K}_v, \mathcal{T}_h \rangle \in \mathcal{C}_1$ be an arbitrary input, and let $Q$ be an arbitrary $\mathcal{L}_2$-query encountered in a run of $\mathsf{ibq}^x[\mathcal{T}_h, \Gamma, \mathcal{L}]$ on $\mathcal{K}_v$. Since $\mathcal{L}_2$ is a fragment of $\mathcal{L}_1$, $Q$ is an $\mathcal{L}_1$-query as well. Thus, $\mathsf{ibq}^x$ is an import-by-query algorithm for $\mathcal{L}_1$ and $\mathcal{C}_1$. □

The following theorem shows that oracles of certain types can simulate oracles of other types. This is important because if $\Omega_1$ can simulate $\Omega_2$ and we show that no import by query algorithm exists for a particular class of inputs applicable to $\Omega_1$, then also no such algorithm exists that is applicable to $\Omega_2$.

**Theorem 1.** *Let $\leq$ be the smallest partial order on the class of all oracles that satisfies the following conditions for each TBox $\mathcal{T}_h$, each signature $\Gamma$, and each description logic $\mathcal{L}$:*

1. *$\Omega^{\mathsf{c}}_{\mathcal{T}_h, \Gamma, \mathcal{L}} \leq \Omega^{\mathsf{a}}_{\mathcal{T}_h, \Gamma, \mathcal{L}} \leq \Omega^{\mathsf{e}}_{\mathcal{T}_h, \Gamma, \mathcal{L}}$; and*

2. *if for each $\mathcal{L}$-ABox $\mathcal{A}$ and each $\mathcal{L}$-assertion $\alpha$ we have that $\mathcal{A} \cup \{\neg\alpha\}$ is an $\mathcal{L}$-ABox, then $\Omega^{\mathsf{e}}_{\mathcal{T}_h, \Gamma, \mathcal{L}} \leq \Omega^{\mathsf{a}}_{\mathcal{T}_h, \Gamma, \mathcal{L}}$ holds as well.*

*Let $\mathcal{L}$ be a description logic, let $\mathcal{C}$ be a class of inputs, and let $x_1, x_2 \in \{\mathsf{a}, \mathsf{c}, \mathsf{e}\}$ be oracle types such that $\Omega^{x_1}_{\mathcal{T}_h, \Gamma, \mathcal{L}} \leq \Omega^{x_2}_{\mathcal{T}_h, \Gamma, \mathcal{L}}$ for each $\langle \Gamma, \mathcal{K}_v, \mathcal{T}_h \rangle \in \mathcal{C}$. Then, each import-by-query algorithm $\mathsf{ibq}^{x_1}$ for $\mathcal{L}$ and $\mathcal{C}$ can be transformed into an import-by-query algorithm $\mathsf{ibq}^{x_2}$ for $\mathcal{L}$ and $\mathcal{C}$ such that, for each input $\langle \Gamma, \mathcal{K}_v, \mathcal{T}_h \rangle \in \mathcal{C}$, $\mathsf{ibq}^{x_1}[\mathcal{T}_h, \Gamma, \mathcal{L}]$ has a run on $\mathcal{K}_v$ with $n$ oracle queries if and only if $\mathsf{ibq}^{x_2}[\mathcal{T}_h, \Gamma, \mathcal{L}]$ has a run on $\mathcal{K}_v$ with $n$ oracle queries.*

*Proof.* Let $\mathsf{ibq}^{x_1}$ be an arbitrary import-by-query algorithm for $\mathcal{L}$ and $\mathcal{C}$, and consider an arbitrary input $\langle \Gamma, \mathcal{K}_v, \mathcal{T}_h \rangle \in \mathcal{C}$. Conditions 1 and 2 ensure that $\Omega^{x_1}_{\mathcal{T}_h, \Gamma, \mathcal{L}}$ is *reducible* to $\Omega^{x_2}_{\mathcal{T}_h, \Gamma, \mathcal{L}}$ in the sense that a computable total function $f$ exists from the domain of $\Omega^{x_1}_{\mathcal{T}_h, \Gamma, \mathcal{L}}$ to the domain of $\Omega^{x_2}_{\mathcal{T}_h, \Gamma, \mathcal{L}}$ such that for each query $Q$ accepted by $\Omega^{x_1}_{\mathcal{T}_h, \Gamma, \mathcal{L}}$, we have $\Omega^{x_1}_{\mathcal{T}_h, \Gamma, \mathcal{L}}(Q) = \Omega^{x_2}_{\mathcal{T}_h, \Gamma, \mathcal{L}}(f(Q))$. In particular, an ABox satisfiability oracle is reducible to an ABox entailment oracle via $f(\mathcal{A}) = (\mathcal{A}, \perp)$ for each ABox $\mathcal{A}$. Furthermore, if Condition 2 holds, then an ABox entailment oracle is reducible to an ABox satisfiability oracle via $f(\mathcal{A}, \alpha) = \mathcal{A} \cup \{\neg\alpha\}$. Finally, a concept satisfiability oracle is reducible to an ABox satisfiability oracle via $f(C) = \{C(a)\}$ for $a$ a fresh individual.

Algorithm $\mathsf{ibq}^{x_2}$ can then simply simulate $\mathsf{ibq}^{x_1}$ on each input $\langle \Gamma, \mathcal{K}_v, \mathcal{T}_h \rangle \in \mathcal{C}$; furthermore, whenever $\mathsf{ibq}^{x_1}[\mathcal{T}_h, \Gamma, \mathcal{L}]$ poses a query $Q$ to $\Omega^{x_1}_{\mathcal{T}_h, \Gamma, \mathcal{L}}$, then $\mathsf{ibq}^{x_2}[\mathcal{T}_h, \Gamma, \mathcal{L}]$ computes $f(Q)$ and poses the query $f(Q)$ to $\Omega^{x_2}_{\mathcal{T}_h, \Gamma, \mathcal{L}}$. Since $\mathsf{ibq}^{x_1}$ is an import-by-query algorithm for $\mathcal{L}$ and $\mathcal{C}$, so is $\mathsf{ibq}^{x_2}$. Furthermore, for each input, there is a one-to-one correspondence between the runs of both algorithms with corresponding runs posing exactly the same number of oracle queries. □

We next show that, if the shared signature $\Gamma$ contains only atomic concepts, there is a close correspondence between ABox and concept satisfiability oracles.

**Theorem 2.** *Let $\mathcal{L}$ be a description logic and let $\mathcal{C}[\Gamma^{\mathcal{C}}, \mathcal{K}^{\mathcal{C}}_v, \mathcal{T}^{\mathcal{C}}_h]$ be a class of inputs where $\Gamma^{\mathcal{C}}$ contains only atomic concepts. Then, each import-by-query algorithm $\mathsf{ibq}^{\mathsf{a}}$ for $\mathcal{L}$ and $\mathcal{C}$ can be transformed into an import-by-query algorithm $\mathsf{ibq}^{\mathsf{c}}$ for $\mathcal{L}$ and $\mathcal{C}$ such that the following statements hold for each input $\langle \Gamma, \mathcal{K}_v, \mathcal{T}_h \rangle \in \mathcal{C}$.*

- *For each run of $\mathsf{ibq^a}[\mathcal{T}_h, \Gamma, \mathcal{L}]$ on $\mathcal{K}_v$ with $n$ oracle queries and $m$ the maximum number of individuals in a query ABox, a run of $\mathsf{ibq^c}[\mathcal{T}_h, \Gamma, \mathcal{L}]$ on $\mathcal{K}_v$ with at most $n \times m$ oracle queries exists.*

- *For each run of $\mathsf{ibq^c}[\mathcal{T}_h, \Gamma, \mathcal{L}]$ on $\mathcal{K}_v$ with $n$ oracle queries, a run of $\mathsf{ibq^a}[\mathcal{T}_h, \Gamma, \mathcal{L}]$ on $\mathcal{K}_v$ with at most $n$ oracle queries exists.*

*Proof.* Let $\mathsf{ibq^a}$ be an import-by-query algorithm for $\mathcal{L}$ and $\mathcal{C}$. We define $\mathsf{ibq^c}$ such that, on each input $\langle \Gamma, \mathcal{K}_v, \mathcal{T}_h \rangle \in \mathcal{C}$, algorithm $\mathsf{ibq^c}[\mathcal{T}_h, \Gamma, \mathcal{L}]$ simulates the steps of algorithm $\mathsf{ibq^a}[\mathcal{T}_h, \Gamma, \mathcal{L}]$; furthermore, when $\mathsf{ibq^a}[\mathcal{T}_h, \Gamma, \mathcal{L}]$ queries $\Omega^{\mathsf{a}}_{\mathcal{T}_h, \Gamma, \mathcal{L}}$ with an ABox $\mathcal{A}$, algorithm $\mathsf{ibq^c}[\mathcal{T}_h, \Gamma, \mathcal{L}]$ proceeds as follows.

1. The algorithm transforms $\mathcal{A}$ into an ABox $\mathcal{A}'$ by iterating over all assertions of the form $a \approx b$ in $\mathcal{A}$ and, for each such assertion, replacing one individual (say $a$) with the other one (say $b$) in all assertions.

2. If $\mathcal{A}'$ contains an individual $a$ such that $a \not\approx a \in \mathcal{A}'$ or $\Omega^{\mathsf{c}}_{\mathcal{T}_h, \Gamma, \mathcal{L}}(B_1 \sqcap \ldots \sqcap B_n) = \mathsf{f}$ where $B_1, \ldots, B_n$ are all concepts such that $B_i(a) \in \mathcal{A}'$, then $\mathsf{ibq^c}[\mathcal{T}_h, \Gamma, \mathcal{L}]$ proceeds in the same way as $\mathsf{ibq^a}[\mathcal{T}_h, \Gamma, \mathcal{L}]$ for $\Omega^{\mathsf{a}}_{\mathcal{T}_h, \Gamma, \mathcal{L}}(\mathcal{A}) = \mathsf{f}$; otherwise, $\mathsf{ibq^c}[\mathcal{T}_h, \Gamma, \mathcal{L}]$ proceeds in the same way as $\mathsf{ibq^a}[\mathcal{T}_h, \Gamma, \mathcal{L}]$ for $\Omega^{\mathsf{a}}_{\mathcal{T}_h, \Gamma, \mathcal{L}}(\mathcal{A}) = \mathsf{t}$.

There is an obvious correspondence between the runs of $\mathsf{ibq^a}[\mathcal{T}_h, \Gamma, \mathcal{L}]$ and $\mathsf{ibq^c}[\mathcal{T}_h, \Gamma, \mathcal{L}]$ on $\mathcal{K}_v$; furthermore, whenever $\mathsf{ibq^a}[\mathcal{T}_h, \Gamma, \mathcal{L}]$ issues a query to $\Omega^{\mathsf{a}}_{\mathcal{T}_h, \Gamma, \mathcal{L}}$, then $\mathsf{ibq^c}[\mathcal{T}_h, \Gamma, \mathcal{L}]$ issues at most $m$ queries to $\Omega^{\mathsf{c}}_{\mathcal{T}_h, \Gamma, \mathcal{L}}$ in order to determine how to proceed. Finally, note that the second statement in the theorem directly follows from Theorem 1. $\qquad\square$

We finally show that we can without loss of generality assume $\mathcal{K}_v$ to contain no concept such as $\exists \mathbf{con}.\mathbf{AS}$ in axiom $\delta_3$ in Table 3.

**Definition 5.** *Let $\Gamma$ be a signature. A concept $C$ is $\Gamma$-modal if $\mathsf{sig}(C) \subseteq \Gamma$ and $C$ is of the form $\exists R.D$, $\forall R.D$, $\geq n\, R.D$, or $\leq n\, R.D$.*

Intuitively, $\Gamma$-modal concepts can always be treated as "atomic" from the point of view of $\mathcal{K}_v$, so we can rely on the oracle to compute all relevant consequences of such concepts.

**Theorem 3.** *Let $\mathcal{L}$, $\mathcal{DL}_1$, and $\mathcal{DL}_2$ be description logics such that each $\mathcal{DL}_1$-concept is also an $\mathcal{L}$-concept and $\mathcal{DL}_2$ allows for $\mathcal{DL}_1$-definitions; let $x \in \{\mathsf{a}, \mathsf{c}, \mathsf{e}\}$; let $\mathcal{C}[\Gamma^{\mathcal{C}}, \mathcal{K}_v^{\mathcal{C}}, \mathcal{T}_h^{\mathcal{C}}]$ be a class of inputs where $\mathcal{K}_v^{\mathcal{C}}$ is a $\mathcal{DL}_1$-knowledge base and $\mathcal{T}_h^{\mathcal{C}}$ is a $\mathcal{DL}_2$-TBox; and let $\mathcal{D}[\Gamma^{\mathcal{D}}, \mathcal{K}_v^{\mathcal{D}}, \mathcal{T}_h^{\mathcal{D}}]$ be the class of inputs consisting of all triples $\langle \Gamma, \mathcal{K}_v, \mathcal{T}_h \rangle$ in $\mathcal{C}[\Gamma^{\mathcal{C}}, \mathcal{K}_v^{\mathcal{C}}, \mathcal{T}_h^{\mathcal{C}}]$ in which $\mathcal{K}_v$ contains no $\Gamma$-modal concepts. Then, each import-by-query algorithm $\mathsf{ibq}_2^x$ for $\mathcal{L}$ and $\mathcal{D}$ can be transformed into an import-by-query algorithm $\mathsf{ibq}_1^x$ for $\mathcal{L}$ and $\mathcal{C}$.*

*Proof.* For $\Gamma$ a signature, $C$ a concept, and $\alpha$ a concept, axiom, or knowledge base, we say that $C$ is $\Gamma$-*outermost* in $\alpha$ if $C$ is $\Gamma$-modal and $C$ does not occur in $\alpha$ as a proper subconcept of another $\Gamma$-modal concept.

Let $\langle \Gamma, \mathcal{K}_v, \mathcal{T}_h \rangle \in \mathcal{C}$ be an arbitrary input in $\mathcal{C}$, let $S$ be the set of all $\Gamma$-outermost concepts in $\mathcal{K}_v$, and let $X_C$ be a fresh atomic concept uniquely associated with each $C \in S$. We define $\Gamma'$, $\mathcal{T}_h'$, and $\mathcal{K}_v'$ as follows: $\Gamma' = \Gamma \cup \{X_C \mid C \in S\}$; $\mathcal{K}_v'$ is obtained from $\mathcal{K}_v$ by replacing each $C \in S$ with $X_C$; and $\mathcal{T}_h' = \mathcal{T}_h \cup \{X_C \equiv C \mid C \in S\}$. Clearly, $\mathcal{K}_v \cup \mathcal{T}_h$ is equisatisfiable with $\mathcal{K}'_v \cup \mathcal{T}'_h$, and $\langle \Gamma', \mathcal{K}'_v, \mathcal{T}'_h \rangle \in \mathcal{D}$. Let $\mathsf{ibq}^x_2$ be an arbitrary import-by-query algorithm for $\mathcal{L}$ and $\mathcal{D}$. We define $\mathsf{ibq}^x_1$ as the algorithm that on each $\langle \Gamma, \mathcal{K}_v, \mathcal{T}_h \rangle \in \mathcal{C}$ simulates the steps of $\mathsf{ibq}^x_2$ on input $\langle \Gamma', \mathcal{K}'_v, \mathcal{T}'_h \rangle \in \mathcal{D}$, but with the following modifications:

- $\mathsf{ibq}^x_1[\mathcal{T}_h, \Gamma, \mathcal{L}]$ treats all concepts in $S$ as if they were atomic; and

- whenever $\mathsf{ibq}^x_2[\mathcal{T}'_h, \Gamma', \mathcal{L}]$ queries $\Omega^x_{\mathcal{T}'_h, \Gamma', \mathcal{L}}$ with a query $Q'$, then $\mathsf{ibq}^x_1[\mathcal{T}_h, \Gamma, \mathcal{L}]$ queries $\Omega^x_{\mathcal{T}_h, \Gamma, \mathcal{L}}$ with a query $Q$ obtained from $Q'$ by replacing each occurrence of $X_C$ with $C$.

There is an obvious correspondence between the runs of $\mathsf{ibq}^x_2[\mathcal{T}'_h, \Gamma', \mathcal{L}]$ and $\mathsf{ibq}^x_1[\mathcal{T}_h, \Gamma, \mathcal{L}]$ on $\mathcal{K}_v$, so $\mathsf{ibq}^x_1$ is an import-by-query algorithm for $\mathcal{L}$ and $\mathcal{C}$. $\qquad \square$

## 4. Limitations of the Import-by-Query Framework

In this section, we explore the limitations of the import-by-query framework and show that import-by-query algorithms do not exist under certain conditions. Our negative results apply to classes of input where $\mathcal{K}_v$ and $\mathcal{T}_h$ are expressed in a description logic $\mathcal{DL}$ that is as lightweight as possible, the oracle is based on ABox satisfiability, and the oracle accepts queries expressed in a description logic $\mathcal{L}$ that is as expressive as possible. By Theorem 1 and Proposition 2, our results also apply to all other oracle types, queries expressed in a fragment of $\mathcal{L}$, and all classes of input where $\mathcal{K}_v$ and $\mathcal{T}_h$ are expressed in a description logic that extends $\mathcal{DL}$.

In particular, in Section 4.1 we establish the following general limitations of the import-by-query framework.

- The presence of nominals in $\mathcal{T}_h$ may preclude the existence of an import-by-query algorithm even if $\Gamma = \emptyset$ (cf. Theorem 4).

- Deductive modularity of the TBox of $\mathcal{K}_v$ w.r.t. $\Gamma$ is a *necessary* condition for the existence of an import-by-query algorithm (cf. Theorem 5).

- Deductive modularity, however, is not *sufficient*, even if $\mathcal{K}_v$ and $\mathcal{T}_h$ are in $\mathcal{EL}$ and $\Gamma$ is allowed to contain only atomic concepts (cf. Theorem 6).

In response to these negative results, all import-by-query algorithms proposed in this paper are subjected to the following restrictions:

R1. $\mathcal{T}_h$ is not allowed to contain nominals.

R2. The TBox of $\mathcal{K}_v$ is required to be *semantically* modular w.r.t. $\Gamma$.

We show in Section 5.1 that these two restrictions are *sufficient* to guarantee the existence of an import-by-query algorithm for $\mathcal{K}_v$ in $\mathcal{ALCHIQ}$ and $\mathcal{T}_h$ in $\mathcal{ALCHIQ}$, provided that $\Gamma$ contains only atomic concepts.

In Section 4.2, however, we show that further restrictions on the input are necessary if $\Gamma$ is allowed to contain atomic roles. Roughly speaking, restrictions R1 and R2 are insufficient since the axioms in $\mathcal{K}_v$ can arbitrarily propagate information about the symbols private to $\mathcal{K}_v$ via a role in $\Gamma$ to a "hidden" part of the canonical model of $\mathcal{K}_v \cup \mathcal{T}_h$ (that is, a part of the canonical model that cannot be constructed using only the axioms in $\mathcal{K}_v$); such propagation can occur both via existential (cf. Theorem 7) and universal quantification (cf. Theorem 8). To overcome these negative results, we define in Section 5.1 the *HT-safety* condition that, on the one hand, ensures semantic modularity and, on the other hand, prevents arbitrary transfer of information about the symbols private to $\mathcal{K}_v$ to hidden parts of the canonical model via a role in $\Gamma$. This condition, however, is still insufficient to enable import-by-query reasoning if $\mathcal{T}_h$ contains universal quantifiers, inverse roles, and functional roles, and $\mathcal{K}_v$ entails cyclic axioms of the form $A \sqsubseteq \exists R.A$ for $R \in \Gamma$ and $A \notin \Gamma$ (cf. Theorem 9). To overcome this negative result, in Section 5.1 we introduce an *acyclicity* condition that together with HT-safety guarantees the existence of an import-by-query algorithm based on ABox satisfiability oracles for $\mathcal{K}_v$ and $\mathcal{T}_h$ expressed in $\mathcal{ALCHIQ}$.

Finally, in Section 4.3 we show that no import-by-query algorithm based on concept satisfiability oracles exists for the class of inputs $\mathcal{C}[\Gamma^{\mathcal{C}}, \mathcal{K}_v^{\mathcal{C}}, \mathcal{T}_h^{\mathcal{C}}]$ where $\mathcal{K}_v^{\mathcal{C}}$ is in $\mathcal{EL}$ and it satisfies the HT-safety condition, and $\mathcal{T}_h^{\mathcal{C}}$ is in $\mathcal{EL}$ (cf. Theorem 10). In Section 5.2.2, however, we present an algorithm based on ABox entailment oracles that applies to this class of inputs $\mathcal{C}$. Thus, practically relevant cases exist for which import-by-query reasoning is impossible with concept satisfiability oracles, but it becomes feasible with ABox oracles.

## 4.1 General Limitations

We first show that the presence of nominals in the hidden knowledge base precludes the existence of an import-by-query algorithm if the visible knowledge base is satisfiable only in infinite models. Expressive DLs used in practice often do not have the finite model property, and our negative result holds even if the shared signature is empty; thus, in the rest of this paper we do not further consider DLs with nominals, and we leave an investigation of conditions that enable import-by-query reasoning with such DLs for future work.

**Theorem 4.** *For each description logic $\mathcal{DL}$ without the finite model property, no import-by-query algorithm based on ABox satisfiability oracles exists for $\mathcal{L} = \mathcal{ALCHOIQ}$ and the class of inputs $\mathcal{C}[\Gamma^{\mathcal{C}}, \mathcal{K}_v^{\mathcal{C}}, \mathcal{T}_h^{\mathcal{C}}]$ where $\Gamma^{\mathcal{C}} = \emptyset$, $\mathcal{K}_v^{\mathcal{C}}$ is a $\mathcal{DL}$-knowledge base, and $\mathcal{T}_h^{\mathcal{C}}$ is an $\mathcal{ALCHOIQ}$-TBox.*

*Proof.* Let $\mathcal{C}$ be an arbitrary class of inputs and let $\mathsf{ibq^a}$ be an arbitrary import-by-query algorithm such that $\mathcal{C}$ and $\mathsf{ibq^a}$ both satisfy the theorem's assumptions. Furthermore, let $\langle \Gamma, \mathcal{K}_v, \mathcal{T}_h^1 \rangle \in \mathcal{C}$ be an arbitrary input where $\mathcal{K}_v^{\mathcal{C}}$ is satisfiable only in infinite models, $\Gamma = \emptyset$, and $\mathcal{T}_h^1 = \emptyset$. Since all runs of $\mathsf{ibq^a}[\mathcal{T}_h^1, \Gamma, \mathcal{L}]$ on $\mathcal{K}_v$ are finite, the number of individuals occurring in a query ABox in each such run is bounded by some integer $n$. Let $\mathcal{T}_h^2$ be as follows, where $O_1, \ldots, O_n$ are fresh nominal concepts:

$$\mathcal{T}_h^2 = \{\top \sqsubseteq O_1 \sqcup \ldots \sqcup O_n\} \tag{9}$$

Clearly, $\mathcal{K}_v \cup \mathcal{T}_h^1$ is satisfiable, but $\mathcal{K}_v \cup \mathcal{T}_h^2$ is not. Consider now an arbitrary query ABox $\mathcal{A}$ occurring in a run of $\mathsf{ibq^a}[\mathcal{T}_h^1, \Gamma, \mathcal{L}]$. Since $\Gamma = \emptyset$, $\mathcal{A}$ consists only of assertions of the form $a \approx b$ or $a \not\approx b$; furthermore, $\mathcal{A}$ contains at most $n$ individuals, so $\Omega_{\mathcal{T}_h^1, \Gamma}^{\mathsf{a}}(\mathcal{A}) = \mathsf{t}$ implies $\Omega_{\mathcal{T}_h^2, \Gamma}^{\mathsf{a}}(\mathcal{A}) = \mathsf{t}$, and the converse holds by the monotonicity of first-order logic. But then, by Proposition 1, the runs of $\mathsf{ibq^a}[\mathcal{T}_h^1, \Gamma, \mathcal{L}]$ on $\mathcal{K}_v$ coincide with the runs of $\mathsf{ibq^a}[\mathcal{T}_h^2, \Gamma, \mathcal{L}]$ on $\mathcal{K}_v$, which contradicts the fact that $\mathcal{K}_v \cup \mathcal{T}_h^1$ is satisfiable but $\mathcal{K}_v \cup \mathcal{T}_h^2$ is not. □

We next present a very strong result: deductive modularity is a *necessary* requirement for the existence of an import-by-query algorithm; that is, no import-by-query algorithm exists for any class of inputs that contains a triple $\langle \Gamma, \mathcal{K}_v, \mathcal{T}_h \rangle$ such that the TBox of $\mathcal{K}_v$ is not deductively modular w.r.t. $\Gamma$. Intuitively, without deductive modularity, $\mathcal{K}_v$ can arbitrarily influence the consequences of $\mathcal{T}_h$, and the oracle cannot take this into account since it does not have access to the axioms of $\mathcal{K}_v$. For the sake of generality, we do not impose any conditions on $\Gamma$.

**Theorem 5.** *Let $\mathcal{DL}_1$ be an arbitrary fragment of $\mathcal{ALCHIQ}$; let $\mathcal{DL}_2$ be an arbitrary description logic that extends $\mathcal{EL}$ and allows for $\mathcal{DL}_1$-definitions; let $\Gamma$ be an arbitrary signature; and let $\mathcal{K}_v$ be an arbitrary satisfiable $\mathcal{DL}_1$-knowledge base whose TBox is not deductively modular w.r.t. $\Gamma$. Then, no import-by-query algorithm based on ABox satisfiability oracles exists for $\mathcal{L} = \mathcal{ALCHIQ}$ and the class of inputs $\mathcal{C}[\Gamma^{\mathcal{C}}, \mathcal{K}_v^{\mathcal{C}}, \mathcal{T}_h^{\mathcal{C}}]$ where $\Gamma^{\mathcal{C}} = \Gamma$, $\mathcal{K}_v^{\mathcal{C}} = \mathcal{K}_v$, and $\mathcal{T}_h^{\mathcal{C}}$ is a $\mathcal{DL}_2$-TBox.*

*Proof.* Let $\mathcal{C}$ be a class of inputs satisfying the theorem's conditions, and let $\langle \Gamma, \mathcal{K}_v, \mathcal{T}_h^1 \rangle \in \mathcal{C}$ be an input where $\mathcal{T}_h^1 = \emptyset$. Since $\mathcal{K}_v$ is not deductively modular w.r.t. $\Gamma$, possibly complex $\mathcal{DL}_1$ concepts $C_1$ and $C_2$ exist such that $\mathsf{sig}(C_1) \subseteq \Gamma$, $\mathsf{sig}(C_2) \subseteq \Gamma$, $\mathcal{T}_v \models C_1 \sqsubseteq C_2$, and $\emptyset \not\models C_1 \sqsubseteq C_2$. Let $\mathsf{ibq^a}$ be an import-by-query algorithm for $\mathcal{L} = \mathcal{ALCHIQ}$ and $\mathcal{C}$. Finally, let $\mathcal{T}_h^2$ be as follows, where $A$, $B_1$, $B_2$, and $R$ do not occur in $\Gamma$.

$$\mathcal{T}_h^2 = \{ \ B_1 \equiv C_1, \quad B_2 \equiv C_2, \quad \top \sqsubseteq \exists R.(A \sqcap B_1), \quad A \sqcap B_2 \sqsubseteq \bot \ \} \tag{10}$$

Clearly, $\mathcal{K}_v \cup \mathcal{T}_h^1$ is satisfiable, but $\mathcal{K}_v \cup \mathcal{T}_h^2$ is not. Consider now an arbitrary $\mathcal{L}$-ABox $\mathcal{A}$ such that $\mathsf{sig}(\mathcal{A}) \subseteq \Gamma$. If $\mathcal{A} \cup \mathcal{T}_h^1$ is unsatisfiable, so is $\mathcal{A} \cup \mathcal{T}_h^2$. Conversely, assume that $\mathcal{A} \cup \mathcal{T}_h^1$ is satisfiable in a model $I' = (\triangle^{I'}, \cdot^{I'})$. Since $\emptyset \not\models C_1 \sqsubseteq C_2$, an interpretation $I'' = (\triangle^{I''}, \cdot^{I''})$ and a domain element $x \in \triangle^{I''}$ exist such that $x \in C_1^{I''}$ but $x \notin C_2^{I''}$. Without loss of generality we assume that $\triangle^{I'} \cap \triangle^{I''} = \emptyset$. Let $I$ be the following interpretation:

$\triangle^I = \triangle^{I'} \cup \triangle^{I''}$
$a^I = a^{I'}$ for each individual $a$ occurring in $\mathcal{A}$
$A^I = \{x\}$
$B_1^I = C_1^{I'} \cup C_1^{I''}$
$B_2^I = C_2^{I'} \cup C_2^{I''}$
$R^I = \{\langle o, x \rangle \mid o \in \triangle^I\}$
$X^I = X^{I'} \cup X^{I''}$ for each atomic concept or role $X \in \Gamma$

Now for each $\mathcal{ALCHIQ}$-concept $E$ such that $\mathsf{sig}(E) \subseteq \Gamma$, since $\triangle^{I'}$ and $\triangle^{I''}$ are disjoint, by a straightforward induction on the structure of $E$ one can show that $E^{I'} = E^I \cap \triangle^{I'}$ and $E^{I''} = E^I \cap \triangle^{I''}$. Furthermore, $S^{I''} \subseteq S^I$ for each atomic role $S \in \Gamma$. Thus $I \models \mathcal{A}$, and it is straightforward to check that $I \models \mathcal{T}_h^2$. Consequently, $\Omega^{\mathsf{a}}_{\mathcal{T}_h^1, \Gamma, \mathcal{L}}(\mathcal{A}) = \Omega^{\mathsf{a}}_{\mathcal{T}_h^2, \Gamma, \mathcal{L}}(\mathcal{A})$ for each $\mathcal{L}$-ABox $\mathcal{A}$ with $\mathsf{sig}(\mathcal{A}) \subseteq \Gamma$. Hence, by Proposition 1, the runs of $\mathsf{ibq^a}[\mathcal{T}_h^1, \Gamma, \mathcal{L}]$ on $\mathcal{K}_v$ coincide with the runs of $\mathsf{ibq^a}[\mathcal{T}_h^2, \Gamma, \mathcal{L}]$ on $\mathcal{K}_v$, which contradicts the fact that $\mathcal{K}_v \cup \mathcal{T}_h^1$ is satisfiable but $\mathcal{K}_v \cup \mathcal{T}_h^2$ is not. □

While Theorem 5 shows that deductive modularity is a necessary requirement for an import-by-query algorithm to exist, the following theorem shows that it is not a sufficient requirement, even if $\Gamma$ contains only atomic concepts, $\mathcal{K}_v$ is an $\mathcal{EL}$-knowledge base, and $\mathcal{T}_h$ is an $\mathcal{EL}$-TBox.

**Theorem 6.** *No import-by-query algorithm based on ABox satisfiability oracles exists for $\mathcal{L} = \mathcal{ALCHIQ}$ and the class of inputs $\mathcal{C}[\Gamma^{\mathcal{C}}, \mathcal{K}_v^{\mathcal{C}}, \mathcal{T}_h^{\mathcal{C}}]$ where $\Gamma^{\mathcal{C}}$ contains only atomic concepts, $\mathcal{K}_v^{\mathcal{C}}$ and $\mathcal{T}_h^{\mathcal{C}}$ are in $\mathcal{EL}$, and the TBox of $\mathcal{K}_v^{\mathcal{C}}$ is deductively modular w.r.t. $\Gamma^{\mathcal{C}}$.*

*Proof.* Let $\mathsf{ibq}^{\mathsf{a}}$ be an import-by-query algorithm satisfying the theorem's assumptions, let $\Gamma = \{A, B, C\}$, and let $\mathcal{K}_v$, $\mathcal{T}_h^1$, and $\mathcal{T}_h^2$ be the following $\mathcal{EL}$ knowledge bases:

$$\mathcal{K}_v = \{ \; A(a), \;\; B \sqsubseteq \exists R.C \; \} \tag{11}$$

$$\mathcal{T}_h^1 = \{ \; C \sqsubseteq \bot \; \} \tag{12}$$

$$\mathcal{T}_h^2 = \mathcal{T}_h^1 \cup \{ \; A \sqsubseteq \exists S.B \; \} \tag{13}$$

The TBox of $\mathcal{K}_v$ is clearly deductively modular w.r.t. $\Gamma$, so $\langle \Gamma, \mathcal{K}_v, \mathcal{T}_h^i \rangle \in \mathcal{C}$ for $i \in \{1, 2\}$; furthermore, $\mathcal{K}_v \cup \mathcal{T}_h^1$ is satisfiable, whereas $\mathcal{K}_v \cup \mathcal{T}_h^2$ is not. Consider now an arbitrary query ABox $\mathcal{A}$ such that $\mathsf{sig}(\mathcal{A}) \subseteq \Gamma$; since $\mathcal{A}$ contains only assertions of the form $X(a)$, $\neg X(a)$, $a \approx b$, and $a \not\approx b$ where $\mathsf{sig}(X) \subseteq \Gamma$, we have $\Omega_{\mathcal{T}_h^1, \Gamma, \mathcal{L}}^{\mathsf{a}}(\mathcal{A}) = \Omega_{\mathcal{T}_h^2, \Gamma, \mathcal{L}}^{\mathsf{a}}(\mathcal{A})$. But then, by Proposition 1, the runs of $\mathsf{ibq}^{\mathsf{a}}[\mathcal{T}_h^1, \Gamma, \mathcal{L}]$ on $\mathcal{K}_v$ coincide with the runs of $\mathsf{ibq}^{\mathsf{a}}[\mathcal{T}_h^2, \Gamma, \mathcal{L}]$ on $\mathcal{K}_v$, which contradicts the fact that $\mathcal{K}_v \cup \mathcal{T}_h^1$ is satisfiable but $\mathcal{K}_v \cup \mathcal{T}_h^2$ is not. □

While deductive modularity is not sufficient, semantic modularity is sufficient in some cases: in Section 5.1 we present an import-by-query algorithm that can be applied to the case when $\Gamma$ contains only atomic concepts, $\mathcal{K}_v$ and $\mathcal{T}_h$ are in $\mathcal{ALCHIQ}$, and the TBox of $\mathcal{K}_v$ is semantically modular w.r.t. $\Gamma$.

## 4.2 Limitations of Importing Atomic Roles

In this section, we establish the limitations of the import-by-query framework for the cases when $\Gamma$ is allowed to contain atomic roles. In particular, we show that semantic modularity is not sufficient to guarantee existence of an import-by-query algorithm.

Theorems 7 and 8 demonstrate problems that arise due to certain fundamental limitations of our oracle query languages. To understand the intuition behind these results, assume that the shared signature $\Gamma$ contains one atomic role $R$. Even in the relatively simple DL $\mathcal{EL}$, knowledge base $\mathcal{T}_h$ can imply existence of arbitrarily long $R$-chains using an axiom such as $C \sqsubseteq \exists R.C$. All of the oracle languages that we consider, however, can examine only bounded prefixes of such chains. For example, assume that we use an ABox satisfiability oracle and a query language based on $\mathcal{ALCHIQ}$. Each concept in a query ABox corresponds to a first-order formula, and it is well known that the satisfiability of such a formula in a first-order interpretation depends on the formula's quantifier depth. Since the number of oracle calls in a run of an import-by-query algorithm must be bounded, an import-by-query algorithm can examine only a bounded prefix of a model of $\mathcal{T}_h$. But this leads us to a fundamental problem: if $\mathcal{T}_h$ is changed so that it has "interesting consequences" that can be detected only by examining longer $R$-chains, then such consequences will go undetected by our algorithm and render the algorithm incorrect. Theorem 7 exploits the fact that the "interesting consequences" of $\mathcal{T}_h$ are detected by $\mathcal{K}_v$ using axioms with existentially quantified concepts (i.e., our proof uses axiom $\exists R.B_2 \sqsubseteq B_2$), whereas Theorem (8) analogously uses axioms with universally quantified concepts (i.e., $B \sqsubseteq \forall R.B$).

An alternative intuitive explanation of the results in Theorems 7 and 8 is to think of the culprit axioms $\exists R.B_2 \sqsubseteq B_2$ and $B \sqsubseteq \forall R.B$ in $\mathcal{K}_v$ as propagating information from $\mathcal{K}_v$ into $\mathcal{T}_h$. In order not to miss the "interesting consequences" of $\mathcal{T}_h$, an import-by-query algorithm must examine a "sufficiently large" portion of the hidden part of a canonical model of $\mathcal{K}_v \cup \mathcal{T}_h$ in order to correctly evaluate the culprit axioms. This, however, is impossible because no bound on the portion size can be determined from the algorithm's inputs.

**Theorem 7.** *No import-by-query algorithm based on ABox satisfiability oracles exists for $\mathcal{L} = \mathcal{ALCHIQ}$ and the class of inputs $\mathcal{C}[\Gamma^\mathcal{C}, \mathcal{K}_v^\mathcal{C}, \mathcal{T}_h^\mathcal{C}]$ where $\Gamma^\mathcal{C}$ is arbitrary, $\mathcal{K}_v^\mathcal{C}$ and $\mathcal{T}_h^\mathcal{C}$ are expressed in $\mathcal{EL}$, and the TBox of $\mathcal{K}_v^\mathcal{C}$ is semantically modular w.r.t. $\Gamma^\mathcal{C}$.*

*Proof.* Let $\mathsf{ibq^a}$ be an import-by-query algorithm satisfying the theorem's assumptions, let $\Gamma = \{A_1, A_2, R\}$, and let $\mathcal{K}_v$ be the following $\mathcal{EL}$ knowledge base:

$$\mathcal{K}_v = \{\ B_1(a),\ \ B_1 \sqsubseteq \exists S.A_1,\ \ A_2 \sqsubseteq B_2,\ \ \exists R.B_2 \sqsubseteq B_2,\ \ \exists S.B_2 \sqsubseteq \bot\ \} \tag{14}$$

The TBox of $\mathcal{K}_v$ is semantically modular w.r.t. $\Gamma$: for each interpretation $I$ of the symbols in $\Gamma$, the interpretation $J$ such that $X^J = X^I$ for each $X \in \Gamma$, $B_1^J = \emptyset$, $B_2^J = \triangle^J$, and $S^J = \emptyset$ is a model of the TBox of $\mathcal{K}_v$. Let $\mathcal{T}_h^1$ be the following $\mathcal{EL}$ TBox:

$$\mathcal{T}_h^1 = \{\ A_1 \sqsubseteq C,\ \ C \sqsubseteq \exists R.C\ \} \tag{15}$$

Since each run of $\mathsf{ibq^a}[\mathcal{T}_h^1, \Gamma, \mathcal{L}]$ on $\mathcal{K}_v$ is finite, an integer $n$ exists such that each query ABox occurring in a run contains concepts of quantifier depth at most $n$. Let $\mathcal{T}_h^2$ be the following $\mathcal{EL}$ TBox:

$$\mathcal{T}_h^2 = \{A_1 \sqsubseteq \underbrace{\exists R \ldots \exists R}_{n+1 \text{ times}} . A_2\ \} \tag{16}$$

Clearly, $\mathcal{K}_v \cup \mathcal{T}_h^1$ is satisfiable, whereas $\mathcal{K}_v \cup \mathcal{T}_h^2$ is not. Consider an arbitrary query ABox $\mathcal{A}$ occurring in a run of $\mathsf{ibq^a}[\mathcal{T}_h^1, \Gamma, \mathcal{L}]$. We next show that $\Omega^{\mathsf{a}}_{\mathcal{T}_h^1, \Gamma, \mathcal{L}}(\mathcal{A}) = \Omega^{\mathsf{a}}_{\mathcal{T}_h^2, \Gamma, \mathcal{L}}(\mathcal{A})$.

Assume that $\mathcal{T}_h^1 \cup \mathcal{A}$ is satisfiable. Since $\mathcal{A}$ is expressed in $\mathcal{ALCHIQ}$ and $\mathcal{T}_h^1$ is in $\mathcal{EL}$, a canonical forest model $I = (\triangle^I, \cdot^I)$ of $\mathcal{T}_h^1 \cup \mathcal{A}$ exists (e.g., such a model can be obtained by applying the hypertableau algorithm to $\mathcal{T}_h^1$ and $\mathcal{A}$). Due to (15), for each $x \in A_1^I$, an infinite sequence $\{\alpha_0^x, \alpha_1^x, \alpha_2^x, \ldots\} \subseteq \triangle^I$ exists such that $\alpha_0^x = x$ and $\langle \alpha_i^x, \alpha_{i+1}^x \rangle \in R^I$ for each $0 \leq i$. Let $J = (\triangle^J, \cdot^J)$ be the interpretation defined as follows:

$$\triangle^J = \triangle^I \qquad A_2^J = A_2^I \cup \{\alpha_{n+1}^x \mid x \in A_1^I\} \qquad X^J = X^I \text{ for each } X \neq A_2$$

Clearly, $J \models \mathcal{T}_h^2$. Furthermore, since $I \models \mathcal{A}$, $\mathcal{A}$ contains concepts of quantifier depth at most $n$, and $I$ and $J$ "coincide up to depth $n$," we have $J \models \mathcal{A}$. Thus, $\mathcal{T}_h^2 \cup \mathcal{A}$ is satisfiable.

Assume that $\mathcal{T}_h^2 \cup \mathcal{A}$ is satisfiable. Then a canonical forest model $I = (\triangle^I, \cdot^I)$ of $\mathcal{T}_h^2 \cup \mathcal{A}$ exists. Due to (16), for each $x \in A_1^I$, a finite sequence $\{\alpha_0^x, \alpha_1^x, \alpha_2^x, \ldots, \alpha_{n+1}^x\} \subseteq \triangle^I$ exists such that $\alpha_0^x = x$ and $\langle \alpha_i^x, \alpha_{i+1}^x \rangle \in R^I$ for each $0 \le i < n$. Let $J = (\triangle^J, \cdot^J)$ be the interpretation defined as follows:

$$\triangle^J = \triangle^I \qquad\qquad R^J = R^I \cup \{\langle \alpha_{n+1}^x, \alpha_{n+1}^x \rangle \mid x \in A_1^I\}$$
$$C^J = \{\alpha_0^x, \dots, \alpha_{n+1}^x \mid x \in A_1^I\} \qquad X^J = X^I \text{ for each } X \notin \{R, C\}$$

Clearly, $J \models \mathcal{T}_h^1$. Furthermore, since $I \models \mathcal{A}$, $C \notin \mathsf{sig}(\mathcal{A})$, $\mathcal{A}$ contains only concepts of quantifier depth at most $n$, and $I$ and $J$ "coincide up to depth $n$", we have $J \models \mathcal{A}$. Thus, $\mathcal{T}_h^1 \cup \mathcal{A}$ is satisfiable.

By Proposition 1, the runs of $\mathsf{ibq^a}[\mathcal{T}_h^1, \Gamma, \mathcal{L}]$ on $\mathcal{K}_v$ coincide with the runs of $\mathsf{ibq^a}[\mathcal{T}_h^2, \Gamma, \mathcal{L}]$ on $\mathcal{K}_v$, which contradicts the fact that $\mathcal{K}_v \cup \mathcal{T}_h^1$ is satisfiable but $\mathcal{K}_v \cup \mathcal{T}_h^2$ is not. □

**Theorem 8.** *No import-by-query algorithm based on ABox satisfiability oracles exists for $\mathcal{L} = \mathcal{ALCHIQ}$ and the class of inputs $\mathcal{C}[\Gamma^{\mathcal{C}}, \mathcal{K}_v^{\mathcal{C}}, \mathcal{T}_h^{\mathcal{C}}]$ where $\Gamma^{\mathcal{C}}$ is arbitrary, $\mathcal{K}_v^{\mathcal{C}}$ is expressed in $\mathcal{FL}_0$, $\mathcal{T}_h^{\mathcal{C}}$ is expressed in $\mathcal{EL}$, and the TBox of $\mathcal{K}_v^{\mathcal{C}}$ is semantically modular w.r.t. $\Gamma^{\mathcal{C}}$.*

*Proof.* Let $\mathsf{ibq^a}$ be an import-by-query algorithm satisfying the theorem's assumptions, let $\Gamma = \{A_1, A_2, R\}$, and let $\mathcal{K}_v$ be the following $\mathcal{FL}_0$ knowledge base.

$$\mathcal{K}_v = \{\ A_1(a),\ \ B(a),\ \ B \sqsubseteq \forall R.B,\ \ A_2 \sqcap B \sqsubseteq \bot\ \} \tag{17}$$

The TBox of $\mathcal{K}_v$ is semantically modular w.r.t. $\Gamma$: for each interpretation $I$ for $\Gamma$, the interpretation $J$ such that $X^J = X^I$ for each $X \in \Gamma$ and $B^J = \emptyset$ is a model of the TBox of $\mathcal{K}_v$. Let $\mathcal{T}_h^1$ be the $\mathcal{EL}$ TBox (15) given in the proof of Theorem 7. Since each run of $\mathsf{ibq^a}[\mathcal{T}_h^1, \Gamma, \mathcal{L}]$ on $\mathcal{K}_v$ is finite, an integer $n$ exists such that each query ABox occurring in a run contains concepts of quantifier depth at most $n$. Let $\mathcal{T}_h^2$ be the the $\mathcal{EL}$ TBox (16) from Theorem 7. Clearly, $\mathcal{K}_v \cup \mathcal{T}_h^1$ is satisfiable, whereas $\mathcal{K}_v \cup \mathcal{T}_h^2$ is not. Using arguments analogous to those from the proof of Theorem 7, one can show that $\Omega_{\mathcal{T}_h^1, \Gamma, \mathcal{L}}^{\mathsf{a}}(\mathcal{A}) = \Omega_{\mathcal{T}_h^2, \Gamma, \mathcal{L}}^{\mathsf{a}}(\mathcal{A})$ for each query ABox $\mathcal{A}$ occurring in a run of $\mathsf{ibq^a}[\mathcal{T}_h^1, \Gamma, \mathcal{L}]$. By Proposition 1, the runs of $\mathsf{ibq^a}[\mathcal{T}_h^1, \Gamma, \mathcal{L}]$ on $\mathcal{K}_v$ coincide with the runs of $\mathsf{ibq^a}[\mathcal{T}_h^2, \Gamma, \mathcal{L}]$ on $\mathcal{K}_v$, which contradicts the fact that $\mathcal{K}_v \cup \mathcal{T}_h^1$ is satisfiable but $\mathcal{K}_v \cup \mathcal{T}_h^2$ is not. □

A possible way to overcome these negative results is to prevent the axioms in $\mathcal{K}_v$ from propagating information via the roles in $\Gamma$ into the hidden part of a canonical model of $\mathcal{K}_v \cup \mathcal{T}_h$. In Section 5.1, we achieve this by requiring $\mathcal{K}_v$ to be *HT-safe*. Roughly speaking, such $\mathcal{K}_v$ is semantically modular w.r.t. $\Gamma$, but, in addition, it can be translated into a set of HT-rules $\mathcal{R}_v$ where variables $x$ and $y$ in each role atom of the form $R(x, y)$ with $R \in \Gamma$ are "guarded" by suitable concepts. For example, although the knowledge base $\mathcal{K}_v$ in (17) is semantically modular w.r.t. $\Gamma = \{A_1, A_2, R\}$, the axiom $B \sqsubseteq \forall R.B \in \mathcal{K}_v$ violates the HT-safety condition since the body of its corresponding HT-rule $B(x) \land R(x, y) \to B(y)$ does not contain a "guard" concept atom for variable $y$. In order to streamline the presentation and ensure that all notions needed to enable import-by-query reasoning are defined in one place, we formalize HT-safety in Definition 6 in Section 5.1. Unfortunately, as Theorem 9 shows, HT-safety alone does not ensure existence of an import-by-query algorithm.

**Theorem 9.** *No import-by-query algorithm based on ABox satisfiability oracles exists for $\mathcal{L} = \mathcal{ALCHIQ}$ and the class of inputs $\mathcal{C}[\Gamma^{\mathcal{C}}, \mathcal{K}_v^{\mathcal{C}}, \mathcal{T}_h^{\mathcal{C}}]$ where $\Gamma^{\mathcal{C}}$ is arbitrary, $\mathcal{K}_v^{\mathcal{C}}$ is expressed in $\mathcal{EL}$, $\mathcal{T}_h^{\mathcal{C}}$ is expressed in Horn-$\mathcal{ALCIF}$, and the TBox of $\mathcal{K}_v^{\mathcal{C}}$ is HT-safe w.r.t. $\Gamma^{\mathcal{C}}$.*

*Proof.* Let $\mathsf{ibq^a}$ be an import-by-query algorithm satisfying the theorem's assumptions, let $\Gamma = \{B, R\}$, and let $\mathcal{K}_v$ be the following $\mathcal{EL}$ knowledge base:

$$\mathcal{K}_v = \{\, A(a), \;\; B(a), \;\; A \sqsubseteq \exists R.A \,\} \tag{18}$$

The TBox of $\mathcal{K}_v$ is semantically modular w.r.t. $\Gamma$: for each interpretation $I$ for $\Gamma$, the interpretation $J$ such that $X^J = X^I$ for each $X \in \Gamma$ and $A^J = \emptyset$ is a model of the TBox of $\mathcal{K}_v$. According to Definition 6, the TBox of $\mathcal{K}_v$ is then HT-safe as well. Let $\mathcal{T}_h^1$ be the following Horn-$\mathcal{ALCIF}$ TBox:

$$\mathcal{T}_h^1 = \{\, B \sqcap C \sqsubseteq \bot, \;\; B \sqsubseteq \forall R.C, \;\; C \sqsubseteq \forall R.C, \;\; \top \sqsubseteq \,\leq 1\,R^- \,\} \tag{19}$$

Since each run of $\mathsf{ibq^a}[\mathcal{T}_h^1, \Gamma, \mathcal{L}]$ on $\mathcal{K}_v$ is finite, integers $n$ and $m$ exist such that each query ABox occurring in a run contains at most $n$ individuals and concepts of quantifier depth at most $m$. Let $k = n + m$ and let $D_0, \ldots, D_k$ be distinct and fresh atomic concepts. Let $\mathcal{T}_h^2$ be the following Horn-$\mathcal{ALC}$ TBox:

$$\begin{aligned}\mathcal{T}_h^2 = \mathcal{T}_h^1 \cup {}& \{\, D_i \sqcap D_j \sqsubseteq \bot, \mid 0 \leq i < j \leq k \,\} \cup \{\, D_{j-1} \sqsubseteq \forall R.D_j \mid 1 \leq i \leq k \,\} \\ \cup {}& \{\, B \sqsubseteq D_0, \;\; D_k \sqsubseteq \bot \,\}\end{aligned} \tag{20}$$

Clearly, $\mathcal{K}_v \cup \mathcal{T}_h^1$ is satisfiable, whereas $\mathcal{K}_v \cup \mathcal{T}_h^2$ is not. Consider an arbitrary query ABox $\mathcal{A}$ occurring in a run of $\mathsf{ibq^a}[\mathcal{T}_h^1, \Gamma, \mathcal{L}]$. We next show that $\Omega_{\mathcal{T}_h^1, \Gamma, \mathcal{L}}^{\mathsf{a}}(\mathcal{A}) = \Omega_{\mathcal{T}_h^2, \Gamma, \mathcal{L}}^{\mathsf{a}}(\mathcal{A})$. This clearly holds if $\mathcal{T}_h^1 \cup \mathcal{A}$ is unsatisfiable, so assume that $\mathcal{T}_h^1 \cup \mathcal{A}$ is satisfiable. Then, there exists a canonical forest model $I = (\triangle^I, \cdot^I)$ of $\mathcal{T}_h^1 \cup \mathcal{A}$. Consider now an arbitrary domain element $x \in B^I$. We say that a domain element $y \in \triangle^I$ is *reachable* from $x$ in $\ell$ steps if a sequence of domain elements $\alpha_0 = x, \alpha_1, \alpha_2, \ldots, \alpha_\ell = y$ exist such that $\langle \alpha_i, \alpha_{i+1} \rangle \in R^I$ for each $1 \leq i < \ell$. For each such $x$ and $y$, the axioms of $\mathcal{T}_h^1$ ensure the following properties:

1. Each such sequence is unique and it consists of unique domain elements. This is because $R^I$ is an inverse-functional relation so, for each $0 \leq i < \ell$, domain element $\alpha_i$ is the only element such that $\langle \alpha_i, \alpha_{i+1} \rangle \in R^I$, so $\alpha_i \neq \alpha_j$ for $0 < i < j \leq \ell$; furthermore, $\alpha_0 \in B^I$ and $\alpha_0 \notin C^I$, and $\alpha_i \in C^I$ for $0 < i \leq \ell$, which ensures $\alpha_0 \neq \alpha_i$ for $0 < i \leq \ell$.

2. No $x' \in B^I$ distinct from $x$ exists such that $y$ is reachable form $x'$. This is because $x_i \notin B^I$ for each $0 < i \leq \ell$ and $R^I$ is inverse-functional.

3. $\ell < k$. This is because $\mathcal{A}$ contains at most $n$ individuals and all concepts in $\mathcal{A}$ are of quantifier depth at most $m$.

Let $J = (\triangle^J, \cdot^J)$ be an interpretation defined as follows:

$$\begin{aligned}\triangle^J \;&=\; \triangle^I & X^J = X^I \text{ for each } X \in \mathsf{sig}(\mathcal{T}_h^1) & \qquad D_k^J = \emptyset \\ D_i^J \;&=\; \bigcup_{x \in B^I} \{y \in \triangle^I \mid y \text{ is reachable from } x \text{ in } i \text{ steps }\} & \text{for each } 0 \leq i < k\end{aligned}$$

Interpretations $I$ and $J$ coincide on the symbols from $\mathcal{T}_h^1$, so $J \models \mathcal{A} \cup \mathcal{T}_h^1$. Furthermore, if $y \in D_i^I$ with $i < k$, by properties 1–3 then $y \notin D_j^I$ for each $j \neq i$, so $J \models \mathcal{A} \cup \mathcal{T}_h^2$. But then, by Proposition 1, the runs of $\mathsf{ibq^a}[\mathcal{T}_h^1, \Gamma, \mathcal{L}]$ on $\mathcal{K}_v$ coincide with the runs of $\mathsf{ibq^a}[\mathcal{T}_h^2, \Gamma, \mathcal{L}]$ on $\mathcal{K}_v$, which contradicts the fact that $\mathcal{K}_v \cup \mathcal{T}_h^1$ is satisfiable but $\mathcal{K}_v \cup \mathcal{T}_h^2$ is not. $\qquad\square$

The proof of Theorem 9 uses $\mathcal{K}_v$ that implies $A \sqsubseteq \exists R^n.A$ for arbitrary $n$, where $A \notin \Gamma$ and $R \in \Gamma$. Furthermore, axioms in $\mathcal{T}_h$ containing universally quantified concepts propagate information along an $R$-chain to an unknown level $m$. An import-by-query algorithm cannot determine the depth to which it must examine a model of $\mathcal{K}_v$, which precludes the termination requirement of Definition 4. In Section 5.1, we present a sufficient acyclicity restriction on $\mathcal{K}_v$ that bounds $n$ and ensures the existence of an import-by-query algorithm.

## 4.3 ABox vs. Concept Satisfiability Oracles

In this section we show that, for $\mathcal{K}_v$ an $\mathcal{EL}$-knowledge base and and $\mathcal{T}_h$ an $\mathcal{EL}$-TBox, no import-by-query algorithm based on concept satisfiability oracles exists, even if $\mathcal{K}_v$ is HT-safe w.r.t. $\Gamma$. This is interesting because in Section 5.2.2 we present an algorithm based on ABox entailment oracles that can handle such a case. Thus, ABox oracles are strictly more expressive than concept satisfiability oracles.

**Theorem 10.** *No import-by-query algorithm based on concept satisfiability oracles exists for $\mathcal{L} = \mathcal{ALCHIQ}$ and the class of inputs $\mathcal{C}[\Gamma^{\mathcal{C}}, \mathcal{K}_v^{\mathcal{C}}, \mathcal{T}_h^{\mathcal{C}}]$ where $\Gamma^{\mathcal{C}}$ is arbitrary, $\mathcal{K}_v^{\mathcal{C}}$ and $\mathcal{T}_h^{\mathcal{C}}$ are expressed in $\mathcal{EL}$, and the TBox of $\mathcal{K}_v^{\mathcal{C}}$ is HT-safe w.r.t. $\Gamma^{\mathcal{C}}$.*

*Proof.* Let $\mathsf{ibq^c}$ be an import-by-query algorithm satisfying the theorem's assumptions, let $\Gamma = \{R\}$, and let $\mathcal{K}_v$ be the following $\mathcal{EL}$ knowledge base:

$$\mathcal{K}_v = \{\ A(a),\ \ A \sqsubseteq \exists R.A\ \} \tag{21}$$

By Definition 6, the TBox of $\mathcal{K}_v$ is HT-safe. Let $\mathcal{T}_h^1 = \emptyset$. Each run of $\mathsf{ibq^c}[\mathcal{T}_h^1, \Gamma, \mathcal{L}]$ on $\mathcal{K}_v$ is finite, so an integer $n$ exists such that each query concept occurring in a run contains concepts of quantifier depth at most $n$. Let $\mathcal{T}_h^2$ be the following $\mathcal{EL}$ TBox:

$$\mathcal{T}_h^2 = \{\ \underbrace{\exists R.\ldots.\exists R}_{n + 1\ \text{times}}.\top \sqsubseteq \bot\ \} \tag{22}$$

Clearly, $\mathcal{K}_v \cup \mathcal{T}_h^1$ is satisfiable, whereas $\mathcal{K}_v \cup \mathcal{T}_h^2$ is not. Furthermore, it is straightforward to see that, for each $\mathcal{ALCHIQ}$ concept $C$ of quantifier depth at most $n$ with $\mathsf{sig}(C) \subseteq \Gamma$, we have $\mathcal{T}_h^1 \models C \sqsubseteq \bot$ if and only if $\mathcal{T}_h^2 \models C \sqsubseteq \bot$, so $\Omega_{\mathcal{T}_h^1, \Gamma}^{\mathsf{c}}(C) = \Omega_{\mathcal{T}_h^2, \Gamma}^{\mathsf{c}}(C)$. But then, by Proposition 1, the runs of $\mathsf{ibq^c}[\mathcal{T}_h^1, \Gamma, \mathcal{L}]$ on $\mathcal{K}_v$ coincide with the runs of $\mathsf{ibq^c}[\mathcal{T}_h^2, \Gamma, \mathcal{L}]$ on $\mathcal{K}_v$, which contradicts the fact that $\mathcal{K}_v \cup \mathcal{T}_h^1$ is satisfiable but $\mathcal{K}_v \cup \mathcal{T}_h^2$ is not. □

Note that the knowledge base $\mathcal{K}_v$ used in the proof of Theorem 10 is analogous to the one from the proof of Theorem 9—that is, it entails a cyclic axiom of the form $A \sqsubseteq \exists R.A$ with $R \in \Gamma$ but $A \notin \Gamma$. The negative result from Theorem 9, however, does not apply in this case because $\mathcal{T}_h$ is expressed in $\mathcal{EL}$. The algorithm presented in Section 5.2.2 can handle such knowledge bases via an ABox entailment oracle. Intuitively, this is because ABoxes can encode cyclic structures, whereas concepts cannot.

## 5. Import-by-Query Algorithms

In this section, we identify several cases in which import-by-query algorithms exist. For simplicity, throughout this section we assume that $\mathcal{K}_v$ does not contain $\Gamma$-modal concepts; by Theorem 3 this is without loss of generality.

To overcome the negative results from Section 4, in Sections 5.1.1 and 5.1.2 we introduce the HT-safety and acyclicity conditions, respectively, that $\mathcal{K}_v$ must satisfy in order to prevent the undesirable interactions between the axioms of $\mathcal{K}_v$ and $\mathcal{T}_h$. Furthermore, in the rest of this paper we assume that $\mathcal{K}_v$ is preprocessed as described by Motik et al. (2009) into the corresponding set of HT-rules $\mathcal{R}_v$ and ABox $\mathcal{A}_v$; this will be convenient because HT-rules do not contain nested quantifiers. We thus formulate HT-safety and acyclicity in terms of $\mathcal{R}_v$ and $\mathcal{A}_v$, and we define $\mathcal{K}_v$ as being HT-safe (resp. acyclic) if the corresponding $\mathcal{R}_v$ and $\mathcal{A}_v$ are HT-safe (resp. acyclic). All our algorithms take as inputs $\mathcal{R}_v$ and $\mathcal{A}_v$, and we specify the allowed inputs using classes $\mathcal{C}[\Gamma^{\mathcal{C}}, \mathcal{R}_v^{\mathcal{C}} \cup \mathcal{A}_v^{\mathcal{C}}, \mathcal{T}_h^{\mathcal{C}}]$ of triples $\langle \Gamma^{\mathcal{C}}, \mathcal{R}_v^{\mathcal{C}} \cup \mathcal{A}_v^{\mathcal{C}}, \mathcal{T}_h^{\mathcal{C}} \rangle$ where $\mathcal{R}_v$ is a set of HT-clauses, $\mathcal{A}_v$ is a normalized ABox, and $\mathsf{sig}(\mathcal{R}_v^{\mathcal{C}} \cup \mathcal{A}_v^{\mathcal{C}}) \cap \mathsf{sig}(\mathcal{T}_h^{\mathcal{C}}) \subseteq \Gamma^{\mathcal{C}}$.

In Section 5.1 we present a general import-by-query algorithm based on ABox satisfiability oracles that is applicable to the case when $\mathcal{K}_v$ imports both atomic concepts and roles, and $\mathcal{K}_v$ and $\mathcal{T}_h$ are expressed in $\mathcal{ALCHIQ}$. In order for the algorithm to be applicable, however, $\mathcal{K}_v$ must be both HT-safe and acyclic. If $\Gamma$ contains only atomic concepts, then acyclicity is vacuously satisfied for each $\mathcal{K}_v$ and HT-safety becomes equivalent to semantic modularity; thus, if only atomic concepts are shared, our algorithms is applicable whenever $\mathcal{K}_v$ is semantically modular w.r.t. $\Gamma$.

The algorithm from Section 5.1, however, is unlikely to be suitable for practice due to a high degree of nondeterminism. Therefore, in Section 5.2.1 we present an import-by-query algorithm based on ABox entailment oracles that, we believe, is suited for implementation and optimization. The algorithm requires $\mathcal{T}_h$ to be a Horn knowledge base, which allows the algorithm to be more goal-oriented.

The practical algorithm from Section 5.2.1 can readily be applied to $\mathcal{EL}$ knowledge bases, but it is not guaranteed to be optimal. Therefore, in Section 5.2.2 we present an $\mathcal{EL}$-specific import-by-query algorithm for the case when $\mathcal{K}_v$ and $\mathcal{T}_h$ are expressed in $\mathcal{EL}$. In addition to being optimal on $\mathcal{EL}$ knowledge bases, this $\mathcal{EL}$-specific algorithm does not require $\mathcal{K}_v$ to be acyclic and it somewhat relaxes the HT-safety requirement.

## 5.1 Import-by-Query in $\mathcal{ALCHIQ}$

We next present an import-by-query algorithm based on ABox satisfiability oracles that is applicable to a set of HT-rules $\mathcal{R}_v$ and a TBox $\mathcal{T}_h$ in $\mathcal{ALCHIQ}$. No assumptions are made on the type of symbols in $\Gamma$: $\mathcal{R}_v$ can reuse both atomic concepts and roles from $\mathcal{T}_h$.

### 5.1.1 HT-Safety

We now define the HT-safety condition that allows us to overcome the negative results of Theorems 7 and 8, and that also guarantees semantic modularity required to overcome the negative results of Theorems 5 and 6. If $\Gamma$ contains only atomic concepts, then HT-safety reduces to the semantic modularity of $\mathcal{R}_v$ w.r.t. $\Gamma$.

The notion of HT-safety for $\mathcal{R}_v$ consists of the following building blocks. We first identify the *safe* concepts—that is, concepts private to $\mathcal{R}_v$ that should not be "propagated" into the models of $\mathcal{T}_h$. Next, we transform $\mathcal{R}_v$ into a *reduct* by replacing in $\mathcal{R}_v$ all safe concepts with $\bot$, and we require the reduct to be semantically modular w.r.t. $\Gamma$. The latter property ensures that any interpretation for the symbols in $\Gamma$ can be extended to an interpretation of the symbols in $\mathcal{R}_v$ by interpreting safe concepts as the empty set. Finally, as motivated in Section 4.2, we impose a syntactic restriction on each HT-rule $\varrho \in \mathcal{R}_v$: for each body atom $R(x, y)$ in $\varrho$ with $R \in \Gamma$, we require variables $x$ and $y$ to be "guarded" by a safe concept.

**Definition 6.** *Let $\mathcal{R}_v$ be a set of HT-rules and let $\Gamma$ be a signature. The set of safe concepts of $\mathcal{R}_v$ and $\Gamma$ is the smallest set $\mathsf{safe}(\mathcal{R}_v, \Gamma)$ such that, for each HT-rule $\varrho \in \mathcal{R}_v$ whose body contains an atom of the form $R(x, y_i)$ or $R(y_i, x)$ with $R \in \Gamma$ and an atom of the form $A(x)$ or $A(y_i)$ with $A \notin \Gamma$, we have $A \in \mathsf{safe}(\mathcal{R}_v, \Gamma)$.*

*The reduct of $\mathcal{R}_v$ w.r.t. $\Gamma$ is the set of rules obtained from $\mathcal{R}_v$ by removing each rule containing a concept in $\mathsf{safe}(\mathcal{R}_v, \Gamma)$ in the body, and then removing from the head of the remaining rules each atom containing a concept in $\mathsf{safe}(\mathcal{R}_v, \Gamma)$.*

*The set $\mathcal{R}_v$ is HT-safe w.r.t. $\Gamma$ if*

1. *the reduct of $\mathcal{R}_v$ w.r.t. $\Gamma$ is semantically modular w.r.t. $\Gamma$, and*

2. *for each rule $\varrho \in \mathcal{R}_v$ and each body atom of $\varrho$ of the form $R(x, y_i)$ or $R(y_i, x)$ with $R \in \Gamma$, the body of $\varrho$ contains atoms $A(x)$ and $B(y_i)$ such that $A, B \in \mathsf{safe}(\mathcal{R}_v, \Gamma)$.*

HT-safety invalidates the proofs of Theorems 7 and 8: the knowledge bases $\mathcal{K}_v$ used in the proofs of these two theorems are not HT-safe w.r.t. the respective signatures $\Gamma$. In particular, consider $\mathcal{K}_v$ used in the proof of Theorem 7. The set of HT-rules $\mathcal{R}_v$ obtained from the TBox of $\mathcal{K}_v$ is shown below.

$$B_1 \sqsubseteq \exists S.A_1 \qquad \rightsquigarrow \qquad B_1(x) \rightarrow \exists S.A_1(x) \qquad (23)$$

$$A_2 \sqsubseteq B_2 \qquad \rightsquigarrow \qquad A_2(x) \rightarrow B_2(x) \qquad (24)$$

$$\exists R.B_2 \sqsubseteq B_2 \qquad \rightsquigarrow \qquad R(x, y) \wedge B_2(y) \rightarrow B_2(x) \qquad (25)$$

$$\exists S.B_2 \sqsubseteq \bot \qquad \rightsquigarrow \qquad S(x, y) \wedge B_2(y) \rightarrow \bot \qquad (26)$$

Now $\mathsf{safe}(\mathcal{R}_v, \Gamma) = \{B_2\}$. It is straightforward to see that the reduct of $\mathcal{R}_v$ w.r.t. $\Gamma$, shown below, is not semantically modular w.r.t. $\Gamma = \{A_1, A_2, R\}$.

$$B_1(x) \rightarrow \exists S.A_1(x) \qquad (27)$$

$$A_2(x) \rightarrow \bot \qquad (28)$$

Consider now $\mathcal{K}_v$ used in the proof of Theorem 8. The set of HT-rules $\mathcal{R}_v$ obtained from the TBox of $\mathcal{K}_v$ is shown below.

$$B \sqsubseteq \forall R.B \qquad \rightsquigarrow \qquad B(x) \wedge R(x, y) \rightarrow B(y) \qquad (29)$$

$$A_2 \sqcap B \sqsubseteq \bot \qquad \rightsquigarrow \qquad A_2(x) \wedge B(x) \rightarrow \bot \qquad (30)$$

Now $\mathsf{safe}(\mathcal{R}_v, \Gamma) = \{B\}$, so the reduct of $\mathcal{R}_v$ w.r.t. $\Gamma$ is empty and thus semantically modular w.r.t. $\Gamma = \{A_1, A_2, R\}$; however, the first HT-rule does not satisfy Condition 2 from Definition 6 since the rule body does not contain an atom of the form $A(y)$ with $A \in \mathsf{safe}(\mathcal{R}_v, \Gamma)$.

Note that, if $\Gamma$ contains only atomic concepts, then $\mathsf{safe}(\mathcal{R}_v, \Gamma) = \emptyset$. The reduct of $\mathcal{R}_v$ w.r.t. $\Gamma$ is then equal to $\mathcal{R}_v$, so Condition 1 from Definition 6 holds if and only if $\mathcal{R}_v$ is semantically modular w.r.t. $\Gamma$; furthermore, Condition 2 vacuously holds for $\mathcal{R}_v$. Thus, HT-safety reduces to semantic modularity w.r.t. $\Gamma$ if only atomic concepts are shared. The following proposition shows that, given an interpretation for the symbols in $\Gamma$, we can obtain a model of $\mathcal{R}_v$ by interpreting safe concepts as the empty set.

**Proposition 3.** *Let $\mathcal{R}_v$ be a set of HT-rules that is HT-safe w.r.t. $\Gamma$. Then, for each interpretation $I$ of the symbols in $\Gamma$, a model $J$ of $\mathcal{R}_v$ exists such that $\triangle^J = \triangle^I$, $X^J = X^I$ for each symbol $X \in \Gamma$, and $X^J = \emptyset$ for each atomic concept $X \in \mathsf{safe}(\mathcal{R}_v, \Gamma)$.*

*Proof.* Let $I$ be an interpretation for the symbols in $\Gamma$, and let $\mathcal{R}'_v$ be the reduct of $\mathcal{R}_v$ w.r.t $\Gamma$. Since $\mathcal{R}'_v$ is semantically modular w.r.t. $\Gamma$, a model $I'$ of $\mathcal{R}'_v$ exists such that $\triangle^{I'} = \triangle^I$ and $X^{I'} = X^I$ for each symbol $X \in \Gamma$. Let $J$ be the interpretation obtained from $I'$ by setting $X^J = \emptyset$ for each $X \in \mathsf{safe}(\mathcal{R}_v, \Gamma)$. Consider now an arbitrary HT-rule $\varrho \in \mathcal{R}_v$. If some $A \in \mathsf{safe}(\mathcal{R}_v, \Gamma)$ occurs in the body of $\varrho$, then $A^J = \emptyset$ clearly implies $J \models \varrho$. Otherwise, let $\varrho' \in \mathcal{R}'_v$ be the rule obtained by removing in $\varrho$ all head atoms that contain a safe concept; then $I' \models \varrho'$ clearly implies $J \models \varrho$. Consequently, $J \models \mathcal{R}_v$. $\qquad\square$

Finally, note that HT-safety is not a syntactic condition; in fact, checking HT-safety is undecidable in general because it requires checking semantic modularity of a set of HT-rules w.r.t. a signature. As mentioned in Section 2.3, however, several practically useful syntactic conditions are known that guarantee semantic modularity (Cuenca Grau, Horrocks, Kazakov, & Sattler, 2008), and any such condition can be used to obtain a purely syntactic HT-safety notion.

### 5.1.2 ACYCLICITY

The negative result of Theorem 9 relies on $\mathcal{K}_v$ containing a cyclic axiom $A \sqsubseteq \exists R.A$ with $R \in \Gamma$ and $A \notin \Gamma$. We next present a sufficient condition that can detect such cycles in polynomial time.

Our test involves a set of function-free first-order formulae with equality $\mathsf{D}(\mathcal{R}_v, \mathcal{A}_v)$ whose consequences "summarize" the models of $\mathcal{R}_v \cup \mathcal{T}_h \cup \mathcal{A}_v$; more precisely, the projection of each canonical model of $\mathcal{R}_v \cup \mathcal{T}_h \cup \mathcal{A}_v$ to the symbols in $\mathsf{sig}(\mathcal{R}_v)$ can be homomorphically embedded into the set of ground facts entailed by $\mathsf{D}(\mathcal{R}_v, \mathcal{A}_v)$. Intuitively, since the axioms of $\mathcal{T}_h$ are not available, the facts entailed by $\mathsf{D}(\mathcal{R}_v, \mathcal{A}_v)$ should reflect all possible consequences of $\mathcal{T}_h$ and all information that can be derived using $\mathcal{R}_v \cup \mathcal{A}_v$. Theory $\mathsf{D}(\mathcal{R}_v, \mathcal{A}_v)$ also keeps track of the paths in the "visible" part of the canonical models of $\mathcal{R}_v \cup \mathcal{T}_h \cup \mathcal{A}_v$ by using two special binary predicates: Succ keeps track the "successorship" relation between domain elements, and $\Gamma$-Desc keeps track the "descendant" relation via roles contained in $\Gamma$. The acyclicity condition then checks whether the $\Gamma$-Desc relation as entailed by $\mathsf{D}(\mathcal{R}_v, \mathcal{A}_v)$ is cyclic; if this is not the case, we can establish a bound on the length of paths of roles in $\Gamma$.

**Definition 7.** *Let $\mathcal{R}_v$ be a set of HT-rules, let $\mathcal{A}_v$ be an ABox, and let $\Gamma$ be a signature. For each atomic concept $A \in \mathsf{sig}(R_v) \cup \mathsf{sig}(\mathcal{A}_v)$, let $v_A$ and $v_{\neg A}$ be individuals uniquely associated with $A$ and $\neg A$, respectively; furthermore, let Succ and $\Gamma$-Desc be binary predicates not occurring in $\mathcal{R}_v$ or $\mathcal{A}_v$. Function $\mathsf{tt}(\cdot)$ maps each atom $\alpha$ occurring in $\mathcal{R}_v \cup \mathcal{A}_v$ into a conjunction of atoms as follows, where $z$ is an arbitrary term:*

- $\mathsf{tt}(\neg A(z)) = \top$;

- $\mathsf{tt}(\geq n\, R.C(z)) = \mathsf{ar}(R, z, v_C) \wedge \mathsf{tt}(C(v_C)) \wedge Succ(z, v_C)$; *and*

- $\mathsf{tt}(\alpha) = \alpha$ *for each atom $\alpha$ of the form not covered by the above two cases.*

*Furthermore, $\mathsf{D}(\mathcal{R}_v, \mathcal{A}_v)$ is the set of function-free formulas of first-order logic with equality defined as follows, where all variables are implicitly universally quantified.*

- *For each assertion $\alpha \in \mathcal{A}_v$, set $\mathsf{D}(\mathcal{R}_v, \mathcal{A}_v)$ contains $\mathsf{tt}(\alpha)$.*

- *For each individual $c$ occurring in $\mathcal{A}_v$ and each atomic concept $A \in \Gamma$, set $\mathsf{D}(\mathcal{R}_v, \mathcal{A}_v)$ contains $A(c)$.*

- *For each HT-rule $\varrho \in \mathcal{R}_v$ of the form (1) and each $1 \leq j \leq n$, set $\mathsf{D}(\mathcal{R}_v, \mathcal{A}_v)$ contains the following formula:*

$$\mathsf{tt}(U_1) \wedge \ldots \wedge \mathsf{tt}(U_m) \rightarrow \mathsf{tt}(V_j) \tag{31}$$

- *For each atomic concept $A \in \Gamma$, set $\mathsf{D}(\mathcal{R}_v, \mathcal{A}_v)$ contains the following formula:*

$$Succ(z_1, z_2) \rightarrow A(z_2) \tag{32}$$

- *For all atomic roles $R, R' \in \Gamma$, set $\mathsf{D}(\mathcal{R}_v, \mathcal{A}_v)$ contains the following formulae:*

$$R'(z_1, z_2) \rightarrow R(z_1, z_2) \tag{33}$$
$$R'(z_1, z_2) \rightarrow R(z_2, z_1) \tag{34}$$
$$R(z, z_1) \wedge R'(z, z_2) \rightarrow z_1 \approx z_2 \tag{35}$$
$$R(z_1, z) \wedge R'(z_2, z) \rightarrow z_1 \approx z_2 \tag{36}$$
$$R(z_1, z) \wedge R'(z, z_2) \rightarrow z_1 \approx z_2 \tag{37}$$

- *For each atomic role $R \in \Gamma$, set $\mathsf{D}(\mathcal{R}_v, \mathcal{A}_v)$ contains the following formulae:*

$$Succ(z_1, z_2) \wedge R(z_1, z_2) \rightarrow \Gamma\text{-}Desc(z_1, z_2) \tag{38}$$
$$\Gamma\text{-}Desc(z_1, z_2) \wedge \Gamma\text{-}Desc(z_2, z_3) \rightarrow \Gamma\text{-}Desc(z_1, z_3) \tag{39}$$

*Set $\mathsf{D}(\mathcal{R}_v, \mathcal{A}_v)$ contains a* harmful cycle *if $\mathsf{D}(\mathcal{R}_v, \mathcal{A}_v) \models \Gamma\text{-}Desc(v_C, v_C)$ for some $v_C$. Furthermore, $\mathcal{R}_v \cup \mathcal{A}_v$ is acyclic w.r.t. $\Gamma$ if $\mathsf{D}(\mathcal{R}_v, \mathcal{A}_v)$ does not contain a harmful cycle.*

The set of formulae $\mathsf{D}(\mathcal{R}_v, \mathcal{A}_v)$ can be straightforwardly transformed into an equivalent datalog program with equality using the well-known equivalences of first-order logic; therefore, we often refer to $\mathsf{D}(\mathcal{R}_v, \mathcal{A}_v)$ as a datalog program.

Acyclicity allows us to express axioms $\delta_6$ and $\delta_8$ from Table 3. Intuitively, acyclicity ensures that the "visible parts" of the canonical forest models of $\mathcal{R}_v \cup \mathcal{T}_h \cup \mathcal{A}_v$ do not contain infinite chains of roles from $\Gamma$; we use this property in our algorithm to define a suitable blocking condition. We explain this intuition by means of an example. Let $\Gamma = \{C, R, U\}$ where $C$ is a concept and $R$ and $U$ roles, $\mathcal{A}_v = \{A(a)\}$, and $\mathcal{R}_v$ contains the following HT-rules; the corresponding formulae in $\mathsf{D}(\mathcal{R}_v, \mathcal{A}_v)$ are shown after the "$\rightsquigarrow$" symbol. Note that $\mathcal{R}_v$ is HT-safe w.r.t. $\Gamma$.

$$A(x) \rightarrow \exists R.B(x) \quad \rightsquigarrow \quad A(x) \rightarrow R(x, v_B) \wedge B(v_B) \wedge Succ(x, v_B) \tag{40}$$
$$A(x) \rightarrow \exists S.C(x) \quad \rightsquigarrow \quad A(x) \rightarrow S(x, v_C) \wedge C(v_C) \wedge Succ(x, v_C) \tag{41}$$

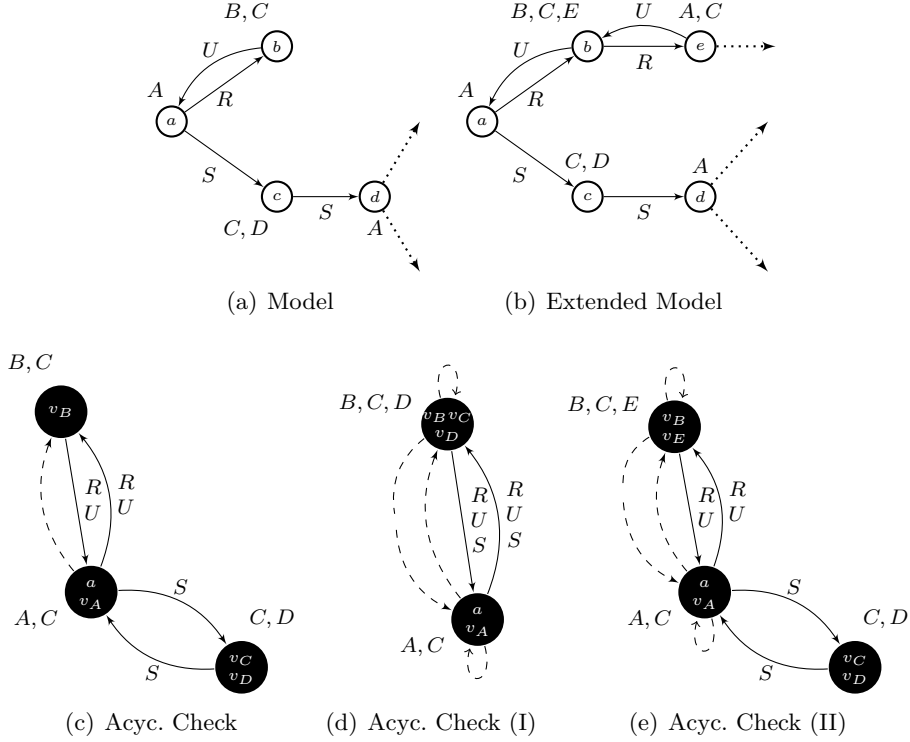

(a) Model  (b) Extended Model

(c) Acyc. Check  (d) Acyc. Check (I)  (e) Acyc. Check (II)

Figure 1: Canonical Models and Acyclicity

$$A(x) \to \exists S.D(x) \quad \rightsquigarrow \quad A(x) \to S(x, v_D) \land D(v_D) \land \mathrm{Succ}(x, v_D) \quad (42)$$

$$S(x, y_1) \land S(x, y_2) \to y_1 \approx y_2 \quad \rightsquigarrow \quad S(x, y_1) \land S(x, y_2) \to y_1 \approx y_2 \quad (43)$$

$$C(x) \land D(x) \to \exists S.A(x) \quad \rightsquigarrow \quad C(x) \land D(x) \to S(x, v_A) \land A(v_A) \land \mathrm{Succ}(x, v_A) \quad (44)$$

Consider also the following hidden TBox expressed in $\mathcal{ALCHIQ}$:

$$\mathcal{T}_h = \{\ \top \sqsubseteq\ \leq 1\, R.\top,\ \ R \sqsubseteq U^-,\ \ \exists U.\top \sqsubseteq C\ \} \quad (45)$$

Figure 1(a) shows a canonical model $I$ of $\mathcal{R}_v \cup \mathcal{T}_h \cup \mathcal{A}_v$. Furthermore, Figure 1(c) shows the ground atoms entailed by $\mathsf{D}(\mathcal{R}_v, \mathcal{A}_v)$ represented as a graph $G$ in which solid arrows show roles $R$, $U$, and $S$, and dashed arrows show the special predicate Γ-Desc; for clarity, the atoms involving the special predicate Succ have not been included in this and the following figures. Note that $\mathsf{D}(\mathcal{R}_v, \mathcal{A}_v)$ entails $R(v_A, v_B)$ and $R(a, v_B)$; these atoms, together with rules (34) and (35), entail $v_A \approx v_a$; consequently, $v_A$ and $v_a$ are represented in Figure 1(c) by the same node. Structure $G$ "summarizes" $I$ in the sense that $I$ can be homomorphically embedded into $G$. The repetitive structure of $I$ is represented in $G$ as a cycle over nodes $v_A$ and $v_C$ via $S$; however, since $S$ is not a shared symbol (i.e., $S \notin \Gamma$), this does not give rise to a harmful cycle. Consequently, $\mathcal{R}_v \cup \mathcal{A}_v$ is acyclic w.r.t. Γ, which guarantees that the "visible" part of a model of $\mathcal{R}_v \cup \mathcal{T}_h \cup \mathcal{A}_v$ does not contain $R$-chains of unbounded length, regardless of the contents of $\mathcal{T}_h$. Accordingly, the canonical model $I$ of $\mathcal{R}_v \cup \mathcal{T}_h \cup \mathcal{A}_v$ shown in Figure 1(a) contains no such $R$-chains.

Note, however, that $G$ overestimates the canonical model $I$; for example, $G$ contains an individual $v_A$ that is an instance of both $A$ and $C$, which is not reflected in $I$. Now let us assume that $\mathcal{R}'_v$ is $\mathcal{R}_v$ extended with the following HT-rule:

$$A(x) \wedge C(x) \rightarrow \exists R.C(x) \quad \leadsto \quad A(x) \wedge C(x) \rightarrow R(x, v_C) \wedge C(v_C) \wedge \mathsf{Succ}(x, v_C) \quad (46)$$

The canonical model of $\mathcal{R}'_v \cup \mathcal{T}_h \cup \mathcal{A}_v$ is clearly the same as that of $\mathcal{R}_v \cup \mathcal{T}_h \cup \mathcal{A}_v$; however, $\mathsf{D}(\mathcal{R}'_v, \mathcal{A}_v)$ contains a harmful cycle, as shown in Figure 1(d). Intuitively, $\mathsf{D}(\mathcal{R}'_v, \mathcal{A}_v)$ provides us with a conservative overestimate of the canonical models, which can in some cases detect "cycles" that do not really exist in canonical models. This is a necessary consequence of the fact that acyclicity can be checked in polynomial time.

Definition 7, however, provides us with a sufficient check. For example, let $\mathcal{R}''_v$ be $\mathcal{R}_v$ extended with the following HT-rules:

$$A(x) \rightarrow \exists R.E(x) \quad \leadsto \quad A(x) \rightarrow R(x, v_E) \wedge E(v_E) \wedge \mathsf{Succ}(x, v_E) \quad (47)$$

$$B(x) \wedge C(x) \wedge E(x) \rightarrow \exists R.A(x) \quad \leadsto \quad \begin{aligned} B(x) \wedge C(x) \wedge E(x) \rightarrow \\ R(x, v_A) \wedge A(v_A) \wedge \mathsf{Succ}(x, v_A) \end{aligned} \quad (48)$$

The canonical model of $\mathcal{R}''_v \cup \mathcal{T}_h \cup \mathcal{A}_v$ and the ground atoms entailed by $\mathsf{D}(\mathcal{R}''_v, \mathcal{A}_v)$ are shown in Figures 1(b) and 1(e), respectively. The HT-rules in $\mathcal{R}''_v \setminus \mathcal{R}_v$ enforce the existence of an infinite $R$-chain, which is reflected as a harmful cycle (e.g., the self-loop on $v_A$).

Acyclicity can indeed be checked in polynomial time, as shown next.

**Proposition 4.** *Acyclicity of $\mathcal{R}_v \cup \mathcal{A}_v$ w.r.t. $\Gamma$ can be checked in polynomial time.*

*Proof.* Let $\mathsf{D}(\mathcal{R}_v, \mathcal{A}_v)$ be as specified in Definition 7. The number of fresh individuals of the form $v_A$ and $v_{\neg A}$ is clearly linear in the size of $\mathcal{R}_v$, $\mathcal{A}_v$, and $\Gamma$, so the size of $\mathsf{D}(\mathcal{R}_v, \mathcal{A}_v)$ is polynomial in the size of $\mathcal{R}_v$, $\mathcal{A}_v$, and $\Gamma$.

We can compute the set of all positive ground atoms that follow from $\mathsf{D}(\mathcal{R}_v, \mathcal{A}_v)$ in polynomial time using forward chaining. All predicates in $\mathsf{D}(\mathcal{R}_v, \mathcal{A}_v)$ are of bounded arity, so the number of such atoms is polynomial in the size of $\mathsf{D}(\mathcal{R}_v, \mathcal{A}_v)$. This straightforwardly implies the claim of this proposition if we show that, given a set of facts and a rule $\varrho \in \mathsf{D}(\mathcal{R}_v, \mathcal{A}_v)$, we can compute the set of entailed facts in polynomial time. Rules not of the form (31) contain a bounded number of variables, so the set of entailed facts can be computed in polynomial time by simply considering all possible mappings of variables to individuals. Assume now that $\varrho$ is of the form (31). The number of variables in $\varrho$ is linear in the size of $\mathcal{R}_v$, so there are exponentially many mappings of variables to individuals. We can, however, determine the values for $x$ and $y_i$ that make the body true as follows. For each variable $y_i$, let $P_i$ be a binary relation that initially contains all pairs of individuals occurring in $\mathsf{D}(\mathcal{R}_v, \mathcal{A}_v)$; this relation will eventually contain all pairs of values for $x$ and $y_i$ that make the body of $\varrho$ true. We then remove from each $P_i$ all pairs that do not satisfy all body atoms of $\varrho$ that contain only variables $x$ and $y_i$. Next, for all $P_i$ and $P_j$, we remove all pairs $\langle c, c' \rangle$ from $P_i$ for which no $c''$ exists such that $\langle c, c'' \rangle \in P_j$. We then consider each consequent atom $\alpha$ of $\varrho$; if $\alpha$ contains only variables $x$ and $y_i$, we infer all ground atoms obtained by replacing $x$ with $c$ and $y_i$ with $c'$ for each $\langle c, c' \rangle \in P_i$; if $\alpha$ contains only variables $y_i$ and $y_j$, we infer all ground atoms $c' \approx c''$ such that an individual $c$ exists where $\langle c, c' \rangle \in P_i$ and $\langle c, c'' \rangle \in P_j$. This can clearly be done by polynomially many steps in the number of individuals in $\mathsf{D}(\mathcal{R}_v, \mathcal{A}_v)$ and the maximal number of variables in a rule in $\mathcal{R}_v$. $\qquad \square$

Table 4: Additional Derivation Rules

| | | |
|---|---|---|
| $A$-cut | If | an individual $s$ in $\mathcal{A}$ and an atomic concept $A \in \Gamma$ exist such that |
| | | 1. $s$ is not indirectly blocked in $\mathcal{A}$ and |
| | | 2. $\{A(s), \neg A(s)\} \cap \mathcal{A} = \emptyset$ |
| | then | $\mathcal{A}_1 := \mathcal{A} \cup \{A(s)\}$ and $\mathcal{A}_2 := \mathcal{A} \cup \{\neg A(s)\}$. |
| $R$-cut | If | individuals $s$ and $t$ in $\mathcal{A}$ and atomic roles $R, R' \in \Gamma$ exist such that |
| | | 1. neither $s$ nor $t$ is indirectly blocked in $\mathcal{A}$, |
| | | 2. $R'(s,t) \in \mathcal{A}$, and |
| | | 3. $\{R(s,t), \neg R(s,t)\} \cap \mathcal{A} = \emptyset$ |
| | then | $\mathcal{A}_1 := \mathcal{A} \cup \{R(s,t)\}$ and $\mathcal{A}_2 := \mathcal{A} \cup \{\neg R(s,t)\}$. |
| $R^-$-cut | If | individuals $s$ and $t$ in $\mathcal{A}$ and atomic roles $R, R' \in \Gamma$ exist such that |
| | | 1. neither $s$ nor $t$ is indirectly blocked in $\mathcal{A}$, |
| | | 2. $R'(s,t) \in \mathcal{A}$, and |
| | | 3. $\{R(t,s), \neg R(t,s)\} \cap \mathcal{A} = \emptyset$, |
| | then | $\mathcal{A}_1 := \mathcal{A} \cup \{R(t,s)\}$ and $\mathcal{A}_2 := \mathcal{A} \cup \{\neg R(t,s)\}$. |
| $\approx$-cut | If | individuals $s$, $s_1$, and $s_2$ in $\mathcal{A}$ exist such that |
| | | 1. none of $s$, $s_1$, and $s_2$ is indirectly blocked in $\mathcal{A}$, |
| | | 2. $\{s_1 \approx s_2, s_1 \not\approx s_2\} \cap \mathcal{A} = \emptyset$, and |
| | | 3. atomic roles $R, R' \in \Gamma$ exist such that |
| | | 3.1 $\{R(s, s_1), R'(s, s_2)\} \subseteq \mathcal{A}$ or |
| | | 3.2 $\{R(s_1, s), R'(s_2, s)\} \subseteq \mathcal{A}$ or |
| | | 3.2 $\{R(s_1, s), R'(s, s_2)\} \subseteq \mathcal{A}$ |
| | then | $\mathcal{A}_1 := \mathcal{A} \cup \{s_1 \approx s_2\}$ and $\mathcal{A}_2 := \mathcal{A} \cup \{s_1 \not\approx s_2\}$. |
| $\Omega^{\mathtt{a}}$-rule | If 1. | $\bot \notin \mathcal{A}$ and |
| | 2. | a connected component $\mathcal{A}'$ of $\mathcal{A}\vert_\Gamma$ exists such that $\Omega^{\mathtt{a}}_{\mathcal{T}_h, \Gamma}(\mathcal{A}') = \mathsf{f}$ |
| | then | $\mathcal{A}_1 := \mathcal{A} \cup \{\bot\}$. |

### 5.1.3 An Import-by-Query Algorithm

We next present our import-by-query algorithm applicable to $\mathcal{R}_v \cup \mathcal{A}_v$ that is HT-safe and acyclic w.r.t. $\Gamma$. The algorithm modifies the standard hypertableau algorithm as follows. First, several cut rules nondeterministically guess all "relevant" assertions involving the symbols in $\Gamma$. Second, the $\Omega^{\mathtt{a}}$-rule checks whether the guesses are indeed consistent with $\mathcal{T}_h$. Third, a relaxed blocking condition ensures termination.

**Definition 8.** *Let $\mathcal{C}[\Gamma^{\mathcal{C}}, \mathcal{R}_v^{\mathcal{C}} \cup \mathcal{A}_v^{\mathcal{C}}, \mathcal{T}_h^{\mathcal{C}}]$ be the class of inputs where $\mathcal{R}_v^{\mathcal{C}} \cup \mathcal{A}_v^{\mathcal{C}}$ is acyclic w.r.t. $\Gamma^{\mathcal{C}}$, $\mathcal{R}_v^{\mathcal{C}}$ is HT-safe w.r.t. $\Gamma^{\mathcal{C}}$, and $\mathcal{T}_h^{\mathcal{C}}$ is an $\mathcal{ALCHIQ}$ TBox. The $\mathcal{ALCHIQ}$ $\Omega^{\mathtt{a}}$-algorithm takes a triple $\langle \Gamma, \mathcal{R}_v \cup \mathcal{A}_v, \mathcal{T}_h \rangle \in \mathcal{C}$ and is obtained by modifying Definition 1 as follows.*

**Blocking.** *An unnamed individual $s$ is blocking-relevant in $\mathcal{A}$ if, for $s'$ the predecessor of $s$, we have*

$$\mathcal{L}_{\mathcal{A}}(s, s') \cap \Gamma = \mathcal{L}_{\mathcal{A}}(s', s) \cap \Gamma = \emptyset.$$

*Then, each individual $s$ in an ABox $\mathcal{A}$ is assigned a blocking status in the same way as in Definition 1, with the difference that $s$ is directly blocked by $t$ if, in addition to the conditions in Definition 1, both $s$ and $t$ are blocking-relevant.*

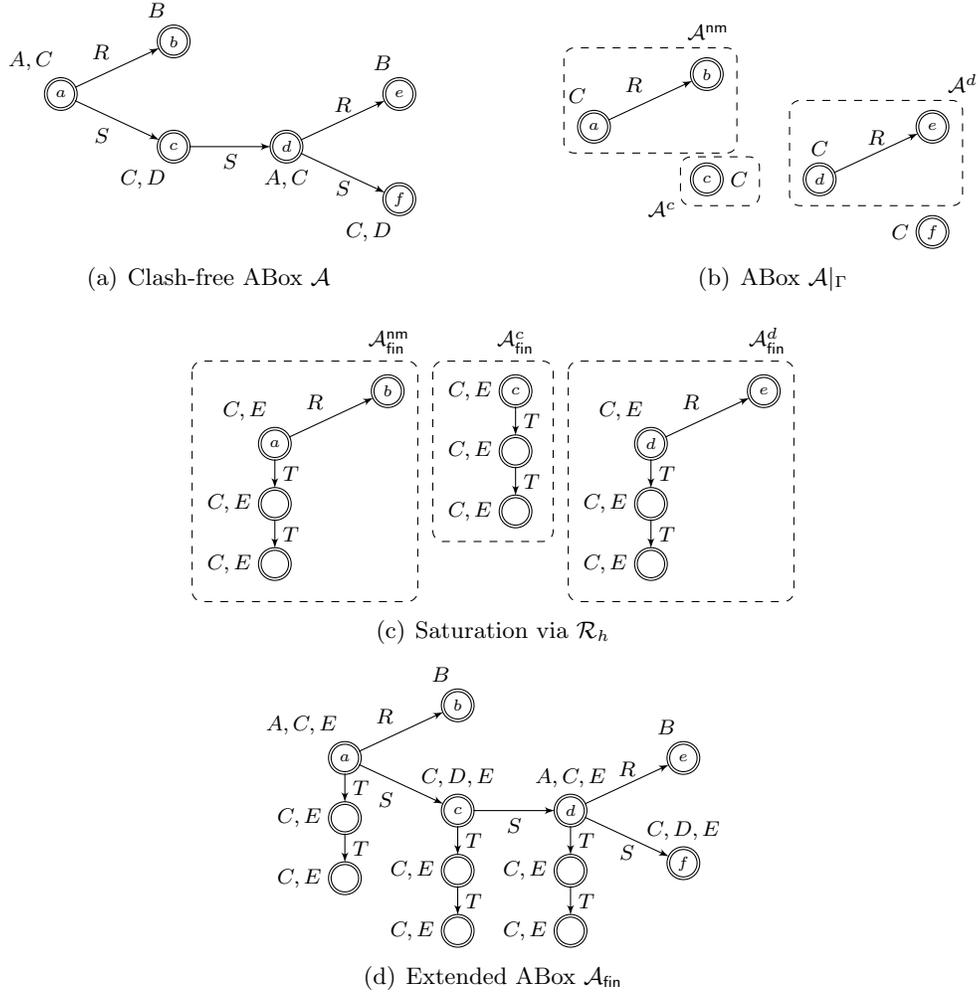

Figure 2: Completeness of the $\mathcal{ALCHIQ}$ $\Omega^{\mathsf{a}}$-algorithm

**Derivation Rules.** *The derivation rules are given in Tables 2 and 4, where $\mathcal{A}|_\Gamma$ is the ABox obtained from $\mathcal{A}$ by removing each assertion containing an indirectly blocked individual and each assertion $\alpha$ such that $\mathsf{sig}(\alpha) \not\subseteq \Gamma$.*

In Section 5.1.4 we show that some of the cut rules in Table 4 are not needed if we know that $\mathcal{T}_h$ is expressed in a description logic between $\mathcal{ALC}$ and $\mathcal{ALCHIQ}$. Our algorithm is indeed an import-by-query algorithm, as we show next.

**Theorem 11.** *The $\mathcal{ALCHIQ}$ $\Omega^{\mathsf{a}}$-algorithm is an import-by-query algorithm based on ABox satisfiability oracles for the class of inputs $\mathcal{C}[\Gamma^\mathcal{C}, \mathcal{R}_v^\mathcal{C} \cup \mathcal{A}_v^\mathcal{C}, \mathcal{T}_h^\mathcal{C}]$ from Definition 8. The algorithm can be implemented such that it runs in N2ExpTime in $N$, and the total number of oracle queries and the size of each query are both at most exponential in $N$, where $N = |\mathcal{R}_v \cup \mathcal{A}_v| + |\Gamma|$ for the input $\mathcal{R}_v$, $\mathcal{A}_v$, and $\Gamma$.*

The proof of Theorem 11 is lengthy and quite technical, so we defer it to the appendix and next discuss only the intuitions. The derivation rules from Table 2 are clearly sound.

Furthermore, due to acyclicity, the chains of assertions involving roles from $\Gamma$ are bounded in length, which enables blocking and ensures termination. We next sketch the completeness argument. In particular, for completeness we need to show that the existence of a clash-free ABox in a derivation to which no rule is applicable implies the satisfiability of the input. Let $\mathcal{A}$ be a clash-free ABox labeling the leaf of a derivation for $\langle \Gamma, \mathcal{R}_v \cup \mathcal{A}_v, \mathcal{T}_h \rangle$, and let $\mathcal{R}_h$ be the set of HT-rules corresponding to $\mathcal{T}_h$. Each model of $\mathcal{R}_v \cup \mathcal{A} \cup \mathcal{T}_h$ can be extended to a model of $\mathcal{R}_v \cup \mathcal{A}_v \cup \mathcal{T}_h$, so it suffices to show the satisfiability of $\mathcal{R}_v \cup \mathcal{A} \cup \mathcal{T}_h$. To this end, we extend $\mathcal{A}$ to a clash-free ABox $\mathcal{A}_{\mathsf{fin}}$ such that no derivation rule of the standard hypertableau algorithm is applicable to $\mathcal{R}_v \cup \mathcal{R}_h$ and $\mathcal{A}_{\mathsf{fin}}$; thus, $\mathcal{R}_v \cup \mathcal{A}_{\mathsf{fin}} \cup \mathcal{T}_h$ is satisfiable, and since $\mathcal{A} \subseteq \mathcal{A}_{\mathsf{fin}}$ so is $\mathcal{R}_v \cup \mathcal{A} \cup \mathcal{T}_h$ by monotonicity. The construction of $\mathcal{A}_{\mathsf{fin}}$ proceeds as follows:

1. We split the projection $\mathcal{A}|_\Gamma$ of $\mathcal{A}$ to $\Gamma$. In particular, we define $\mathcal{A}^{\mathsf{nm}}$ as the ABox containing all assertions of $\mathcal{A}|_\Gamma$ involving individuals reachable from a named individual; furthermore, for each nonblocked blocking-relevant individual $t$, we define $\mathcal{A}^t$ as the ABox containing all assertions of $\mathcal{A}|_\Gamma$ involving individuals reachable from $t$.

2. We apply the standard hypertableau algorithm to $\mathcal{R}_h$ and each of the connected components of $\mathcal{A}^{\mathsf{nm}}$, and $\mathcal{R}_h$ and each $\mathcal{A}^t$; let $\mathcal{A}^{\mathsf{nm}}_{\mathsf{fin}}$ and $\mathcal{A}^t_{\mathsf{fin}}$ be clash-free ABoxes labeling leaves of the respective derivations. The $\Omega^{\mathsf{a}}$-rule is not applicable to $\mathcal{A}$ so such ABoxes exist.

3. We define $\mathcal{A}_{\mathsf{fin}}$ as the union of $\mathcal{A}$, $\mathcal{A}^{\mathsf{nm}}_{\mathsf{fin}}$, all $\mathcal{A}^t_{\mathsf{fin}}$, and all assertions $C(s)$ such that $s$ is blocked in $\mathcal{A}$ by the blocker $s'$, $C(s') \in \mathcal{A}^{s'}_{\mathsf{fin}}$, and $\mathsf{sig}(C) \subseteq \mathsf{sig}(\mathcal{R}_h)$.

Let us call the individuals from $\mathcal{A}$ *old*, and the individuals introduced in the second step *new*; we then observe the following. (1) Due to the cut rules, the second step above cannot derive fresh assertions involving only old individuals and the symbols in $\Gamma$ without leading to a contradiction. (2) Each of the connected components of $\mathcal{A}^{\mathsf{nm}}$ and each $\mathcal{A}^t$ are disjoint, so the HT-rules from $\mathcal{R}_h$ can be applied in $\mathcal{A}_{\mathsf{fin}}$ only to subsets that correspond to a connected component of $\mathcal{A}^{\mathsf{nm}}$ and $\mathcal{A}^t$. (3) Due to (1), no HT-rule from $\mathcal{R}_v$ can become applicable to assertions involving only old individuals. (4) Due to HT-safety, no HT-rule from $\mathcal{R}_v$ can become applicable to an assertion of $\mathcal{A}_{\mathsf{fin}}$ that involves a new individual. (5) Due to (1) and the third step from the construction above, if an individual $s$ is blocked in $\mathcal{A}$, $\mathcal{A}^{\mathsf{nm}}_{\mathsf{fin}}$, or $\mathcal{A}^t_{\mathsf{fin}}$, then $s$ is blocked in $\mathcal{A}_{\mathsf{fin}}$ as well. Then, (1)–(5) imply that no derivation rule of the standard hypertableau algorithm is applicable to $\mathcal{R}_v \cup \mathcal{R}_h$ and $\mathcal{A}_{\mathsf{fin}}$, which proves completeness.

We explain this intuition on an example where $\Gamma = \{C, R\}$, $\mathcal{A}_v = \{A(a)\}$, $\mathcal{R}_v$ consists of HT-rules (40)–(44), and $\mathcal{T}_h$ is defined as follows:

$$\mathcal{T}_h = \{\, \exists R.\top \sqsubseteq C, \ \ C \sqsubseteq \exists T.C, \ \ C \sqsubseteq E \,\} \tag{49}$$

As shown in Section 5.1.2, $\mathcal{R}_v \cup \mathcal{A}_v$ is acyclic w.r.t. $\Gamma$, so the $\mathcal{ALCHIQ}$ $\Omega^{\mathsf{a}}$-algorithm is applicable. The algorithm produces a derivation in which a leaf is labeled with the ABox $\mathcal{A}$ shown in Figure 2(a); for readability, we show neither the negative assertions nor the assertions involving complex concepts. Individual $f$ is directly blocked by $c$ in $\mathcal{A}$, and assertions $C(a)$ and $C(d)$ are introduced by the $A$-cut rule. To construct $\mathcal{A}_{\mathsf{fin}}$, the assertions containing a symbol not in $\Gamma$ are removed, resulting in the ABox $\mathcal{A}|_\Gamma$ shown in

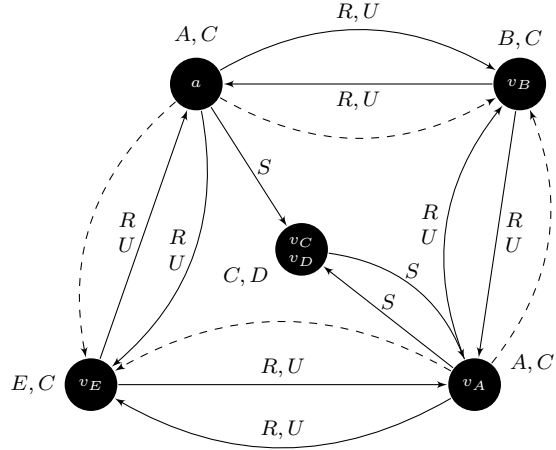

Figure 3: Acyclicity Check for $\mathcal{T}_h$ in $\mathcal{ALCHI}$

Figure 2(b). This ABox is then split into connected components $\mathcal{A}^{nm}$, $\mathcal{A}^c$, and $\mathcal{A}^d$; note that $c$ and $d$ are the only nonblocked blocking-relevant individuals. Next, $\mathcal{A}^{nm}$, $\mathcal{A}^c$, and $\mathcal{A}^d$ are completed w.r.t. $\mathcal{R}_h$ using the standard hypertableau algorithm; Figure 2(c) shows the resulting ABoxes $\mathcal{A}^{nm}_{fin}$, $\mathcal{A}^c_{fin}$, and $\mathcal{A}^d_{fin}$. Note that $C(a)$ and $C(d)$ in $\mathcal{A}$ are consistent with the axiom $\exists R.\top \sqsubseteq C$ from $\mathcal{T}_h$. Finally, $\mathcal{A}_{fin}$ is obtained by taking the union of $\mathcal{A}$, $\mathcal{A}^{nm}_{fin}$, $\mathcal{A}^c_{fin}$, and $\mathcal{A}^d_{fin}$, and adding $E(f)$; the latter is because $f$ is blocked by $c$ and $E(c) \in \mathcal{A}^c_{fin}$. The result is shown in Figure 2(d); clearly, no derivation rule of the standard hypertableau algorithm is applicable to $\mathcal{A}_{fin}$.

### 5.1.4 Hidden Ontology in DLs Between $\mathcal{ALC}$ and $\mathcal{ALCHIQ}$

The main limitation of the acyclicity condition from Definition 7 stems from the fact that we must anticipate all possible consequences of $\mathcal{T}_h$. Both the acyclicity conditions and the derivation rules from Table 4 can be simplified if the hidden ontology is known to be expressed in a description logic between $\mathcal{ALC}$ and $\mathcal{ALCHIQ}$.

- If $\mathcal{T}_h$ is known not to use cardinality restrictions, then we can omit rules (35)–(37) in the definition of $\mathsf{D}(\mathcal{R}_v, \mathcal{A}_v)$, and the $\approx$-cut rule in Table 4 is not required.

- If $\mathcal{T}_h$ is known not to use inverse roles, then we can omit rules (34), (36), and (37) in the definition of $\mathsf{D}(\mathcal{R}_v, \mathcal{A}_v)$, the $R^-$-cut rule is not required, and Conditions 3.2 and 3.3 can be removed from the $\approx$-cut rule.

- If $\mathcal{T}_h$ is known not to use role hierarchies, then we can omit rules (33) and (34) in the definition of $\mathsf{D}(\mathcal{R}_v, \mathcal{A}_v)$, the $R$-cut in Table 4 is not required, and the $R^-$-cut rule need only be applied if $R$ and $R'$ are the same.

These simplifications allow our approach to be applied to a wider range of visible ontologies. For example, consider the set $\mathcal{R}''_v$ consisting of HT-rules (40)–(44) and (47)–(48), for which we obtained a harmful cycle w.r.t. $\Gamma = \{C, R, U\}$, as shown in Figure 1(e). If $\mathcal{T}_h$ is known to be expressed in $\mathcal{ALCHI}$ (and so does not use cardinality restrictions), we can omit formulas of the form (35)–(37) in the definition of $\mathsf{D}(\mathcal{R}_v, \mathcal{A}_v)$; the ground atoms entailed by such $\mathsf{D}(\mathcal{R}_v, \mathcal{A}_v)$ are shown in Figure 3. This change makes $\mathcal{R}_v \cup \mathcal{A}_v$ acyclic w.r.t. $\Gamma$: we can now be sure that, for an arbitrary hidden TBox expressed in $\mathcal{ALCHI}$, no infinite $R$-chains need to be considered during reasoning with $\mathcal{R}_v$.

## 5.2 Practical Import-by-Query Algorithms

The algorithm presented in Section 5.1.3 is not suited for practical implementation because the derivation rules in Table 4 incur a huge amount of nondeterminism. In this section, we present practical import-by-query algorithms in which nondeterministic rules are replaced with "on demand" oracle calls, which makes the algorithms "more goal-oriented." Our algorithms make no assumptions about the kinds of symbols contained in $\Gamma$: both atomic concepts and roles can be shared.

### 5.2.1 Importing Horn Ontologies

In this section, we present a practical algorithm that applies when $\mathcal{T}_h$ is expressed in the Horn-$\mathcal{ALCHIQ}$ fragment of $\mathcal{ALCHIQ}$. It is well known that $\mathcal{T}_h$ can then be transformed into a set of Horn HT-rules. This allows us to eliminate the nondeterministic cut rules, use an ABox entailment oracle instead of an ABox satisfiability oracle, and define oracle query rules that deterministically "complete" the query ABox $\mathcal{A}$ with the missing assertions entailed by $\mathcal{T}_h \cup \mathcal{A}$. Such an algorithm issues oracle queries on demand, so it is goal oriented and thus more amenable to implementation.

**Definition 9.** *Let* $\mathcal{C}[\Gamma^{\mathcal{C}}, \mathcal{R}_v^{\mathcal{C}} \cup \mathcal{A}_v^{\mathcal{C}}, \mathcal{T}_h^{\mathcal{C}}]$ *be the class of inputs where* $\mathcal{R}_v^{\mathcal{C}} \cup \mathcal{A}_v^{\mathcal{C}}$ *is acyclic w.r.t.* $\Gamma^{\mathcal{C}}$, $\mathcal{R}_v^{\mathcal{C}}$ *is HT-safe w.r.t.* $\Gamma^{\mathcal{C}}$, *and* $\mathcal{T}_h^{\mathcal{C}}$ *is a Horn-$\mathcal{ALCHIQ}$ TBox. The Horn-$\mathcal{ALCHIQ}$ $\Omega^{\mathsf{e}}$-algorithm takes a triple* $\langle \Gamma, \mathcal{R}_v \cup \mathcal{A}_v, \mathcal{T}_h \rangle \in \mathcal{C}$ *and is obtained from Definition 8 by replacing the derivation rules from Table 4 with those in Table 5.*

Our algorithm is indeed an import-by-query algorithm with the same worst-case complexity as the algorithm for the non-Horn case.

**Theorem 12.** *The Horn-$\mathcal{ALCHIQ}$ $\Omega^{\mathsf{e}}$-algorithm is an import-by-query algorithm based on ABox entailment oracles for the class of inputs* $\mathcal{C}[\Gamma^{\mathcal{C}}, \mathcal{R}_v^{\mathcal{C}} \cup \mathcal{A}_v^{\mathcal{C}}, \mathcal{T}_h^{\mathcal{C}}]$ *from Definition 9. The algorithm can be implemented such that it runs in* N2ExpTime *in $N$, and the total number of oracle queries and the size of each query are both also at most exponential in $N$, where* $N = |\mathcal{R}_v \cup \mathcal{A}_v| + |\Gamma|$ *for the input $\mathcal{R}_v$, $\mathcal{A}_v$, and $\Gamma$.*

The proof of Theorem 12 is obtained by a modification of the one for Theorem 11 and is given in the appendix.

### 5.2.2 Import-by-Query in $\mathcal{EL}$

In this section, we present an import-by-query algorithm based on ABox entailment oracles that can handle the case when both $\mathcal{K}_v$ and $\mathcal{T}_h$ are expressed in $\mathcal{EL}$. In this setting, only Theorems 5 and 7 provide clues about features that hinder existence of an import-by-query algorithm. In particular, it is no longer necessary for $\mathcal{K}_v$ to be acyclic.

Our algorithm is again based on the hypertableau framework, so $\mathcal{K}_v$ is first converted into a set $\mathcal{R}_v$ of $\mathcal{EL}$-rules and a normalized ABox $\mathcal{A}_v$. Since $\mathcal{EL}$ does not allow for inverse roles or universal quantification, there is no danger of information propagating from a successor to a predecessor; therefore, we can relax the HT-safety condition as shown in Definition 10.

Table 5: Additional Derivation Rules for Horn KBs

| | | |
|---|---|---|
| $\Omega^{\mathsf{e}}$-concept | If | a connected component $\mathcal{A}'$ of $\mathcal{A}\|_\Gamma$, an individual $s$ in $\mathcal{A}'$, and an atomic concept $A \in \Gamma \cup \{\perp\}$ exist such that |
| | 1. | $s$ is not indirectly blocked in $\mathcal{A}$, |
| | 2. | $A(s) \notin \mathcal{A}$, and |
| | 3. | $\Omega^{\mathsf{e}}_{\mathcal{T}_h, \Gamma}(\mathcal{A}', A(s)) = \mathsf{t}$ |
| | then | $\mathcal{A}_1 := \mathcal{A} \cup \{A(s)\}$ |
| $\Omega^{\mathsf{e}}$-role | If | a connected component $\mathcal{A}'$ of $\mathcal{A}\|_\Gamma$, individuals $s$ and $t$ in $\mathcal{A}'$, and atomic roles $R, R' \in \Gamma$ exist such that |
| | 1. | neither $s$ not $t$ is indirectly blocked in $\mathcal{A}$, |
| | 2. | $R'(s,t) \in \mathcal{A}'$ or $R'(t,s) \in \mathcal{A}'$, |
| | 3. | $R(s,t) \notin \mathcal{A}$, and |
| | 4. | $\Omega^{\mathsf{e}}_{\mathcal{T}_h, \Gamma}(\mathcal{A}', R(s,t)) = \mathsf{t}$ |
| | then | $\mathcal{A}_1 := \mathcal{A} \cup \{R(s,t)\}$ |
| $\Omega^{\mathsf{e}}$-$\approx$ | If | a connected component $\mathcal{A}'$ of $\mathcal{A}\|_\Gamma$ and individuals $s$, $s_1$, and $s_2$ in $\mathcal{A}'$ exist such that |
| | 1. | none of $s$, $s_1$, and $s_2$ are indirectly blocked in $\mathcal{A}$, |
| | 2. | $s_1 \approx s_2 \notin \mathcal{A}$, |
| | 3. | atomic roles $R, R' \in \Gamma$ exist such that |
| | 3.1 | $\{R(s,s_1), R'(s,s_2)\} \subseteq \mathcal{A}$ or |
| | 3.2 | $\{R(s_1,s), R'(s_2,s)\} \subseteq \mathcal{A}$ or |
| | 3.3 | $\{R(s_1,s), R'(s,s_2)\} \subseteq \mathcal{A}$, and |
| | 4. | $\Omega^{\mathsf{e}}_{\mathcal{T}_h, \Gamma}(\mathcal{A}', s_1 \approx s_2) = \mathsf{t}$ |
| | then | $\mathcal{A}_1 := \mathcal{A} \cup \{s_1 \approx s_2\}$ |

**Definition 10.** *Let $\mathcal{R}_v$ be a set of $\mathcal{EL}$-rules, and let $\Gamma$ be a signature Then, $\mathcal{R}_v$ is $\mathcal{EL}$-safe w.r.t. $\Gamma$ if*

- *it satisfies Condition 1 from Definition 6, and*

- *for each rule $\varrho \in \mathcal{R}_v$ and each body atom of $\varrho$ of the form $R(x, y_i)$ with $R \in \Gamma$, the body of $\varrho$ contains an atom $B(y_i)$ such that $B \in \mathsf{safe}(\mathcal{R}_v, \Gamma)$.*

Our algorithm takes a set $\mathcal{R}_v$ of $\mathcal{EL}$-safe rules and a normalized ABox $\mathcal{A}_v$. It applies the standard $\mathcal{EL}$ hypertableau derivation rules; furthermore, just like the Horn-$\mathcal{ALCHIQ}$ $\Omega^{\mathsf{e}}$-algorithm from Section 5.2.1, it uses the oracle to complete the ABoxes encountered in a derivation with the relevant concept assertions.

**Definition 11.** *Let $\mathcal{C}[\Gamma^{\mathcal{C}}, \mathcal{R}_v^{\mathcal{C}} \cup \mathcal{A}_v^{\mathcal{C}}, \mathcal{T}_h^{\mathcal{C}}]$ be the class of inputs where $\mathcal{R}_v^{\mathcal{C}}$ is a set of $\mathcal{EL}$-rules that is $\mathcal{EL}$-safe w.r.t. $\Gamma^{\mathcal{C}}$, $\mathcal{A}_v^{\mathcal{C}}$ is a normalized ABox, and $\mathcal{T}_h^{\mathcal{C}}$ is an $\mathcal{EL}$ TBox. The $\mathcal{EL}$ $\Omega^{\mathsf{e}}$-algorithm takes a triple $\langle \Gamma, \mathcal{R}_v \cup \mathcal{A}_v, \mathcal{T}_h \rangle \in \mathcal{C}$ and is obtained by extending the algorithm from Definition 2 with the $\Omega^{\mathsf{e}}$-concept derivation rule shown in Table 5.*

Our algorithm is indeed an import-by-query algorithm, and it can be implemented to run in polynomial time, as shown by the following theorem. In contrast to algorithms we have presented thus far, the $\mathcal{EL}$ $\Omega^{\mathsf{e}}$-algorithm is both optimal and amenable to implementation.

**Theorem 13.** *The $\mathcal{EL}$ $\Omega^{\mathsf{e}}$-algorithm is an import-by-query algorithm based on ABox entailment oracles for the class of inputs $\mathcal{C}[\Gamma^{\mathcal{C}}, \mathcal{R}_v^{\mathcal{C}} \cup \mathcal{A}_v^{\mathcal{C}}, \mathcal{T}_h^{\mathcal{C}}]$ from Definition 11. The algorithm can be implemented such that it runs in* PTime *in $N$ with a polynomial number in $N$ of calls to $\Omega_{\mathcal{T}_h, \Gamma}^{\mathsf{e}}$, where $N = |\mathcal{R}_v \cup \mathcal{A}_v| + |\Gamma|$ for the input $\mathcal{R}_v$, $\mathcal{A}_v$, and $\Gamma$.*

The proof of Theorem 13 is rather technical and lengthy, and it is given in the appendix. The intuition behind the proof, however, is the same as in the case of the $\mathcal{ALCHIQ}$ $\Omega^{\mathsf{a}}$-algorithm, and the differences are due to the fact that the ABoxes produced by the $\mathcal{EL}$ $\Omega^{\mathsf{e}}$-algorithm have a specific shape.

## 6. A Lower Bound on the Complexity of Import-by-Query Reasoning

In this section we show that no import-by-query algorithm that handles the same input as our $\mathcal{ALCHIQ}$ $\Omega^{\mathsf{a}}$-algorithm can make only a polynomial number (in $|\Gamma|$) of queries each of which is of polynomial size (in $|\Gamma|$). This result applies already if $\Gamma$ contains only atomic concepts, so the only requirement for the $\mathcal{ALCHIQ}$ $\Omega^{\mathsf{a}}$-algorithm to be applicable is that the TBox of $\mathcal{K}_v$ is semantically modular w.r.t. $\Gamma$.

**Theorem 14.** *Let $\mathcal{C}[\Gamma^{\mathcal{C}}, \mathcal{K}_v^{\mathcal{C}}, \mathcal{T}_h^{\mathcal{C}}]$ be the class of inputs where $\Gamma^{\mathcal{C}}$ contains only atomic concepts, $\mathcal{K}_v^{\mathcal{C}}$ is an $\mathcal{ALCHIQ}$ knowledge base that is semantically modular in $\Gamma^{\mathcal{C}}$, and $\mathcal{T}_h^{\mathcal{C}}$ is an $\mathcal{ALCHIQ}$ TBox. Then, no import-by-query algorithm $\mathsf{ibq}^{\mathsf{a}}$ based on ABox satisfiability oracles for $\mathcal{L} = \mathcal{ALCHIQ}$ and $\mathcal{C}$ exists such that, for each input $\langle \Gamma, \mathcal{K}_v, \mathcal{T}_h \rangle \in \mathcal{C}$, the total number of oracle queries in all possible runs of $\mathsf{ibq}^{\mathsf{a}}[\mathcal{T}_h, \Gamma, \mathcal{L}]$ on $\mathcal{K}_v$, as well as the size of each query, are both polynomial in $|\Gamma|$.*

*Proof.* Assume that $\mathsf{ibq}^{\mathsf{a}}$ is an algorithm that satisfies the theorem's assumptions; then, integers $c_1$ and $c_2$ exist such that, for each input $\langle \Gamma, \mathcal{K}_v, \mathcal{T}_h \rangle \in \mathcal{C}$, the total number of oracle queries in all possible runs of $\mathsf{ibq}^{\mathsf{a}}[\mathcal{T}_h, \Gamma, \mathcal{L}]$ on $\mathcal{K}_v$ is smaller than or equal to $|\Gamma|^{c_1}$, and the maximal size of each query ABox is smaller than or equal to $|\Gamma|^{c_2}$.

We next construct a particular input in $\mathcal{C}$ for which we show that $\mathsf{ibq}^{\mathsf{a}}$ violates the above assumption. Let $k$ be an arbitrary integer such that $k^{c_1+c_2} < 2^k$; such $k$ exists since $c_1$ and $c_2$ are fixed. Let $\Gamma = \{A_1, \dots, A_k\}$ be arbitrary atomic concepts, and let $Z$, $B$, $C_1$, ..., $C_k$, $\overline{C}_1$, ..., $\overline{C}_k$ be atomic concepts not occurring in $\Gamma$. Then, we define $\mathcal{K}_v = \mathcal{T}_v \cup \mathcal{A}_v$ such that $\mathcal{A}_v = \{Z(a)\}$ and $\mathcal{T}_v$ contains the following axioms:

$$B \sqsubseteq \exists R.B \tag{50}$$

$$Z \sqsubseteq B \sqcap \overline{C}_1 \sqcap \dots \sqcap \overline{C}_k \tag{51}$$

$$C_j \sqcap \overline{C}_j \sqsubseteq \bot \qquad 1 \leq j \leq k \tag{52}$$

$$\top \sqsubseteq (\neg C_1 \sqcup \forall R.\overline{C}_1) \sqcap (\neg \overline{C}_1 \sqcup \forall R.C_1) \tag{53}$$

$$C_{j-1} \sqcap \exists R.\overline{C}_{j-1} \sqsubseteq (\neg C_j \sqcup \forall R.\overline{C}_j) \sqcap (\neg \overline{C}_j \sqcup \forall R.C_j) \quad 1 < j \leq k \tag{54}$$

$$\overline{C}_{j-1} \sqcup (C_{j-1} \sqcap \exists R.C_{j-1}) \sqsubseteq (\neg C_j \sqcup \forall R.C_j) \sqcap (\neg \overline{C}_j \sqcup \forall R.\overline{C}_j) \quad 1 < j \leq k \tag{55}$$

$$C_i \sqsubseteq A_i \quad 1 \leq i \leq k \tag{56}$$

$$\overline{C}_i \sqsubseteq \neg A_i \quad 1 \leq i \leq k \tag{57}$$

TBox $\mathcal{T}_v$ uses the well-known "integer counting" technique (Tobies, 2000). Consider an arbitrary model $I$ of $\mathcal{K}_v$. Domain elements of $I$ can be assigned integers between 0 and $2^k - 1$ by means of $2k$ atomic concepts $C_1, \ldots, C_k, \overline{C}_1, \ldots, \overline{C}_k$. Axiom (51) implies that $a^I \in (\overline{C}_k \sqcap \ldots \sqcap \overline{C}_1)^I$, which "initializes the counter" to 0. Axiom (50) ensures that $a^I$ is an origin of an infinite $R$-chain. Axioms (52) ensure that no domain element in this chain is labeled with both $C_j$ and $\overline{C}_j$. Axioms (53), (54), and (55) increment the counter over $R$. Finally, these axioms together with axioms (56) and (57) ensure that each possible number between 0 and $2^k - 1$ is assigned to some domain element of $I$ in the $R$-chain. Clearly, $\mathcal{T}_v$ is semantically modular w.r.t. $\Gamma$ since we can extend each interpretation of the symbols of $\Gamma$ to a model of $\mathcal{T}_v$ by interpreting the symbols not in $\Gamma$ with the empty set.

Let $\mathcal{T}_h^1 = \emptyset$, let $\mathcal{A}_1, \ldots, \mathcal{A}_m$ be the query ABoxes occurring in all possible runs of $\mathsf{ibq}^{\mathsf{a}}[\mathcal{T}_h^1, \Gamma, \mathcal{L}]$ on $\mathcal{K}_v$, and let $n$ be the maximal size of $\mathcal{A}_i$ for $1 \leq i \leq m$. By our assumptions, we have $m \leq k^{c_1}$ and $n \leq k^{c_2}$, which implies $m \times n = k^{c_1 + c_2} < 2^k$ due to the way we chose $k$. For each $1 \leq i \leq m$, let $\mathcal{A}_i'$ be the following ABox equivalent to $\mathcal{A}_i$:

- If $\mathcal{A}_i$ is unsatisfiable, then $\mathcal{A}_i' = \{\bot\}$.

- If $\mathcal{A}_i$ is satisfiable, let $\mathcal{A}_i'$ be an ABox that contains for each individual $s$ exactly one concept assertion of the form $D(s)$ where $D$ is in disjunctive normal form; that is, $D$ is expressed as a disjunction of concepts of the form $(\neg)A_1 \sqcap \ldots \sqcap (\neg)A_k$. Such $\mathcal{A}_i'$ can be obtained from $\mathcal{A}_i$ by applying de Morgan's laws.

Let $D_1, \ldots, D_\ell$ be all disjunctive concepts that occur in some satisfiable ABox $\mathcal{A}_i'$. Each $\mathcal{A}_i'$ contains at most $n$ such concepts, so $1 \leq \ell \leq m \times n$. Furthermore, let $U$ be the subset of $\{D_1, \ldots, D_\ell\}$ containing precisely those $D_i$ that have exactly one disjunct. Finally, let $S$ be a concept of the form $(\neg)A_1 \sqcap \ldots \sqcap (\neg)A_k$ that does not occur in $U$; such $S$ exists because $\ell \leq m \times n < 2^k$. Now let $\mathcal{T}_h^2$ be the following TBox:

$$\mathcal{T}_h^2 = \{S \sqsubseteq \bot\} \tag{58}$$

We next show that, for each $1 \leq j \leq \ell$, concept $D_j$ is satisfiable w.r.t. $\mathcal{T}_h^2$. The claim is trivial if $D_j$ does not contain $S$; otherwise, $D_j$ contains a disjunct $S' \neq S$, so an interpretation satisfying $\mathcal{T}_h^2$ and $D_j$ can be obtained by interpreting $S'$ as a nonempty set.

We next show that $\Omega^{\mathsf{a}}_{\mathcal{T}_h^1, \Gamma, \mathcal{L}}(\mathcal{A}_i') = \Omega^{\mathsf{a}}_{\mathcal{T}_h^2, \Gamma, \mathcal{L}}(\mathcal{A}_i')$ for each $1 \leq i \leq m$; since $\mathcal{A}_i$ and $\mathcal{A}_i'$ are equivalent, then $\Omega^{\mathsf{a}}_{\mathcal{T}_h^1, \Gamma, \mathcal{L}}(\mathcal{A}_i) = \Omega^{\mathsf{a}}_{\mathcal{T}_h^2, \Gamma, \mathcal{L}}(\mathcal{A}_i)$ as well. The statement clearly holds if $\mathcal{A}_i'$ is unsatisfiable, so assume that $\mathcal{A}_i'$ is satisfiable. Since $\mathcal{A}_i'$ consists of assertions of the form $D(s)$ where $D$ is satisfiable w.r.t. $\mathcal{T}_h^2$, an interpretation satisfying $\mathcal{A}_i' \cup \mathcal{T}_h^2$ can be obtained as a disjoint union of the interpretations satisfying each $D$.

By Proposition 1, the runs of $\mathsf{ibq}^{\mathsf{a}}[\mathcal{T}_h^1, \Gamma, \mathcal{L}]$ on $\mathcal{K}_v$ then coincide with the runs of $\mathsf{ibq}^{\mathsf{a}}[\mathcal{T}_h^2, \Gamma, \mathcal{L}]$ on $\mathcal{K}_v$; however, it is straightforward to see that $\mathcal{K}_v \cup \mathcal{T}_h^1$ is satisfiable, whereas $\mathcal{K}_v \cup \mathcal{T}_h^2$ is unsatisfiable, which is a contradiction. $\qquad\square$

# 7. Related Work

There is currently a growing interest in techniques for hiding parts of an ontology $\mathcal{T}_h$. One possible approach is to hide a subset $\Upsilon$ of the signature of $\mathcal{T}_h$ by first extracting from $\mathcal{T}_h$ an $\Upsilon$- *module* $\mathcal{M}_\Upsilon$—a subset of $\mathcal{T}_h$ that preserves all $\Upsilon$-consequences (i.e., all logical consequences formed using only the symbols in $\Upsilon$)—and then publishing the ontology $\mathcal{T}_h \setminus \mathcal{M}_\Upsilon$. In order to ensure that no sensitive information about $\Upsilon$ is being disclosed, the module $\mathcal{M}_\Upsilon$ should be *depleting* (Kontchakov, Pulina, Sattler, Schneider, Selmer, Wolter, & Zakharyaschev, 2009)—that is, ontology $\mathcal{T}_h \setminus \mathcal{M}_\Upsilon$ should be indistinguishable from the empty ontology w.r.t. $\Upsilon$-consequences. This approach ensures that no $\Upsilon$-consequences are disclosed to external applications and offers the additional advantage that one can reason over the union of $\mathcal{K}_v$ and $\mathcal{T}_h \setminus \mathcal{M}_\Upsilon$ using off-the-shelf DL reasoners. Finally, although determining whether a subset of an ontology is a depleting module for a signature is an undecidable problem for many DLs (and hence extraction of minimal depleting modules is often computationally infeasible), several practical techniques for extracting (not necessarily minimal) depleting modules are known (Cuenca Grau, Horrocks, Kazakov, & Sattler, 2008).

An important disadvantage of this approach is that the module $\mathcal{M}_\Upsilon$ may also contain relevant information that is not sensitive (e.g., $\mathcal{M}_\Upsilon$ may entail consequences about symbols $\Gamma$ not in $\Upsilon$) and hence the union of $\mathcal{K}_v$ (which may use symbols from $\Gamma$) and $\mathcal{T}_h \setminus \mathcal{M}_\Upsilon$ may not contain enough information to answer relevant queries. Furthermore, by adopting this approach, the vendor of $\mathcal{T}_h$ would distribute a subset of the axioms of $\mathcal{T}_h$, which may allow competitors to plagiarize parts of $\mathcal{T}_h$. Finally, the published axioms might mention symbols in $\Upsilon$ (even if they do not entail any $\Upsilon$-consequence) and external applications would be aware of the presence of those symbols in the ontology.

Some of these drawbacks can be overcome by publishing an $\Upsilon$-*interpolant* of $\mathcal{T}_h$—an ontology that contains no symbols from $\Upsilon$ and that coincides with $\mathcal{T}_h$ on all logical consequences formed using the symbols not in $\Upsilon$ (Konev et al., 2009; Wang et al., 2009, 2008; Lutz & Wolter, 2011; Nikitina, 2011). In contrast to the module extraction approach, publishing an interpolant ensures that the sensitive information in $\mathcal{T}_h$ (i.e., the information about the symbols from $\mathcal{T}_h$ not mentioned in the interpolant) is not exposed in any way to external applications; furthermore, interpolants preserve all consequences of symbols not in $\Upsilon$. Similarly to the module extraction approach, using interpolation has the additional advantage that the developers of $\mathcal{K}_v$ can reason over the union of $\mathcal{K}_v$ and the interpolant using off-the-shelf DL reasoners.

The interpolation approach may, however, have several drawbacks. First, an interpolant may exist only if $\mathcal{T}_h$ is expressed in a relatively weak DL and satisfies certain syntactic conditions (Konev et al., 2009). In contrast, import-by-query is often possible even if an interpolant of $\mathcal{T}_h$ for the signature of interest does not exist.

Second, although interpolants preserve logical consequences formed using symbols not in $\Upsilon$, they are not *robust under replacement* (Sattler et al., 2009)—that is, the union of $\mathcal{K}_v$ and an $\Upsilon$-interpolant of $\mathcal{T}_h$ is not guaranteed to yield the same consequences as $\mathcal{T}_h \cup \mathcal{K}_v$ for a query $q$ involving no symbols from $\Upsilon$. For example, given $\Upsilon = \{R\}$ and $\mathcal{T}_h = \{A \sqsubseteq \exists R.B\}$, the empty ontology is an $\Upsilon$-interpolant (it preserves all consequences of the form $C \sqsubseteq D$ with $C$ and $D$ arbitrary boolean concepts over the signature $\{A, B\}$); however, for $\mathcal{K}_v = \{B \sqsubseteq \bot\}$ we have that $\mathcal{K}_v \cup \mathcal{T}_h$ entails the consequence $A \sqsubseteq \bot$, whereas the union of $\mathcal{K}_v$ and the (empty) interpolant does not. Thus, once an interpolant has been published, it cannot be imported into $\mathcal{K}_v$ with the guarantee that all relevant consequences will be preserved, unless suitable restrictions are imposed to $\mathcal{K}_v$.

Finally, an $\Upsilon$-interpolant of $\mathcal{T}_h$ can be exponentially larger than $\mathcal{T}_h$, and may reveal more information than what is strictly needed. Although import-by-query algorithms can also formulate in the worst-case exponentially many queries to the oracle, our algorithms may limit the flow of irrelevant information from $\mathcal{T}_h$ to $\mathcal{K}_v$, especially if $\mathcal{T}_h$ is expressed in a Horn DL, in which case our import-by-query algorithms issue queries "on demand." For example, for $\Gamma = \{R, C\}$, $\Upsilon = \emptyset$, $\mathcal{K}_v = \{A \sqsubseteq \exists R.B, B \sqsubseteq C\}$ and $\mathcal{T}_h = \{\exists R.\exists R.C \sqsubseteq C\}$, the $\Upsilon$-interpolant is equal to $\mathcal{T}_h$ and thus publishing the interpolant reveals entire contents of $\mathcal{T}_h$. In contrast, our import-by-query algorithm for $\mathcal{EL}$ would not reveal any positive information about $\mathcal{T}_h$, as it would only disclose the fact that an ABox of the form $\{R(a, b), C(b)\}$ is satisfiable w.r.t. $\mathcal{T}_h$.

The idea of accessing an ontology through an oracle is similar in spirit to the proposal by Calvanese, De Giacomo, Lembo, Lenzerini, and Rosati (2004) for query answering in a peer-to-peer setting. The authors consider the problem of answering a conjunctive query $q$ over KBs $\mathcal{K}_v$ and $\mathcal{K}_h$ and mappings $M$ by reformulating $q$ as queries that can be evaluated over $\mathcal{K}_v$ and $\mathcal{K}_h$ in isolation. The query reformulation algorithm accesses only $\mathcal{K}_v$ and $M$, so $q$ can be answered using an oracle for $\mathcal{K}_h$. In this setting, however, the focus is on the reuse of data, rather than schema. Since a satisfiable $\mathcal{K}_h$ cannot affect the subsumption of concepts in $\mathcal{K}_v$, the results by Calvanese et al. (2004) are not applicable to schema reasoning.

## 8. Conclusion

In this paper, we have proposed and studied the import-by-query framework. Our results provide a flexible way for ontology designers to ensure selective access to their ontologies. Our framework thus provides key theoretical insights into the issues surrounding ontology privacy. Furthermore, we believe our algorithms to be practicable when applied to Horn ontologies; thus, our results provide a starting point for the development of practical import-by-query systems.

The problem of import-by-query is novel, and we see many open questions. For example, a problem that is relevant to both theory and practice is to allow the hidden ontology to selectively export data and not just schema statements.

## Acknowledgments

This is an extended version of the paper "Import-by-Query: Ontology Reasoning under Access Limitations" by Bernardo Cuenca Grau, Boris Motik, and Yevgeny Kazakov published at IJCAI 2009 and the paper "Pushing the Limits of Reasoning over Ontologies with Hidden Content" by Bernardo Cuenca Grau and Boris Motik published at KR 2010.

This research has been supported by the Royal Society and the EPSRC projects ExODA (EP/H051511/1) and HermiT (EP/F065841/1).

# Appendix A. Proof of Theorem 11

We will use the following definitions and intermediate results to prove the theorem.

**Definition 12.** *An ABox $\mathcal{A}$ is an* HT-ABox *if all of its assertions satisfy the following conditions, for $B$ an atomic or a negated atomic concept, $S$ a role, $R$ an atomic role, $a$ and $b$ named individuals, $s$ an individual, and $i$ and $j$ integers.*

1. *Each concept assertion in $\mathcal{A}$ is of the form $B(s)$ or $\geq n\, S.B(s)$.*

2. *Each role assertion in $\mathcal{A}$ is of the form $R(a, b)$, $R(s, s.i)$, or $R(s.i, s)$.*

3. *If an individual $s.i$ occurs in an assertion in $\mathcal{A}$, then $\mathcal{A}$ contains a role assertion of the form $R(s, s.i)$ or $R(s.i, s)$.*

4. *Each equality in $\mathcal{A}$ is of the form $s.i \approx s.j$, $s.i.j \approx s$, $s \approx s$, or $a \approx b.i$.*

*Furthermore, an* extended HT-ABox *$\mathcal{A}$ is additionally allowed to contain assertions of the form $R(s, s)$ and $s.i \approx s$.*

**Lemma 1.** *Let $\mathcal{R}$ be a set of HT-rules and let $\mathcal{A}$ be an ABox. Then, each ABox labeling a node of a derivation for $\mathcal{R}$ and $\mathcal{A}$ is an HT-ABox.*

*Proof.* The proof is a straightforward modification of the proof of Lemma 4 by Motik et al. (2009), which are due the following observations: since HT-rules do not allow for atoms of the form $R(x, x)$ in the head, one cannot derive atoms of the form $R(s, s)$; this, in turn, guarantees that one cannot derive equalities of the form $s.i \approx s$. □

**Lemma 2.** *(Motik et al., 2009, Lemma 6) Let $\mathcal{R}$ be a set of HT-rules and let $\mathcal{A}$ be a clash-free extended HT-ABox not containing indirectly blocked individuals. If no derivation rule is applicable to $\mathcal{R}$ and $\mathcal{A}$, then $\mathcal{R} \cup \mathcal{A}$ is satisfiable.*

**Definition 13.** *The* weakened pairwise anywhere blocking, *abbreviated* w-blocking, *is the same as in Definition 1, with the difference that the following condition is used instead of $\mathcal{L}_{\mathcal{A}}(s') = \mathcal{L}_{\mathcal{A}}(t')$:*

> *For each HT-rule $\varrho \in \mathcal{R}$ containing a body atom of the form $R(x, y)$ or $R(y, x)$ with $R$ an atomic role such that $R \in \mathcal{L}_{\mathcal{A}}(s, s') \cup \mathcal{L}_{\mathcal{A}}(s', s)$, and for each atomic concept $A$ occurring in $\varrho$, we have $A \in \mathcal{L}_{\mathcal{A}}(s')$ if and only if $A \in \mathcal{L}_{\mathcal{A}}(t')$.*

**Lemma 3.** *Lemma 2 holds even if the derivation for $\mathcal{R}$ and $\mathcal{A}$ uses w-blocking.*

*Proof (Sketch).* Let $\mathcal{A}'$ be an ABox labeling a leaf of a derivation for $\mathcal{R}$ and $\mathcal{A}$; let $s$ be an individual that is blocked in $\mathcal{A}'$ by $t$ by w-blocking; and let $s'$ and $t'$ be the parents of $s$ and $t$. For the proof by Motik et al. (2009, Lemma 6) to hold, we must show that no HT-rule is applicable to an interpretation obtained by unraveling $\mathcal{A}'$. Let $\varrho \in \mathcal{R}$ be an arbitrary HT-rule. If $\varrho$ does not contain in the body a role atom with a role $R \in \mathcal{L}_{\mathcal{A}}(s, s') \cup \mathcal{L}_{\mathcal{A}}(s', s)$, then the *Hyp*-rule cannot be applied to $\varrho$ with mapping $\sigma(x) = s$. Furthermore, if $\varrho$ does not contain an atomic concept $A$, then the fact that $A \in \mathcal{L}_{\mathcal{A}}(s')$ but $A \notin \mathcal{L}_{\mathcal{A}}(t')$ or vice versa cannot affect the applicability of $\varrho$. Thus, by a straightforward modification of the proof by Motik et al. (2009, Lemma 6), we can construct a model for $\mathcal{A}$ and $\mathcal{R}$ by unraveling $\mathcal{A}'$. □

It is straightforward to see that the derivation rules in Table 4 do not invalidate Lemma 1—that is, given an HT-ABox, they always produce an HT-ABox.

## A.1 Termination

We first show that the logical consequences of the datalog program $D(\mathcal{R}_v, \mathcal{A}_v)$ from Definition 7 "overestimate" the ABoxes produced by the hypertableau algorithm; that is, we show that each ABox $\rho(t)$ labeling a derivation node can be homomorphically embedded into the set of ground facts entailed by $D(\mathcal{R}_v, \mathcal{A}_v)$.

If $s' = s.i$ and either $R(s, s') \in \rho(t)$ or $R(s', s) \in \rho(t)$ with $R \in \Gamma$, we say that $s'$ is a $\Gamma$-*successor* of $s$.

**Lemma 4.** *Let $\mathcal{R}_v$ be a set of HT-rules, let $\mathcal{A}_v$ be an ABox, let $\Gamma$ be a signature, let $D(\mathcal{R}_v, \mathcal{A}_v)$ be as given in Definition 7, let $\mathcal{T}_h$ be an $\mathcal{ALCHIQ}$ TBox, and let $(T, \rho)$ be a derivation for $\langle \Gamma, \mathcal{R}_v \cup \mathcal{A}_v, \mathcal{T}_h \rangle$. Then, for each derivation node $t \in T$, a mapping $\mu$ from the individuals in $\rho(t)$ to the individuals in $D(\mathcal{R}_v, \mathcal{A}_v)$ exists satisfying all of the following properties for all individuals $s$ and $s'$ occurring in $\rho(t)$:*

1. *$A(s) \in \rho(t)$ with $A$ an atomic concept implies $D(\mathcal{R}_v, \mathcal{A}_v) \models A(\mu(s))$.*

2. *$R(s, s') \in \rho(t)$ implies $D(\mathcal{R}_v, \mathcal{A}_v) \models R(\mu(s), \mu(s'))$.*

3. *If $s'$ is a successor of $s$ in $\rho(t)$, then $D(\mathcal{R}_v, \mathcal{A}_v) \models Succ(\mu(s), \mu(s'))$.*

4. *If $s'$ is a $\Gamma$-successor of $s$ in $\rho(t)$, then $D(\mathcal{R}_v, \mathcal{A}_v) \models \Gamma\text{-}Desc(\mu(s), \mu(s'))$.*

5. *If $\geq n\, R.C(s) \in \rho(t)$ with $R$ a possibly inverse role, then the following conditions hold:*

    (a) *if $C$ is an atomic concept, then $D(\mathcal{R}_v, \mathcal{A}_v) \models C(v_C)$;*

    (b) *$D(\mathcal{R}_v, \mathcal{A}_v) \models \mathsf{ar}(R, \mu(s), v_C)$;*

    (c) *$D(\mathcal{R}_v, \mathcal{A}_v) \models Succ(\mu(s), v_C)$; and*

    (d) *if $R \in \Gamma$, then $D(\mathcal{R}_v, \mathcal{A}_v) \models \Gamma\text{-}Desc(\mu(s), v_C)$.*

6. *If $s \approx s' \in \rho(t)$, then $D(\mathcal{R}_v, \mathcal{A}_v) \models \mu(s) \approx \mu(s')$.*

7. *If $s$ is an unnamed individual in $\rho(t)$, an atomic concept $A \in \mathsf{sig}(\mathcal{R}_v) \cup \mathsf{sig}(\mathcal{A}_v)$ exists such that $\mu(s) = v_A$ or $\mu(s) = v_{\neg A}$.*

*Proof.* We prove the lemma by induction on the structure of the derivation. For $\epsilon \in T$ the root node of the derivation, let $\mu$ map each individual in $\mathcal{A}_v$ to itself. ABox $\rho(\epsilon) = \mathcal{A}_v$ trivially satisfies Properties 3, 4, and 7 since $\mathcal{A}_v$ contains only named individuals. Properties 5 and 6 also hold trivially because $\rho(\epsilon)$ is a normalized ABox and hence it does not contain assertions of the form $\geq n\, R.C(s)$ or of the form $s \approx s'$. Finally, Properties 1 and 2 hold because $\rho(\epsilon) \subseteq D(\mathcal{R}_v, \mathcal{A}_v)$.

For the induction step, assume that, for some derivation node $t \in T$, ABox $\rho(t)$ satisfies the claim for some mapping $\mu$. For each child node $t'$ of $t$ in $T$, we consider the possible ways $\rho(t')$ can be derived from $\rho(t)$.

- $\Omega^{\mathsf{a}}$-rule: All properties hold trivially for $\rho(t')$ and $\mu$.

- **$A$-cut:** All properties hold trivially for $\rho(t')$ and $\mu$ except for Property 1 in case $\rho(t') = \rho(t) \cup \{A(s)\}$ with $A \in \Gamma$. If $s$ is a named individual in $\rho(t)$, then $s$ occurs in $\mathcal{A}_v$ and Property 1 holds because $\mathsf{D}(\mathcal{R}_v, \mathcal{A}_v)$ contains the assertion $A(s)$ for each $A \in \Gamma$ and each $s$ occurring in $\mathcal{A}_v$. If $s$ is unnamed, then $s$ is the successor of some individual $s'$ in $\rho(t)$; by the induction hypothesis (Property 3) we have $\mathsf{D}(\mathcal{R}_v, \mathcal{A}_v) \models \mathsf{Succ}(\mu(s'), \mu(s))$; however, $\mathsf{D}(\mathcal{R}_v, \mathcal{A}_v)$ contains the formula (32) for each $A \in \Gamma$, so we have $\mathsf{D}(\mathcal{R}_v, \mathcal{A}_v) \models A(\mu(s))$, as required.

- **$R$-cut:** All properties hold trivially for $\rho(t')$ and $\mu$ except for Property 2 in case $\rho(t') = \rho(t) \cup \{R(s, s')\}$ with $R \in \Gamma$. By Condition 2 of $R$-cut we have $R'(s, s') \in \rho(t)$ for some atomic role $R' \in \Gamma$, so we have $\mathsf{D}(\mathcal{R}_v, \mathcal{A}_v) \models R'(\mu(s), \mu(s'))$ by the induction assumption. Since $R, R' \in \Gamma$ and $\mathsf{D}(\mathcal{R}_v, \mathcal{A}_v)$ contains formulae (33) for all roles in $\Gamma$, we have $\mathsf{D}(\mathcal{R}_v, \mathcal{A}_v) \models R(\mu(s), \mu(s'))$, so $\rho(t')$ satisfies Property 2 for $\mu$.

- **$R^-$-cut:** All properties hold trivially for $\rho(t')$ and $\mu$ except for Property 2 in case $\rho(t') = \rho(t) \cup \{R(s', s)\}$ with $R \in \Gamma$. By Condition 2 of $R^-$-cut we have $R'(s', s) \in \rho(t)$ for some atomic role $R' \in \Gamma$, so we have $\mathsf{D}(\mathcal{R}_v, \mathcal{A}_v) \models R'(\mu(s), \mu(s'))$ by the induction assumption. Since $R, R' \in \Gamma$ and $\mathsf{D}(\mathcal{R}_v, \mathcal{A}_v)$ contains formulae (34) for all roles in $\Gamma$, we have $\mathsf{D}(\mathcal{R}_v, \mathcal{A}_v) \models R(\mu(s'), \mu(s))$, so $\rho(t')$ satisfies Property 2 for $\mu$.

- **$\bot$-rule:** All properties hold trivially for $\rho(t')$ and $\mu$.

- **$\geq$-rule:** Assume that $\rho(t')$ is defined as follows, where $\geq n\, R.C(s) \in \rho(t)$, $s_i$ are fresh successors of $s$, and $C$ is a possibly negated atomic concept:

$$\rho(t') = \rho(t) \cup \{\, \mathsf{ar}(R, s, s_i),\ \ C(s_i)\ \mid 1 \leq i \leq n \,\} \cup \{\, s_i \not\approx s_j\ \mid 1 \leq i < j \leq n\,\}$$

Let $\mu' = \mu \cup \{s_i \mapsto v_C \mid 1 \leq i \leq n\}$. Properties 5 and 6 hold trivially for $\rho(t')$ and $\mu'$, and it is obvious that Property 7 holds as well. Hence, we focus on showing Properties 1—4. For Property 1, assume that $C$ is an atomic concept; since Property 5(a) holds for $\rho(t)$ and $\mu$ by the induction assumption, we have $\mathsf{D}(\mathcal{R}_v, \mathcal{A}_v) \models C(v_C)$, as required. For Property 2, since Property 5(b) holds for $\rho(t)$ and $\mu$ by the induction assumption, we have $\mathsf{D}(\mathcal{R}_v, \mathcal{A}_v) \models \mathsf{ar}(R, \mu(s), v_C)$, as required. For Property 3, since Property 5(c) holds for $\rho(t)$ and $\mu$ by the induction assumption, we have $\mathsf{D}(\mathcal{R}_v, \mathcal{A}_v) \models \mathsf{Succ}(\mu(s), v_C)$, so Property 3 holds for $\rho(t')$ and $\mu'$. For Property 4, assume that $R \in \Gamma$; Property 5(d) holds for $\rho(t)$ and $\mu$ by the induction assumption, we have $\mathsf{D}(\mathcal{R}_v, \mathcal{A}_v) \models \Gamma\text{-}\mathsf{Desc}(\mu(s), v_C)$, so Property 4 holds for $\rho(t')$ and $\mu'$.

- **$Hyp$-rule:** Assume that $\rho(t') = \rho(t) \cup \{\alpha\}$ for $\alpha$ the head atom of an HT-rule $\varrho$ of the form (2). Properties 3, 4, and 7 hold trivially for $\rho(t')$ and $\mu$, so we focus on the remaining properties. By Condition 2 of the $Hyp$-rule, $\rho(t)$ contains individuals $s, s_1, \ldots, s_n$ such that the statements from the left column from the following table holds. But then, by the induction assumption, the statements from the right column hold as well.

$$
\begin{array}{lcl}
A_i(s) \in \rho(t) & \Rightarrow & \mathsf{D}(\mathcal{R}_v, \mathcal{A}_v) \models A_i(\mu(s)) \\
R_{ij}(s, s_i) \in \rho(t) & \Rightarrow & \mathsf{D}(\mathcal{R}_v, \mathcal{A}_v) \models R_{ij}(\mu(s), \mu(s_i)) \\
S_{ij}(s_i, s) \in \rho(t) & \Rightarrow & \mathsf{D}(\mathcal{R}_v, \mathcal{A}_v) \models S_{ij}(\mu(s_i), \mu(s)) \\
B_{ij}(s_i) \in \rho(t) & \Rightarrow & \mathsf{D}(\mathcal{R}_v, \mathcal{A}_v) \models B_{ij}(\mu(s_i))
\end{array}
$$

For the HT-rule $\varrho$, the datalog program contains the rule (31). Thus, the statements from the following table then hold as well:

$$\mathsf{D}(\mathcal{R}_v, \mathcal{A}_v) \models \mathsf{tt}(C_i(\mu(s)))$$
$$\mathsf{D}(\mathcal{R}_v, \mathcal{A}_v) \models R'_{ij}(\mu(s), \mu(s_i))$$
$$\mathsf{D}(\mathcal{R}_v, \mathcal{A}_v) \models S'_{ij}(\mu(s_i), \mu(s))$$
$$\mathsf{D}(\mathcal{R}_v, \mathcal{A}_v) \models D_{ij}(\mu(s_i))$$
$$\mathsf{D}(\mathcal{R}_v, \mathcal{A}_v) \models \mu(s_i) \approx \mu(s_j)$$

Consequently, Properties 2 and 6 clearly hold; Property 1 also holds since for an atomic concept atom $\alpha$ we have $\mathsf{tt}(\alpha) = \alpha$. To show Property 5, assume that $C_i(\mu(s))$ is of the form $\geq n\, R.C(\mu(s))$, so

$$\mathsf{tt}(C_i(\mu(s))) = \mathsf{ar}(R, \mu(s), v_C) \wedge \mathsf{tt}(C(v_C)) \wedge \mathrm{Succ}(\mu(s), v_C).$$

Then, the following holds:

$$\mathsf{D}(\mathcal{R}_v, \mathcal{A}_v) \models \mathsf{ar}(R, \mu(s), v_C)$$
$$\mathsf{D}(\mathcal{R}_v, \mathcal{A}_v) \models \mathsf{tt}(C(v_C))$$
$$\mathsf{D}(\mathcal{R}_v, \mathcal{A}_v) \models \mathrm{Succ}(\mu(s), v_C)$$

Thus, Properties (5a), (5b), and (5c) hold. Finally, if $R \in \Gamma$, then Property (5d) holds because the datalog program entails assertion $\mathrm{Succ}(\mu(s), v_C)$, and it contains formulae (38) and (34) for all roles in $\Gamma$.

- $\approx$-cut rule: Assume that $\rho(t') = \rho(t) \cup \{\alpha\}$ with $\alpha$ an assertion of the form $s_1 \approx s_2$ or $s_1 \not\approx s_2$. Then, $\rho(t)$ trivially satisfies Properties 1–5 and 7 for $\mu$. Property 6 also holds trivially if $\alpha$ is of the form $s_1 \not\approx s_2$, so assume that $\alpha$ of the form $s_1 \approx s_2$. By the preconditions of the $\approx$-cut rule, an individual $s$ in $\rho(t)$ and atomic roles $R, R' \in \Gamma$ exist such that

$$\{\, R(s, s_1),\ R'(s, s_2)\,\} \subseteq \rho(t) \text{ or}$$
$$\{\, R(s_1, s),\ R'(s_2, s)\,\} \subseteq \rho(t) \text{ or}$$
$$\{\, R(s_1, s),\ R'(s, s_2)\,\} \subseteq \rho(t).$$

By the induction hypothesis (Property 2), then

$$\mathsf{D}(\mathcal{R}_v, \mathcal{A}_v) \models \{\, R(\mu(s), \mu(s_1)),\ R'(\mu(s), \mu(s_2))\,\} \quad \text{or}$$
$$\mathsf{D}(\mathcal{R}_v, \mathcal{A}_v) \models \{\, R(\mu(s_1), \mu(s)),\ R'(\mu(s_2), \mu(s))\,\} \quad \text{or}$$
$$\mathsf{D}(\mathcal{R}_v, \mathcal{A}_v) \models \{\, R(\mu(s_1), \mu(s)),\ R'(\mu(s), \mu(s_2))\,\}.$$

But then, since the datalog program contains formulas (35)–(37) for all roles in $\Gamma$, we have $\mathsf{D}(\mathcal{R}_v, \mathcal{A}_v) \models \mu(s_1) \approx \mu(s_2)$, as required.

- $\approx$-rule: Assume that $\rho(t') = \mathsf{merge}_{\rho(t)}(s \to s')$. Then, by Conditions 1 and 2 of the $\approx$-rule, $s \approx s' \in \rho(t)$ with $s \neq s'$. Furthermore, by the induction assumption, we have $\mathsf{D}(\mathcal{R}_v, \mathcal{A}_v) \models \mu(s) \approx \mu(s')$. Since merging merely replaces $s$ with $s'$, by the semantics of equality $\rho(t')$ satisfies all the required properties. $\qquad \square$

We next use Lemma 4 to prove that the length of chains of role assertions involving a role in $\Gamma$ is bounded.

**Lemma 5.** *Let $\mathcal{R}_v$, $\mathcal{A}_v$, $\Gamma$, $\mathsf{D}(\mathcal{R}_v, \mathcal{A}_v)$, and $(T, \rho)$ be as in Lemma 4 with the additional restriction that $\mathcal{R}_v \cup \mathcal{A}_v$ is acyclic w.r.t. $\Gamma$. Let $N$ be the number of individuals of the form $v_C$ occurring in $\mathsf{D}(\mathcal{R}_v, \mathcal{A}_v)$, let $t \in T$ be an arbitrary derivation node of $(T, \rho)$, and let $s_1, \dots, s_\ell$ be unnamed individuals occurring in $\rho(t)$ such that $s_{i+1}$ is a $\Gamma$-successor of $s_i$ for each $1 \leq i < \ell$. Then, $\ell \leq N$.*

*Proof.* Assume that, for some integer $\ell > N$, unnamed individuals $s_1, \dots, s_\ell$ satisfying the conditions of this lemma exist, and let $\mu$ be a mapping satisfying Lemma 4. By Property 7 in Lemma 4, for each $1 \leq i \leq \ell$ we have $\mu(s_i) = v_{C_i}$ for some $C_i$ (because each $s_i$ is unnamed). Furthermore, by Property 4 in Lemma 4, we also have $\mathsf{D}(\mathcal{R}_v, \mathcal{A}_v) \models \Gamma\text{-Desc}(\mu(s_i), \mu(s_{i+1}))$ for each $1 \leq i < \ell$. But then, since $\ell > N$ and predicate $\Gamma\text{-Desc}(x, y)$ is axiomatized as transitive by formula (39) in $\mathsf{D}(\mathcal{R}_v, \mathcal{A}_v)$, we clearly obtain a harmful cycle, which is a contradiction. $\qquad\square$

We are now ready to prove our main claim.

**Lemma 6** (Termination). *Let $\mathcal{R}_v$, $\mathcal{A}_v$, $\Gamma$, $\mathsf{D}(\mathcal{R}_v, \mathcal{A}_v)$, $N$, and $(T, \rho)$ be as in Lemma 5. Then, $(T, \rho)$ is finite.*

*Proof.* Let the *depth* of an individual $s$ be the number of its ancestors, and let $c$ and $r$ be the numbers of atomic concepts and roles, respectively, occurring in $\mathcal{R}_v$ and $\mathcal{A}_v$; finally, let $\wp = (2^{2cr} + 1)(N + 1) + 1$. Consider now an arbitrary derivation node $t \in T$. Let $s$ be an individual in $\rho(t)$ of depth $i(N + 1) + 1$. By a simple induction on $i$, one can show that $s$ has at least $i$ ancestors that are blocking-relevant. The induction base is straightforward for $i = 0$; furthermore, the induction step holds because, by Lemma 5 and the fact that $\rho(t)$ is an HT-ABox, the depth of the nearest blocking-relevant ancestor of $s$ can be at most $N + 1$ less than the depth of $s$. Thus, each individual $s$ of depth $\wp$ has at least $2^{2cr} + 1$ blocking-relevant ancestors; since there are at most $2^{2cr}$ possible concept and role labelings for an individual and its predecessor, one of the blocking ancestors of $s$ is blocked due to the definition of blocking; hence, $s$ is either directly or indirectly blocked in $\rho(t)$. The rest of the proof of our claim is then analogous to the proof of Lemma 7 by Motik et al. (2009). $\qquad\square$

### A.2 Soundness

**Lemma 7** (Soundness). *Let $\mathcal{R}_v$ be a set of HT-rules, let $\mathcal{T}_h$ be an $\mathcal{ALCHIQ}$ TBox, let $\mathcal{A}$ be an ABox such that $\mathcal{R}_v \cup \mathcal{T}_h \cup \mathcal{A}$ is satisfiable, and let $\mathcal{A}_1, \dots, \mathcal{A}_n$ be the ABoxes obtained by applying a derivation rule from Table 2 or 4 to $\mathcal{R}_v$ and $\mathcal{A}$. Then, $\mathcal{R}_v \cup \mathcal{T}_h \cup \mathcal{A}_i$ is satisfiable for some $1 \leq i \leq n$.*

*Proof.* Let $I$ be a model of $\mathcal{R}_v \cup \mathcal{T}_h \cup \mathcal{A}$, and let us consider the possible derivation rules that derive $\mathcal{A}_1, \dots, \mathcal{A}_n$. The cases for the *Hyp*-, $\geq$-, $\approx$-, and $\perp$-rule are the same as in the proof by Motik et al. (2009, Lemma 5). Furthermore, by the law of excluded middle of first-order logic, the claim is true for $A$, $R$-cut, $R^-$-cut and $\approx$-cut rules. Assume that the $\Omega^a$-rule derives $\perp$—that is, that $\mathcal{T}_h \cup \mathcal{A}'$ is unsatisfiable for some connected component $\mathcal{A}'$ of $\mathcal{A}|_\Gamma$. But then, since $\mathcal{A}' \subseteq \mathcal{A}|_\Gamma \subseteq \mathcal{A}$, by the monotonicity of first-order logic $\mathcal{R}_v \cup \mathcal{T}_h \cup \mathcal{A}$ is unsatisfiable as well, which is a contradiction. $\qquad\square$

### A.3 Completeness

Definition 14 and Proposition 5 show that the part of a model that is implied by $\mathcal{T}_h$ can always be extended to a model of $\mathcal{R}_v$. We say that an assertion is *atomic* if it is of the form $A(a)$ with $A$ an atomic concept, or $R(a, b)$ with $R$ an atomic role.

**Definition 14.** *Let $\Gamma$ be a signature, let $\mathcal{R}_v$ be a set of HT-rules, and let $\mathcal{A}$ be a nonempty clash-free ABox containing exactly one individual such that $\mathsf{sig}(\mathcal{A}) \subseteq \Gamma$. An ABox $\mathcal{A}'$ is an $\mathcal{R}_v$-extension of $\mathcal{A}$ w.r.t. $\Gamma$ if the following conditions hold:*

1. *$\mathcal{A}'$ contains exactly one individual, $\mathcal{A}'|_\Gamma = \mathcal{A}$, and $\mathsf{sig}(\mathcal{A}') \subseteq \mathsf{sig}(\mathcal{R}_v)$;*

2. *no derivation rule from Table 2 is applicable to $\mathcal{A}'$ and $\mathcal{R}_v$; and*

3. *$\mathcal{A}'$ does not contain an assertion of the form $A(s)$ with $A \in \mathsf{safe}(\mathcal{R}_v, \Gamma)$.*

**Proposition 5.** *For each $\Gamma$, $\mathcal{R}_v$, and $\mathcal{A}$ as in Definition 14 where $\mathcal{R}_v$ is additionally HT-safe, at least one $\mathcal{R}_v$-extension $\mathcal{A}'$ of $\mathcal{A}$ w.r.t. $\Gamma$ exists.*

*Proof.* Let $s$ be the individual occurring in $\mathcal{A}$, and let $I = (\triangle^I, \cdot^I)$ be the interpretation for the symbols in $\Gamma$ defined as follows:

$$\triangle^I = \{s\} \qquad A^I = \left\{ \begin{array}{ll} \{s\} & \text{if } A(s) \in \mathcal{A} \\ \emptyset & \text{otherwise} \end{array} \right. \qquad R^I = \left\{ \begin{array}{ll} \{\langle s, s\rangle\} & \text{if } R(s, s) \in \mathcal{A} \\ \emptyset & \text{otherwise} \end{array} \right.$$

Since $\mathcal{R}_v$ is HT-safe w.r.t. $\Gamma$ and $\mathsf{sig}(\mathcal{A}) \subseteq \Gamma$, by Proposition 3 a model $J$ of $\mathcal{R}_v$ exists such that $\triangle^J = \triangle^I$, $X^J = X^I$ for each symbol $X \in \Gamma$, and $X^J = \emptyset$ for each $X \in \mathsf{safe}(\mathcal{R}_v, \Gamma)$. We define the ABox $\mathcal{A}'$ as follows:

$$\begin{aligned}
\mathcal{A}' = \ &\{s \approx s\} \ \cup \\
&\{A(s) \mid s \in A^J \text{ and } A \in \mathsf{sig}(\mathcal{R}_v)\} \ \cup \\
&\{\neg A(s) \mid s \notin A^J \text{ and } A \in \mathsf{sig}(\mathcal{R}_v)\} \ \cup \\
&\{R(s, s) \mid \langle s, s\rangle \in R^J \text{ and } R \in \mathsf{sig}(\mathcal{R}_v)\} \ \cup \\
&\{\geq 1\, R.A(s) \mid s \in (\geq 1\, R.A)^J \text{ and } \{R, A\} \subseteq \mathsf{sig}(\mathcal{R}_v)\} \ \cup \\
&\{\geq 1\, R.\neg A(s) \mid s \in (\geq 1\, R.\neg A)^J \text{ and } \{R, A\} \subseteq \mathsf{sig}(\mathcal{R}_v)\}
\end{aligned}$$

We now show that $\mathcal{A}'$ is an $\mathcal{R}_v$-extension of $\mathcal{A}$ w.r.t. $\Gamma$. Since $J$ coincides with $I$ on the interpretation of all atomic concepts and roles in $\Gamma$, $\mathcal{A}'$ satisfies Properties 1 and 3 of Definition 14. We next show that no hypertableau derivation rule is applicable to $\mathcal{A}'$ and $\mathcal{R}_v$. The $\approx$- and the $\bot$-rule are clearly not applicable to $\mathcal{A}'$. Furthermore, the construction of $\mathcal{A}'$ ensures that $\geq 1\, R.C(s) \in \mathcal{A}'$ if and only if $\{R(s, s), C(s)\} \subseteq \mathcal{A}'$, so the $\geq$-rule is not applicable to $\mathcal{A}'$ either. Finally, assume that the *Hyp*-rule is applicable to an HT-rule $\rho \in \mathcal{R}_v$ and $\mathcal{A}'$ with a mapping $\sigma$. Since $\mathcal{A}'$ contains only the individual $s$, the mapping $\sigma$ maps all variables in $\rho$ to $s$. Since $J \models \mathcal{R}_v$, rule $\varrho$ contains a head atom $V_j$ such that $J \models \sigma(V_j)$. Note that if $V_j$ is of the form $\geq n\, R.C$, then $n = 1$ since $\triangle^J$ contains just one element. Thus, $\sigma(V_j)$ can be of the form $A(s)$, $R(s, s)$, $\geq 1\, R.C(s)$, or $s \approx s$, where $A \in \mathsf{sig}(\mathcal{R}_v)$, $R \in \mathsf{sig}(\mathcal{R}_v)$, and $\mathsf{sig}(C) \subseteq \mathsf{sig}(\mathcal{R}_v)$. But then, by the construction of $\mathcal{A}'$ we have $\sigma(V_j) \in \mathcal{A}'$, which contradicts the assumption that the *Hyp*-rule is applicable to $R_v$ and $\mathcal{A}'$. $\qquad\square$

We are now ready to prove the main claim of the section.

**Lemma 8** (Completeness). *Let $\langle \Gamma, \mathcal{R}_v \cup \mathcal{A}_v, \mathcal{T}_h \rangle$ be an input of the $\mathcal{ALCHIQ}$ $\Omega^{\mathtt{a}}$-algorithm. If a derivation for $\langle \Gamma, \mathcal{R}_v \cup \mathcal{A}_v, \mathcal{T}_h \rangle$ contains a leaf node labeled with a clash-free ABox, then $\mathcal{R}_v \cup \mathcal{A}_v \cup \mathcal{T}_h$ is satisfiable.*

*Proof.* Let $\mathcal{A}$ be an ABox obtained from a clash-free ABox labeling a leaf of a derivation for $\langle \Gamma, \mathcal{R}_v \cup \mathcal{A}_v, \mathcal{T}_h \rangle$ by removing all assertions involving an indirectly blocked individual. Since $\mathcal{R}_v \cup \mathcal{A}_v$ is acyclic w.r.t. $\Gamma$, ABox $\mathcal{A}$ is finite by Lemma 6. Furthermore, $\mathcal{A}$ is clearly an HT-ABox and no derivation rule is applicable to $\mathcal{R}_v$, $\mathcal{A}$, and $\Omega^{\mathtt{a}}_{\mathcal{T}_h, \Gamma}$. Finally, it is straightforward to see that a mapping $h$ from the individuals in $\mathcal{A}_v$ to the individuals in $\mathcal{A}$ exists such that $h(a) = a$ for each individual $a$ occurring in $\mathcal{A}$, $C(a) \in \mathcal{A}_v$ implies $C(h(a)) \in \mathcal{A}$, and $R(a, b) \in \mathcal{A}_v$ implies $R(h(a), h(b)) \in \mathcal{A}$. Hence, each model of $\mathcal{R}_v \cup \mathcal{A} \cup \mathcal{T}_h$ can be extended to a model of $\mathcal{R}_v \cup \mathcal{A}_v \cup \mathcal{T}_h$ by interpreting each individual $a$ not occurring in $\mathcal{A}_v$ in the same way as $h(a)$. Thus, we prove this lemma by showing that $\mathcal{R}_v \cup \mathcal{A} \cup \mathcal{T}_h$ is satisfiable.

Let $\mathcal{R}_h$ be the result of transforming $\mathcal{T}_h$ into a set of HT-rules as described by Motik et al. (2009); then, $\mathcal{R}_v \cup \mathcal{A}_v \cup \mathcal{T}_h$ is equisatisfiable with $\mathcal{R}_v \cup \mathcal{A}_v \cup \mathcal{R}_h$, and each model of the latter is a model of the former as well. Therefore, in the rest of the proof we extend $\mathcal{A}$ to a clash-free extended HT-ABox $\mathcal{A}_{\mathsf{fin}}$ such that no derivation rule from Table 2 is applicable to $\mathcal{R}_v \cup \mathcal{R}_h$ and $\mathcal{A}_{\mathsf{fin}}$. By Lemma 3, $\mathcal{R}_v \cup \mathcal{A}_{\mathsf{fin}} \cup \mathcal{R}_h$ is satisfiable, which, together with $\mathcal{A} \subseteq \mathcal{A}_{\mathsf{fin}}$, implies the satisfiability of $\mathcal{R}_v \cup \mathcal{A} \cup \mathcal{R}_h$. Before proceeding with the construction of $\mathcal{A}_{\mathsf{fin}}$, we next introduce several useful definitions and notational conventions.

- Let $\Gamma_v = \mathsf{sig}(\mathcal{R}_v) \cup \mathsf{sig}(\mathcal{A}_v)$ and let $\Gamma_h = \mathsf{sig}(\mathcal{R}_h)$.

- In this proof, term "blocking" refers to the version of blocking given in Definition 8; term "w-blocking" refers to the version of blocking in Definition 13; and term "s-blocking" refers to the standard blocking given in Definition 1 with the additional requirement that individuals $s$, $s'$, $t$, and $t'$ are all unnamed.

- For each blocked individual $s$, we pick an arbitrary but fixed individual $s'$ that blocks $s$, which we call the *blocker* of $s$.

- The *modified hypertableau algorithm* is the same as the standard hypertableau algorithm from Definition 1 with the difference that it uses s-blocking and that it can be applied to ABoxes that contain unnamed individuals; such individuals are then treated by the algorithm as if they were named. The modified hypertableau algorithm is clearly sound, complete, and terminating.

- The *projection* of an ABox $\mathcal{A}$ to a set of individuals $S$ is the ABox consisting of exactly those assertions from $\mathcal{A}$ that contain only individuals in $S$.

We now proceed with the construction of $\mathcal{A}_{\mathsf{fin}}$. To this end, we split $\mathcal{A}|_{\Gamma}$ into ABoxes $\mathcal{A}^{\mathsf{nm}}$ and $\mathcal{A}^t$ as follows; we use these ABoxes later to construct $\mathcal{A}_{\mathsf{fin}}$.

- The ABox $\mathcal{A}^{\mathsf{nm}}$ is the projection of $\mathcal{A}|_{\Gamma}$ to the set containing all named individuals in $\mathcal{A}$ and all unnamed individuals that are connected to a named individual in $\mathcal{A}|_{\Gamma}$.

    – For each nonblocked blocking-relevant individual $t$ in $\mathcal{A}$, the ABox $\mathcal{A}^t$ is the projection of $\mathcal{A}|_\Gamma$ to the set containing $t$ and all (unnamed) individuals connected to $t$ in $\mathcal{A}|_\Gamma$.

Let $\mathcal{A}_{\text{der}}^{\text{nm}}$ be the result of taking any clash-free ABox labeling a leaf of a derivation for $\mathcal{R}_h \cup \mathcal{A}^{\text{nm}}$ by the modified hypertableau algorithm and then removing all assertions containing an indirectly blocked individual; furthermore, for each nonblocked blocking-relevant individual $t$ in $\mathcal{A}$, let $\mathcal{A}_{\text{der}}^t$ be obtained from $\mathcal{A}^t$ in an analogous way. ABoxes $\mathcal{A}_{\text{der}}^{\text{nm}}$ and $\mathcal{A}_{\text{der}}^t$ exist because $\Omega_{\mathcal{T}_h,\Gamma}^{\mathfrak{a}}(\mathcal{A}') = \mathbf{t}$ for each connected component $\mathcal{A}'$ of $\mathcal{A}^{\text{nm}}$, $\Omega_{\mathcal{T}_h,\Gamma}^{\mathfrak{a}}(\mathcal{A}^t) = \mathbf{t}$ for each $t$, and the modified hypertableau algorithm is sound, complete, and terminating. Since the supply of unnamed individuals is unlimited, we assume without loss of generality that the $\geq$-rule always introduces individuals that are "globally fresh"—that is, that do not occur in any other ABox.

We next extend $\mathcal{A}_{\text{der}}^{\text{nm}}$ and each $\mathcal{A}_{\text{der}}^t$ with assertions necessary to satisfy $\mathcal{R}_v$. Let $\mathcal{A}'$ be $\mathcal{A}^{\text{nm}}$ (resp. some $\mathcal{A}^t$) and let $\mathcal{A}_{\text{der}}'$ be $\mathcal{A}_{\text{der}}^{\text{nm}}$ (resp. the corresponding $\mathcal{A}_{\text{der}}^t$). We say that an individual $u$ is *fresh* in $\mathcal{A}_{\text{der}}'$ if $u$ occurs in $\mathcal{A}_{\text{der}}'$ but not in $\mathcal{A}'$. For each fresh individual $u$ in $\mathcal{A}_{\text{der}}'$, we define $\mathcal{A}_{\text{der}}'[u]$ as an $\mathcal{R}_v$-extension of the projection of $\mathcal{A}_{\text{der}}'|_\Gamma$ to $\{u\}$; without loss of generality, we assume that $\mathcal{A}_{\text{der}}'[u_1] = \mathcal{A}_{\text{der}}'[u_2]$ for all $u_1$ and $u_2$ for which the projections of $\mathcal{A}_{\text{der}}'|_\Gamma$ to $\{u_1\}$ and $\{u_2\}$ are isomorphic (i.e., identical up to the renaming of individuals). Finally, let $\mathcal{A}_{\text{fin}}'$ be the union of $\mathcal{A}_{\text{der}}'$ and $\mathcal{A}_{\text{der}}'[u]$ for each $u$ that is fresh in $\mathcal{A}_{\text{der}}'$; thus, we obtain ABoxes $\mathcal{A}_{\text{fin}}^{\text{nm}}$ and $\mathcal{A}_{\text{fin}}^t$. By Condition 1 of Definition 14, the atomic assertions of $\mathcal{A}_{\text{der}}'|_{\text{sig}(\mathcal{R}_h)}$ coincide with the atomic assertions of $\mathcal{A}'|_{\text{sig}(\mathcal{R}_h)}$. Furthermore, since all individuals involved in s-blocking are required to be unnamed and all isomorphic individuals are extended in the same way, this construction does not affect s-blocking—that is, $u$ is s-blocked in $\mathcal{A}'$ if and only if $u$ is s-blocked in $\mathcal{A}_{\text{der}}'$.

We now define $\mathcal{A}_{\text{fin}}$ as the ABox obtained by

1. taking the union of $\mathcal{A}$, $\mathcal{A}_{\text{fin}}^{\text{nm}}$, and $\mathcal{A}_{\text{fin}}^t$ for each nonblocked blocking-relevant individual $t$ in $\mathcal{A}$, and

2. adding $A(s)$ for each blocked individual $s$ in $\mathcal{A}$ with blocker $s'$ such that $A(s') \in \mathcal{A}_{\text{fin}}^{s'}$ and $A \in \Gamma_h$.[3]

By Lemma 1, $\mathcal{A}_{\text{fin}}^{\text{nm}}$ and all $\mathcal{A}_{\text{fin}}^t$ are HT-ABoxes, and $\mathcal{A}_{\text{fin}}$ is clearly an extended HT-ABox. We next show that no hypertableau derivation rule is applicable to $\mathcal{R}_v \cup \mathcal{R}_h$ and $\mathcal{A}_{\text{fin}}$.

To this end, we first show that $\mathcal{A}_{\text{fin}}$ satisfies the following property (*): if $\alpha \in \mathcal{A}_{\text{fin}}$ is an atomic assertion or an assertion of the form $a \approx b$ such that $\text{sig}(\alpha) \subseteq \Gamma_v$ and all individuals mentioned in $\alpha$ occur in $\mathcal{A}$, then $\alpha \in \mathcal{A}$. In particular, note that the extension of $\mathcal{A}_{\text{der}}^{\text{nm}}$ and $\mathcal{A}_{\text{der}}^t$ to $\mathcal{A}_{\text{fin}}^{\text{nm}}$ and $\mathcal{A}_{\text{fin}}^t$, respectively, does not introduce an atomic assertion $\alpha$ that involves an individual from $\mathcal{A}$ and for which $\text{sig}(\alpha) \cap (\Gamma_v \setminus \Gamma) \neq \emptyset$; hence, the only possibility for $\alpha \in \mathcal{A}_{\text{fin}}$, $\alpha \notin \mathcal{A}$, and $\text{sig}(\alpha) \subseteq \Gamma_v$ is if $\alpha \in \mathcal{A}_{\text{der}}^{\text{nm}}$ or $\alpha \in \mathcal{A}_{\text{der}}^t$ for some $t$. We consider next the former case; the latter one is analogous. We prove (*) by induction on the application of the derivation rules in the construction of $\mathcal{A}_{\text{der}}^{\text{nm}}$. To this end, we show that each ABox $\mathcal{A}'$ in a derivation for $\mathcal{A}^{\text{nm}}$ and $\mathcal{R}_h$ satisfies the following properties:

---

3. Note that, since $s$ is blocked, it is blocking-relevant.

1. If $\alpha \in \mathcal{A}'$ is an atomic assertion or an assertion of the form $a \approx b$ such that $\mathsf{sig}(\alpha) \subseteq \Gamma_v$ and all individuals mentioned in $\alpha$ occur in $\mathcal{A}$, then $\alpha \in \mathcal{A}$ or $\neg\alpha \in \mathcal{A}$.

2. If $R(a,b) \in \mathcal{A}'$ such that $a$ and $b$ occur in $\mathcal{A}$ and $R \in \Gamma_h \setminus \Gamma$, then $S \in \Gamma$ exists such that $S(a,b) \in \mathcal{A}$ or $S(b,a) \in \mathcal{A}$.

3. If $a \approx b \in \mathcal{A}'$ such that $a$ occurs in $\mathcal{A}$, then $R \in \Gamma$ and an individual $c$ occurring in $\mathcal{A}$ exist such that $R(a,c) \in \mathcal{A}$ or $R(c,a) \in \mathcal{A}$.

The base case is trivial. We next consider ways in which an assertion in $\mathcal{A}'$ can be derived. An application of the $\bot$-rule or the $\geq$-rule clearly preserves (1)–(3). In an application of the $\approx$-rule, the modified hypertableau algorithm treats the individuals in $\mathcal{A}$ as named; furthermore, if $a \approx b \in \mathcal{A}'$ and $a$ and $b$ occur in $\mathcal{A}$, by (1) we have $a \approx b \in \mathcal{A}$, so $a = b$ since the $\approx$-rule is not applicable to $\mathcal{A}$; but then, it is straightforward to see that (1)–(3) remain preserved. Finally, the following types of assertions are relevant in an application of the *Hyp*-rule to an HT-rule $\varrho \in \mathcal{R}_h$:

- $A(a)$ with $a$ in $\mathcal{A}$ and $A \in \Gamma$. Since the $A$-cut rule is not applicable to $\mathcal{A}$, we have $A(a) \in \mathcal{A}$ or $\neg A(a) \in \mathcal{A}$, so (1) holds.

- $R(a,b)$ with $a$ and $b$ in $\mathcal{A}$. The body of $\varrho$ then contains an atom that is matched to an assertion $R'(a,b) \in \mathcal{A}'$ or $R'(b,a) \in \mathcal{A}'$ with $R' \in \Gamma_v$ that satisfies the induction assumption; thus, $S \in \Gamma$ exists such that $S(a,b) \in \mathcal{A}$ or $S(b,a) \in \mathcal{A}$, so (2) holds. Furthermore, if $R \in \Gamma$, then this assertion satisfies the preconditions of the $R$-cut and the $R^-$-cut rule; since these rules are not applicable to $\mathcal{A}$, we have $R(a,b) \in \mathcal{A}$ or $\neg R(a,b) \in \mathcal{A}$, so (1) holds.

- $a \approx b$ with $a$ in $\mathcal{A}$. The body of $\varrho$ then contains an atom that is matched to an assertion $R'(a,c) \in \mathcal{A}'$ or $R'(c,a) \in \mathcal{A}'$ with $R' \in \Gamma_v$ that satisfies the induction assumption; thus, $S \in \Gamma$ exists such that $S(a,c) \in \mathcal{A}$ or $S(c,a) \in \mathcal{A}$, so (3) holds. Furthermore, if $b$ is in $\mathcal{A}$, then the body of $\varrho$ also contains an atom that is matched to an assertion $R''(a,c) \in \mathcal{A}'$ or $R''(c,a) \in \mathcal{A}'$ that satisfies the induction assumption; thus, $S' \in \Gamma$ exists such that $S'(a,c) \in \mathcal{A}$ or $S'(c,a) \in \mathcal{A}$. The precondition of the $\approx$-cut rule is then satisfied and, since the rule is not applicable to $\mathcal{A}$, we have $a \approx b \in \mathcal{A}$ or $a \not\approx b \in \mathcal{A}$, so (1) holds.

This completes the proof of (1)–(3). Property (*) is a straightforward consequence of (1): a derivation of an assertion $\alpha$ such that $\mathsf{sig}(\alpha) \subseteq \Gamma_v$ and all individuals mentioned in $\alpha$ occur in $\mathcal{A}$ either makes no difference or it leads to a contradiction. A straightforward consequence of (*) is that (59) and (60) hold for all individuals $u$ and $v$ that occur in $\mathcal{A}$:

$$\mathcal{L}_{\mathcal{A}_{\mathsf{fin}}}(u) \cap \Gamma_v = \mathcal{L}_{\mathcal{A}}(u) \tag{59}$$

$$\mathcal{L}_{\mathcal{A}_{\mathsf{fin}}}(u,v) \cap \Gamma_v = \mathcal{L}_{\mathcal{A}}(u,v) \tag{60}$$

We now show that no derivation rule of the hypertableau algorithm with w-blocking is applicable to $\mathcal{R}_v \cup \mathcal{R}_h$ and $\mathcal{A}_{\mathsf{fin}}$. We do so by considering the possible derivation rules.

($\geq$-rule) Assume that the $\geq$-rule is applicable to an assertion $\geq n\, R.C(s) \in \mathcal{A}_{\mathsf{fin}}$, so $s$ is not w-blocked in $\mathcal{A}_{\mathsf{fin}}$. We show that then $s$ is not blocked in $\mathcal{A}$, or $s$ is not s-blocked in $\mathcal{A}_{\mathsf{fin}}^{\mathsf{nm}}$, or $s$ is not s-blocked in some $\mathcal{A}_{\mathsf{fin}}^t$. We have the following cases.

- $\geq n\, R.C(s) \in \mathcal{A}$. Assume that $s$ is blocked in $\mathcal{A}$ with blocker $t$, and let $s'$ and $t'$ be the predecessors of $s$ and $t$, respectively. By the definition of blocking, (61)–(65) hold:

$$\mathcal{L}_{\mathcal{A}}(s) = \mathcal{L}_{\mathcal{A}}(t) \tag{61}$$

$$\mathcal{L}_{\mathcal{A}}(s') = \mathcal{L}_{\mathcal{A}}(t') \tag{62}$$

$$\mathcal{L}_{\mathcal{A}}(s, s') = \mathcal{L}_{\mathcal{A}}(t, t') \tag{63}$$

$$\mathcal{L}_{\mathcal{A}}(s', s) = \mathcal{L}_{\mathcal{A}}(t', t) \tag{64}$$

$$\mathcal{L}_{\mathcal{A}}(s, s') \cup \mathcal{L}_{\mathcal{A}}(s', s) \subseteq \Gamma_v \setminus \Gamma \tag{65}$$

By (59) and (60), the following properties hold as well:

$$\mathcal{L}_{\mathcal{A}_{\mathsf{fin}}}(s) \cap \Gamma_v = \mathcal{L}_{\mathcal{A}_{\mathsf{fin}}}(t) \cap \Gamma_v \tag{66}$$

$$\mathcal{L}_{\mathcal{A}_{\mathsf{fin}}}(s') \cap \Gamma_v = \mathcal{L}_{\mathcal{A}_{\mathsf{fin}}}(t') \cap \Gamma_v \tag{67}$$

Furthermore, the second item in the construction of $\mathcal{A}_{\mathsf{fin}}$ ensures that $\mathcal{L}_{\mathcal{A}_{\mathsf{fin}}}(s)$ and $\mathcal{L}_{\mathcal{A}_{\mathsf{fin}}}(t)$ coincide on each concept $C \in \Gamma_h$, which ensures the following property:

$$\mathcal{L}_{\mathcal{A}_{\mathsf{fin}}}(s) = \mathcal{L}_{\mathcal{A}_{\mathsf{fin}}}(t) \tag{68}$$

By (65), $\mathcal{A}|_{\Gamma}$ does not contains an assertion involving individuals $s$ and $s'$, or individuals $t$ and $t'$. By the construction of $\mathcal{A}_{\mathsf{fin}}$, the following properties hold:

$$\mathcal{L}_{\mathcal{A}_{\mathsf{fin}}}(s, s') = \mathcal{L}_{\mathcal{A}_{\mathsf{fin}}}(t, t') \tag{69}$$

$$\mathcal{L}_{\mathcal{A}_{\mathsf{fin}}}(s', s) = \mathcal{L}_{\mathcal{A}_{\mathsf{fin}}}(t', t) \tag{70}$$

Consider now each rule $\varrho \in \mathcal{R}_v \cup \mathcal{R}_h$. If $\varrho \in \mathcal{R}_h$, then no role in the body of $\varrho$ occurs in $\mathcal{L}_{\mathcal{A}_{\mathsf{fin}}}(s, s') \cup \mathcal{L}_{\mathcal{A}_{\mathsf{fin}}}(s', s)$, so $\varrho$ satisfies the condition of weakened pairwise anywhere blocking. If $\varrho \in \mathcal{R}_v$, then $\varrho$ satisfies the condition of weakened pairwise anywhere blocking due to (67). Together with (68)–(70), this implies that $s$ is w-blocked by $t$, which is a contradiction. Consequently, $s$ is not blocked in $\mathcal{A}$.

- $\geq n\, R.C(s) \in \mathcal{A}_{\mathsf{fin}}^{\mathsf{nm}}$ and $\geq n\, R.C(s) \notin \mathcal{A}$. If $s$ occurs in $\mathcal{A}$ or if $s$ is a successor of an individual that occurs in $\mathcal{A}$, then $s$ is not s-blocked in $\mathcal{A}_{\mathsf{fin}}^{\mathsf{nm}}$ since the modified hypertableau algorithm treats the individuals occurring in $\mathcal{A}$ as named and such individuals cannot be s-blocked. Otherwise, by the construction of $\mathcal{A}_{\mathsf{fin}}$, $\mathcal{L}_{\mathcal{A}_{\mathsf{fin}}}(u) = \mathcal{L}_{\mathcal{A}_{\mathsf{fin}}^{\mathsf{nm}}}(u)$ and $\mathcal{L}_{\mathcal{A}_{\mathsf{fin}}}(u, v) = \mathcal{L}_{\mathcal{A}_{\mathsf{fin}}^{\mathsf{nm}}}(u, v)$ for all individuals $u$ and $v$ occurring in $\mathcal{A}_{\mathsf{fin}}^{\mathsf{nm}}$ but not in $\mathcal{A}$; again, $s$ is not s-blocked in $\mathcal{A}_{\mathsf{fin}}^{\mathsf{nm}}$.

- $\geq n\, R.C(s) \in \mathcal{A}_{\mathsf{fin}}^{t}$ for some $t$ and $\geq n\, R.C(s) \notin \mathcal{A}$. This case is completely analogous to the previous one.

Let $\mathcal{A}'$ be the ABox for which the above property holds; note that $\geq n\, R.C(s) \in \mathcal{A}'$. The $\geq$-rule is not applicable to $s$ in $\mathcal{A}'$, so $\mathcal{A}'$ contains individuals $u_1, \ldots, u_n$ such that

$$\{\mathsf{ar}(R, s, u_i), C(u_i) \mid 1 \leq i \leq n\} \cup \{u_i \not\approx u_j \mid 1 \leq i < j \leq n\} \subseteq \mathcal{A}'.$$

By the construction of $\mathcal{A}_{\mathsf{fin}}$ we have $\mathcal{A}' \subseteq \mathcal{A}_{\mathsf{fin}}$, which then contradicts the assumption that the $\geq$-rule is applicable to $s$ and $\mathcal{A}_{\mathsf{fin}}$.

($\perp$-rule, first variant) Property (59) holds for each individual $s$ occurring in $\mathcal{A}$, and (71) and (72) hold for each individual $s$ occurring in $\mathcal{A}_{\mathsf{fin}}^{\mathsf{nm}}$ and $\mathcal{A}_{\mathsf{fin}}^t$, respectively.

$$\mathcal{L}_{\mathcal{A}_{\mathsf{fin}}}(s) \cap \Gamma_h = \mathcal{L}_{\mathcal{A}_{\mathsf{fin}}^{\mathsf{nm}}}(s) \cap \Gamma_h \qquad (71)$$

$$\mathcal{L}_{\mathcal{A}_{\mathsf{fin}}}(s) \cap \Gamma_h = \mathcal{L}_{\mathcal{A}_{\mathsf{fin}}^t}(s) \cap \Gamma_h \qquad (72)$$

Thus, $\{A(s), \neg A(s)\} \subseteq \mathcal{A}_{\mathsf{fin}}$ implies $\{A(s), \neg A(s)\} \subseteq \mathcal{A}'$, where $\mathcal{A}'$ can be $\mathcal{A}$, or $\mathcal{A}_{\mathsf{fin}}^{\mathsf{nm}}$, or some $\mathcal{A}_{\mathsf{fin}}^t$. Since the first variant of the $\perp$-rule is not applicable to $\mathcal{A}'$, it is not applicable to $\mathcal{A}_{\mathsf{fin}}$ either.

($\perp$-rule, second variant) Property (60) holds for each pair of individuals $s$ and $t$ occurring in $\mathcal{A}$. Furthermore, $\mathcal{A}_{\mathsf{fin}}^{\mathsf{nm}}$ and $\mathcal{A}_{\mathsf{fin}}^t$ do not contain negative assertions other than those already present in $\mathcal{A}$. Since the second variant of the $\perp$-rule is not applicable to $\mathcal{A}$, $\mathcal{A}_{\mathsf{fin}}^{\mathsf{nm}}$, and all $\mathcal{A}_{\mathsf{fin}}^t$, it is not applicable to $\mathcal{A}_{\mathsf{fin}}$ either.

($\perp$-rule, third variant) Suppose that the $\perp$-rule is applicable to an assertion of the form $s \not\approx s \in \mathcal{A}_{\mathsf{fin}}$. By the construction of $\mathcal{A}_{\mathsf{fin}}$, then $s \not\approx s \in \mathcal{A}'$ for $\mathcal{A}'$ being $\mathcal{A}$, $\mathcal{A}_{\mathsf{fin}}^{\mathsf{nm}}$, or $\mathcal{A}_{\mathsf{fin}}^t$ for some $t$. But then, since the $\perp$-rule is not applicable to $\mathcal{A}'$, it is not applicable to $\mathcal{A}_{\mathsf{fin}}$ either.

($\approx$-rule) Assume now that the $\approx$-rule is applicable to $\mathcal{A}_{\mathsf{fin}}$. Then, an assertion $s \approx s'$ in $\mathcal{A}_{\mathsf{fin}}$ exists with $s \neq s'$. By the construction of $\mathcal{A}_{\mathsf{fin}}$, we have that $s \approx s' \in \mathcal{A}'$, with $\mathcal{A}' = \mathcal{A}$, or $\mathcal{A}' = \mathcal{A}_{\mathsf{fin}}^{\mathsf{nm}}$, or $\mathcal{A}' = \mathcal{A}_{\mathsf{fin}}^t$ for some $t$. But then, since the $\approx$-rule is not applicable to $\mathcal{A}'$, it is not applicable to $\mathcal{A}_{\mathsf{fin}}$ either.

(Hyp-rule) Assume that the Hyp-rule is applicable to $\mathcal{A}_{\mathsf{fin}}$ and an HT-rule $\varrho \in \mathcal{R}_v \cup \mathcal{R}_h$ of the form (2). Thus, a mapping $\sigma$ from the variables in $\varrho$ to the individuals $\mathcal{A}_{\mathsf{fin}}$ exists such that $\sigma(U_i) \in \mathcal{A}_{\mathsf{fin}}$ for each $1 \leq i \leq m$, but $\sigma(V_j) \notin \mathcal{A}_{\mathsf{fin}}$ for each $1 \leq j \leq n$. Let $s = \sigma(x)$ and $u_i = \sigma(y_i)$. We have the following possibilities:

- $\varrho \in \mathcal{R}_h$. Let $\mathcal{A}'$ be the ABox chosen among $\mathcal{A}_{\mathsf{fin}}^{\mathsf{nm}}$ and $\mathcal{A}_{\mathsf{fin}}^t$ containing the individual $s$. Consider now each $u_i$. Then $\varrho$ contains an atom of the form $R_{ij}(x, y_i)$ or $R_{ij}(y_i, x)$ with $R_{ij} \in \Gamma_h$, so $\mathcal{A}_{\mathsf{fin}}$ contains an assertion of the form $R_{ij}(s, u_i)$ or $R_{ij}(u_i, s)$. By the definition of blocking, for each pair of individuals $u$ and $v$ that belong to different $\mathcal{A}^{\mathsf{nm}}$ and $\mathcal{A}^t$, the ABox $\mathcal{A}$ does not contain an assertion of the form $T(u, v)$ with $T \in \Gamma_h$; but then, by the construction of $\mathcal{A}_{\mathsf{fin}}$, if $u$ and $v$ belong to different $\mathcal{A}_{\mathsf{fin}}^{\mathsf{nm}}$ and $\mathcal{A}_{\mathsf{fin}}^t$, the ABox $\mathcal{A}_{\mathsf{fin}}$ does not contain such an assertion either. Thus, all $u_i$ occur in $\mathcal{A}'$, so the Hyp-rule is applicable to $\varrho$ and $\mathcal{A}'$, which is a contradiction.

- $\varrho \in \mathcal{R}_v$. We first show the following property (**): if $s$ or some $u_i$ does not occur in $\mathcal{A}$, then $s = u_j$ for each $u_j$. We consider first the case when $s$ does not occur in $\mathcal{A}$. Consider an arbitrary $u_j$. Since $\varrho$ is an HT-rule, the body of $\varrho$ contains an atom of the form $R_{jk}(x, y_j)$ or $R_{jk}(y_j, x)$, so $\mathcal{A}_{\mathsf{fin}}$ contains an assertion of the form $R_{jk}(s, u_j)$ or $R_{jk}(u_j, s)$. We have the following two possibilities for $R_{jk}$.

  - $R_{jk} \in \Gamma_v \setminus \Gamma$. By the construction of $\mathcal{A}_{\mathsf{fin}}$, assertion $R_{jk}(s, u_j)$ or $R_{jk}(u_j, s)$ with $s$ not occurring in $\mathcal{A}$ must have been introduced via some $\mathcal{R}_v$-extension, so $u_j = s$.

  - $R_{jk} \in \Gamma$. Since $\varrho$ is HT-safe w.r.t. $\Gamma$, $\varrho$ contains an atom of the form $A(x)$ such that $A \in \mathsf{safe}(\mathcal{R}_v, \Gamma)$ in the body. By Condition 3 of Definition 14, $A(s) \notin \mathcal{A}_{\mathsf{fin}}$, which is a contradiction.

The case when some $u_i$ does not occur in $\mathcal{A}$ is symmetric; the only difference is that in case $R_{jk} \in \Gamma$ we consider a body atom $B(y_j)$ of $\varrho$ such that $B \in \mathsf{safe}(\mathcal{R}_v, \Gamma)$.

Let $\mathcal{A}' = \mathcal{A}$ if $s$ occurs in $\mathcal{A}$, and let $\mathcal{A}'$ be the ABox that contains $s$ otherwise. A straightforward consequence of (**) is that $\sigma(U_i) \in \mathcal{A}'$ for each $1 \leq i \leq m$; furthermore, $\mathcal{A}' \subseteq \mathcal{A}_{\mathsf{fin}}$ and $\sigma(V_j) \notin \mathcal{A}_{\mathsf{fin}}$ imply $\sigma(V_j) \notin \mathcal{A}'$ for each $1 \leq j \leq n$. But then, the *Hyp*-rule is applicable to $\mathcal{A}'$ for $\varrho$ and $\sigma$, which is a contradiction.

Thus, no derivation rule of the hypertableau algorithm with w-blocking is applicable to $\mathcal{R}_v \cup \mathcal{R}_h$ and $\mathcal{A}_{\mathsf{fin}}$, so $\mathcal{R}_v \cup \mathcal{R}_h \cup \mathcal{A}_{\mathsf{fin}}$ is satisfiable by Lemma 3. As explained earlier, this then proves the claim of this lemma. $\qquad\square$

Lemmas 6, 7, and 8 immediately imply Theorem 11.

## Appendix B. Proof of Theorem 12

The termination argument for the Horn-$\mathcal{ALCHIQ}$ $\Omega^{\mathsf{e}}$-algorithm is analogous to the non-Horn case: for each derivation for $\langle \Gamma, \mathcal{R}_v \cup \mathcal{A}_v, \mathcal{T}_h \rangle$, and each node $t$ in the derivation, we can find an embedding $\mu$ as in Lemma 4; the proof is a straightforward variant of the proof given for the non-Horn case. Termination then follows exactly as in the non-Horn case. Soundness is a consequence of the soundness of the standard hypertableau algorithm together with the following lemma.

**Lemma 9.** *Let $\mathcal{R}_v$ be a set of HT-rules, let $\mathcal{T}_h$ be a Horn-$\mathcal{ALCHIQ}$ TBox, and let $\mathcal{A}$ an ABox such that $\mathcal{R}_v \cup \mathcal{T}_h \cup \mathcal{A}$ is satisfiable. Furthermore, let $\mathcal{A}_1$ be the ABox obtained by applying a derivation rule from Table 5 to $\mathcal{R}_v$ and $\mathcal{A}$. Then, $\mathcal{R}_v \cup \mathcal{T}_h \cup \mathcal{A}_1$ is satisfiable.*

*Proof.* Let $I$ be a model of $\mathcal{R}_v \cup \mathcal{T}_h \cup \mathcal{A}$, and let us assume that a derivation rule from Table 5 derives $\mathcal{A}_1 = \mathcal{A} \cup \{\alpha\}$. By the preconditions of the $\Omega^{\mathsf{e}}$-concept, $\Omega^{\mathsf{e}}$-role, and $\Omega^{\mathsf{e}}$-$\approx$ rules, then $\Omega^{\mathsf{e}}_{\mathcal{T}_h, \Gamma}(\mathcal{A}', \alpha) = \mathsf{t}$ for some connected component $\mathcal{A}'$ of $\mathcal{A}|_\Gamma$, so $\mathcal{T}_h \cup \mathcal{A}' \models \alpha$. Since $\mathcal{A}' \subseteq \mathcal{A}|_\Gamma \subseteq \mathcal{A}$, we have that $I \models \mathcal{T}_h \cup \mathcal{A}'$, so $I \models \alpha$, which implies our claim. $\qquad\square$

We now show completeness of the algorithm. If a set of HT-rules $\mathcal{R}$ is Horn, then each derivation of the hypertableau algorithm contains exactly one leaf node, so we can identify a derivation with a sequence of ABoxes $\mathcal{A}_0, \ldots, \mathcal{A}_n$. The following proposition is a straightforward consequence of the fact that $\mathcal{R}$ is a Horn set of HT-rules.

**Proposition 6.** *Let $\mathcal{R}$ be a set of Horn HT-rules, let $\mathcal{A}$ an ABox, and let $\mathcal{A}_0, \ldots, \mathcal{A}_n$ be a derivation for $\mathcal{R}$ and $\mathcal{A}$. Then, for each assertion $\alpha$ that mentions only the individuals from $\mathcal{A}$ such that $\alpha \in \mathcal{A}_i$ for some $1 \leq i \leq n$, we have $\mathcal{R} \cup \mathcal{A} \models \alpha$.*

**Lemma 10** (Completeness). *Let $\langle \Gamma, \mathcal{R}_v \cup \mathcal{A}_v, \mathcal{T}_h \rangle$ be an input of the Horn-$\mathcal{ALCHIQ}$ $\Omega^{\mathsf{e}}$-algorithm. If a derivation for $\langle \Gamma, \mathcal{R}_v \cup \mathcal{A}_v, \mathcal{T}_h \rangle$ contains a leaf node labeled with a clash-free ABox, then $\mathcal{R}_v \cup \mathcal{A}_v \cup \mathcal{T}_h$ is satisfiable.*

*Proof.* The proof is analogous to the proof of Lemma 8: given an ABox $\mathcal{A}$ labeling a derivation leaf, we construct an ABox $\mathcal{A}_{\mathsf{fin}}$ such that no derivation rule of the hypertableau algorithm with w-blocking is applicable to $\mathcal{R}_v \cup \mathcal{R}_h$ and $\mathcal{A}_{\mathsf{fin}}$. The construction and the bulk of the proof are exactly the same as in Lemma 8, and we next prove only properties that are affected by the difference in the derivation rules.

The preconditions of the derivation rules in Table 5 clearly ensure that, whenever a derivation rule is applied to an HT-ABox, the result is also an HT-ABox; consequently, $\mathcal{A}_{\mathsf{fin}}$ is an extended HT-ABox.

We next show that property (*) holds despite the change in the derivation rules: if $\alpha \in \mathcal{A}_{\mathsf{fin}}$ is an atomic assertion or an assertion of the form $a \approx b$ such that $\mathsf{sig}(\alpha) \subseteq \Gamma_v$ and all individuals mentioned in $\alpha$ occur in $\mathcal{A}$, then $\alpha \in \mathcal{A}$. In particular, note that the construction of $\mathcal{A}_{\mathsf{fin}}$ does not introduce an atomic assertion $\alpha$ that involves an individual from $\mathcal{A}$ and for which $\mathsf{sig}(\alpha) \cap (\Gamma_v \setminus \Gamma) \neq \emptyset$. Assume now that $\mathsf{sig}(\alpha) \subseteq \Gamma$ and all individuals in $\alpha$ occur in $\mathcal{A}$. By Proposition 6 we have $\mathcal{R}_h \cup \mathcal{A} \models \alpha$. Furthermore, in the same say as in Lemma 8 one can show that the preconditions of the $\Omega^{\mathsf{e}}$-concept, $\Omega^{\mathsf{e}}$-role, or $\Omega^{\mathsf{e}}$-$\approx$ rule are satisfied in $\mathcal{A}$; since the relevant rule in not applicable to $\mathcal{A}$, we have $\alpha \in \mathcal{A}$, which proves our claim.

The rest of the proof is exactly the same as in Lemma 8. □

Theorem 12 follows immediately from Lemmas 9 and 10.

## Appendix C. Proof of Theorem 13

For each set of $\mathcal{EL}$-rules $\mathcal{R}$ and each ABox $\mathcal{A}$, each derivation of the $\mathcal{EL}$ hypertableau algorithm contains exactly one leaf node, so we identify a derivation with a sequence of ABoxes $\mathcal{A}_0, \mathcal{A}_1, \ldots, \mathcal{A}_n$. Since $\mathcal{A}_{j-1} \subseteq \mathcal{A}_j$ for each $1 \leq i \leq n$, the ABox labeling the derivation leaf is uniquely defined by $\mathcal{R}$ and $\mathcal{A}$. The following lemma captures the relevant properties of the standard $\mathcal{EL}$ hypertableau algorithm, and it can be proved by a slight variation of the proofs by Motik and Horrocks (2008) and Baader et al. (2005).

**Lemma 11.** *Let $\mathcal{R}$ be a set of $\mathcal{EL}$-rules, let $\mathcal{A}$ be an ABox containing only named individuals, and let $\mathcal{A}_f$ be the ABox labeling a leaf of a derivation for $\mathcal{R}$ and $\mathcal{A}$. Then the following properties hold for each pair of atomic concepts $A, B \in \mathsf{sig}(\mathcal{R})$ and each individual $s$ in $\mathcal{A}$:*

1. *$A(s) \in \mathcal{A}_f$ if and only if $\mathcal{R} \cup \mathcal{A} \models A(s)$.*

2. *$B(a_A) \in \mathcal{A}_f$ if and only if $\mathcal{R} \models A \sqsubseteq B$.*

3. *For each $\mathcal{A}' \subseteq \mathcal{A}$ and each $\mathcal{R}' \subseteq \mathcal{R}$, we have $\mathcal{A}'_f \subseteq \mathcal{A}_f$, where $\mathcal{A}'_f$ is the ABox labeling a leaf of a derivation for $\mathcal{R}'$ and $\mathcal{A}'$.*

Just like in the $\mathcal{EL}$ hypertableau algorithm, each derivation of the $\mathcal{EL}$ $\Omega^{\mathsf{e}}$-algorithm contains exactly one leaf node, and the ABox labeling the derivation leaf is uniquely defined by $\langle \Gamma, \mathcal{R}_v \cup \mathcal{A}_v, \mathcal{T}_h \rangle$. We next show several useful properties of this algorithm.

**Lemma 12.** *Let $\langle \Gamma, \mathcal{R}_v \cup \mathcal{A}_v, \mathcal{T}_h \rangle$ be an input of the $\mathcal{EL}$ $\Omega^{\mathsf{e}}$-algorithm and let $\mathcal{A}^{\mathsf{e}}$ be the ABox labeling a leaf of a derivation for $\langle \Gamma, \mathcal{R}_v \cup \mathcal{A}_v, \mathcal{T}_h \rangle$. Then the following holds.*

1. *Let $\mathcal{R}_h$ be the set of $\mathcal{EL}$-rules corresponding to $\mathcal{T}_h$ as described by Motik et al. (2009), and let $\mathcal{A}^{\mathcal{EL}}$ be the ABox labeling a leaf of a derivation of the standard $\mathcal{EL}$ hypertableau algorithm for $\mathcal{R}_v \cup \mathcal{R}_h$ and $\mathcal{A}_v$; then, $\mathcal{A}^{\mathsf{e}} \subseteq \mathcal{A}^{\mathcal{EL}}$.*

    *2. If $\perp \notin \mathcal{A}^{\mathbf{e}}$ and $B(a_A) \in \mathcal{A}^{\mathbf{e}}$ with $A \in \Gamma$, then $B \notin \mathsf{safe}(\mathcal{R}_v, \Gamma)$.*

*Proof.* (Claim 1) Let $\mathcal{A}_0, \ldots, \mathcal{A}_n$ be a derivation of the $\mathcal{EL}$ $\Omega^{\mathbf{e}}$-algorithm for $\mathcal{R}_v$, $\mathcal{A}_v$, and $\Omega^{\mathbf{e}}_{\mathcal{T}_h,\Gamma}$ such that $\mathcal{A}_0 = \mathcal{A}_v$ and $\mathcal{A}_n = \mathcal{A}^{\mathbf{e}}$. We prove the claim inductively by showing that $\mathcal{A}_j \subseteq \mathcal{A}^{\mathcal{EL}}$ for each $0 \leq j \leq n$. For the induction base, we clearly have $\mathcal{A}_0 \in \mathcal{A}^{\mathcal{EL}}$. Assume now that $\mathcal{A}_{j-1} \subseteq \mathcal{A}^{\mathcal{EL}}$ and let $\mathcal{A}_j$ be obtained from $\mathcal{A}_{j-1}$ by applying a derivation rule of the $\mathcal{EL}$ $\Omega^{\mathbf{e}}$-algorithm. If the *Hyp*-rule is applied to $\mathcal{A}_{j-1}$ and some $\varrho \in \mathcal{R}_v$, then $\varrho \in \mathcal{R}_v \cup \mathcal{R}_h$, $\mathcal{A}_{j-1} \subseteq \mathcal{A}^{\mathcal{EL}}$, and the fact that *Hyp*-rule is not applicable to $\mathcal{A}^{\mathcal{EL}}$ and $\varrho$ imply $\mathcal{A}_j \subseteq \mathcal{A}^{\mathcal{EL}}$. The argument is analogous for the $\exists$-rule. Finally, assume that the $\Omega^{\mathbf{e}}$-concept rule derives $A(s)$ with $A \in \Gamma \cup \{\perp\}$ from $\mathcal{A}_{j-1}$. By the preconditions of the $\Omega^{\mathbf{e}}$-concept rule, then $\Omega^{\mathbf{e}}_{\mathcal{T}_h,\Gamma}(\mathcal{A}', A(s)) = \mathsf{t}$ for some connected component $\mathcal{A}'$ of $\mathcal{A}_{j-1}|_\Gamma$. By Property 1 of Lemma 11 then $A(s) \in \mathcal{A}''$, where $\mathcal{A}''$ is the ABox labeling a leaf of a derivation of the standard $\mathcal{EL}$ hypertableau algorithm for $\mathcal{A}'$ and $\mathcal{R}_h$. Now $\mathcal{A}' \subseteq \mathcal{A}^{\mathcal{EL}}$, $\mathcal{R}_h \subseteq \mathcal{R}_h \cup \mathcal{R}_v$, and Property 3 of Lemma 11 imply $\mathcal{A}'' \subseteq \mathcal{A}^{\mathcal{EL}}$; consequently, $A(s) \in \mathcal{A}^{\mathcal{EL}}$ and $\mathcal{A}_j \subseteq \mathcal{A}^{\mathcal{EL}}$.

    (Claim 2) Consider an arbitrary individual of the form $a_A$ with $A \in \Gamma$ and an arbitrary assertion $B(a_A) \in \mathcal{A}^{\mathbf{e}}$. By Claim 1, $B(a_A) \in \mathcal{A}^{\mathcal{EL}}$, so by Property 2 of Lemma 11 we have $\mathcal{A}_v \cup \mathcal{R}_v \cup \mathcal{R}_h \models A \sqsubseteq B$. Since $\mathcal{R}_v \cup \mathcal{R}_h$ are $\mathcal{EL}$-rules, $\mathcal{A}_v$ does not affect subsumption inferences, so $\mathcal{R}_v \cup \mathcal{R}_h \models A \sqsubseteq B$. Since $\perp \notin \mathcal{A}^{\mathbf{e}}$, an interpretation $I$ exists such that $A^I \neq \emptyset$ and $I \models \mathcal{R}_h$. Assume now that $B \in \mathsf{safe}(\mathcal{R}_v, \Gamma)$. By Proposition 3 and the fact that $\mathcal{R}_v$ is $\mathcal{EL}$-safe, a model $J$ of $\mathcal{R}_v$ exists such that $X^J = X^I$ for each $X \in \mathsf{sig}(\mathcal{R}_h)$, and $B^J = \emptyset$. Thus, $J \models \mathcal{R}_v \cup \mathcal{R}_h$, which contradicts the fact that $\mathcal{R}_v \cup \mathcal{R}_h \models A \sqsubseteq B$. $\square$

    Soundness of the $\mathcal{EL}$ $\Omega^{\mathbf{e}}$-algorithm follows immediately from Property 1 of Lemma 12 and the fact that the standard $\mathcal{EL}$ hypertableau algorithm is sound. We next prove that the algorithm is complete.

**Lemma 13** (Completeness). *Let $\langle \Gamma, \mathcal{R}_v \cup \mathcal{A}_v, \mathcal{T}_h \rangle$ be an input of the $\mathcal{EL}$ $\Omega^{\mathbf{e}}$-algorithm and let $\mathcal{T}_h$ be an $\mathcal{EL}$ TBox, and let $\mathcal{A}^{\mathbf{e}}$ be the ABox labeling a leaf of a derivation for $\langle \Gamma, \mathcal{R}_v \cup \mathcal{A}_v, \mathcal{T}_h \rangle$. If $\perp \notin \mathcal{A}^{\mathbf{e}}$, then $\mathcal{R}_v \cup \mathcal{A}_v \cup \mathcal{T}_h$ is satisfiable.*

*Proof.* Let $\mathcal{R}_h$ be the result of transforming $\mathcal{T}_h$ into a set of $\mathcal{EL}$-rules as described by Motik et al. (2009); then, $\mathcal{R}_v \cup \mathcal{A}_v \cup \mathcal{T}_h$ is equisatisfiable with $\mathcal{R}_v \cup \mathcal{A}_v \cup \mathcal{R}_h$, and each model of the latter is a model of the former as well. Therefore, in the rest of the proof we extend $\mathcal{A}$ to a clash-free ABox $\mathcal{A}_{\mathsf{fin}}$ such that no derivation rule from Table 2 is applicable to $\mathcal{R}_v \cup \mathcal{R}_h$ and $\mathcal{A}_{\mathsf{fin}}$. By Lemma 3, then $\mathcal{R}_v \cup \mathcal{A}_{\mathsf{fin}} \cup \mathcal{R}_h$ is satisfiable, which, together with $\mathcal{A} \subseteq \mathcal{A}_{\mathsf{fin}}$, implies the satisfiability of $\mathcal{R}_v \cup \mathcal{A} \cup \mathcal{R}_h$. Let $\Gamma_v = \mathsf{sig}(\mathcal{R}_v) \cup \mathsf{sig}(\mathcal{A}_v)$ and $\Gamma_h = \mathsf{sig}(\mathcal{R}_h)$.

    We next present the construction of $\mathcal{A}_{\mathsf{fin}}$. The first step is to extend $\mathcal{A}^{\mathbf{e}}$ such that it satisfies $\mathcal{R}_h$, which we achieve by applying the $\mathcal{EL}$ hypertableau algorithm to $\mathcal{R}_h$ and $\mathcal{A}^{\mathbf{e}}$. We assume that the individuals in $\mathcal{A}^{\mathbf{e}}$ of the form $a_A$ are reused whenever $A \in \Gamma_v$. We call the individuals from $\mathcal{A}^{\mathbf{e}}$ *old* and the freshly introduced individuals *new*, and we call an individual $\Gamma$-*relevant* if it is of the form $a_A$ with $A \in \Gamma$.

    We next show that each ABox $\mathcal{A}_j$ in a derivation for $\mathcal{R}_h$ and $\mathcal{A}^{\mathbf{e}}$ satisfies the following properties (*):

1. $\alpha \in \mathcal{A}_j$ implies $\alpha \in \mathcal{A}^{\mathbf{e}}$ whenever $\alpha$ is of the form

        (a) $C(s)$ with $\mathsf{sig}(C) \subseteq \Gamma_v$ and $s$ an old individual, or

(b) $R(s, t)$ with $R \in \Gamma_v$ and $s$ and $t$ old individuals.

2. For each $C(s) \in \mathcal{A}_j$, the following properties hold:

   (a) $\mathsf{sig}(C) \subseteq \Gamma_v$ or $\mathsf{sig}(C) \subseteq \Gamma_h$, and

   (b) if $s$ is a new individual, then $\mathsf{sig}(C) \subseteq \Gamma_h$.

3. For each $R(s, t) \in \mathcal{A}_j$, the following properties hold:

   (a) if $t$ is a new individual, then $R \in \Gamma_h$, and

   (b) if $s$ is new and $t$ is old, then $t$ is $\Gamma$-relevant and $R \in \Gamma_h$.

The proof of (*) is by induction on the application of the derivation rules. For the induction base, we have $\mathcal{A}_0 = \mathcal{A}^{\mathsf{e}}$. Statements (1) and (2a) hold trivially, and (2b) and (3) are vacuously true since $\mathcal{A}^{\mathsf{e}}$ contains only old individuals. Assume now that (1)–(3) hold for $\mathcal{A}_{j-1}$ and consider an application of a derivation rule that derives $\mathcal{A}_j$.

Assume that the $\exists$-rule is applied to $\exists R.A(s) \in \mathcal{A}_{j-1}$, deriving $R(s, a_A)$ and $A(a_A)$. If $\{R, A\} \subseteq \Gamma_v$ and $s$ is old, then $\exists R.A(s) \in \mathcal{A}^{\mathsf{e}}$ by the induction assumption; since the $\exists$-rule is not applicable to $\mathcal{A}^{\mathsf{e}}$, then $\{R(s, a_A), A(a_A)\} \subseteq \mathcal{A}^{\mathsf{e}}$, so (1) holds. Furthermore, if $A \in \Gamma_v \setminus \Gamma$, then $s$ is old by (2b), and $R \in \Gamma_v$ by (2a); but then $\exists R.A(s) \in \mathcal{A}^{\mathsf{e}}$, so the $\exists$-rule cannot be applicable to $\exists R.A(s)$ in $\mathcal{A}_{j-1}$. Consequently, we have $\{R, A\} \subseteq \Gamma_h$, so $A(a_A)$ clearly satisfies (2), and $R(s, a_A)$ clearly satisfies (3a). Finally, $a_A$ can be old only if $A \in \Gamma$, so $R(s, a_A)$ clearly satisfies (3b).

Assume that the *Hyp*-rule is applied to an $\mathcal{EL}$-rule $\rho \in \mathcal{R}_h$ of the form (8), deriving $C(s)$. Then, individuals $t_1, \ldots, t_m$ in $\mathcal{A}_{j-1}$ exist such that $A_i(s) \in \mathcal{A}_{j-1}$ for each $1 \leq i \leq k$ and $\{R_i(s, t_i), B_{i,1}(t_i), \ldots, B_{i,m_i}(t_i)\} \subseteq \mathcal{A}_{j-1}$ for each $1 \leq i \leq m$. ABox $\mathcal{A}_j$ trivially satisfies (1b) and (3), and it satisfies (2) because $\rho \in \mathcal{R}_h$, so $\mathsf{sig}(C) \subseteq \Gamma_h$. To show (1a), assume that $s$ is an old individual and $\mathsf{sig}(C) \subseteq \Gamma_v$; since $\rho \in \mathcal{R}_h$, then $\mathsf{sig}(C) \subseteq \Gamma$. By Property 1 of Lemma 11, then $\mathcal{R}_h \cup \mathcal{A}^{\mathsf{e}} \models C(s)$. Since $\mathsf{sig}(C) \subseteq \Gamma$, we have $\mathcal{R}_h \cup \mathcal{A}^{\mathsf{e}}|_\Gamma \models C(s)$. Since the $\Omega^{\mathsf{e}}$-concept rule is not applicable to $\mathcal{A}^{\mathsf{e}}$, we have $C(s) \in \mathcal{A}^{\mathsf{e}}$, so $\mathcal{A}_j$ satisfies (1a).

This completes the proof of (*). Let $\mathcal{A}_{\mathsf{der}}$ be the ABox labeling a leaf of a derivation of the $\mathcal{EL}$ hypertableau algorithm for $\mathcal{R}_h$ and $\mathcal{A}^{\mathsf{e}}$. Such $\mathcal{A}_{\mathsf{der}}$ is clash-free since $\bot \notin \mathcal{A}^{\mathsf{e}}$ and the $\Omega^{\mathsf{e}}$-rule is not applicable to $\mathcal{A}^{\mathsf{e}}$; furthermore, $\mathcal{A}_{\mathsf{der}}$ satisfies (*).

We now extend $\mathcal{A}_{\mathsf{der}}$ such that the $\mathcal{EL}$-rules in $\mathcal{R}_v$ are satisfied when they are matched to new individuals. To this end, for each new individual $u$ in $\mathcal{A}_{\mathsf{der}}$, let $\mathcal{A}_{\mathsf{der}}[u]$ be an $\mathcal{R}_v$-extension w.r.t. $\Gamma$ of the projection of $\mathcal{A}_{\mathsf{der}}$ on $\{u\}$; such $\mathcal{A}_{\mathsf{der}}[u]$ exists by Proposition 5 and the fact that $\mathcal{R}_v$ is $\mathcal{EL}$-safe. Let $\mathcal{A}_{\mathsf{fin}}$ be the union of $\mathcal{A}_{\mathsf{der}}$ and all such $\mathcal{A}_{\mathsf{der}}[u]$. Since $\mathcal{A}_v \subseteq \mathcal{A}^{\mathsf{e}}$ and $\mathcal{A}^{\mathsf{e}} \subseteq \mathcal{A}_{\mathsf{fin}}$, we have $\mathcal{A}_v \subseteq \mathcal{A}_{\mathsf{fin}}$. Furthermore, since $\bot \notin \mathcal{A}_{\mathsf{der}}$ and $\bot \notin \mathcal{A}_{\mathsf{der}}[u]$ for each $u$ that is new in $\mathcal{A}_{\mathsf{der}}$, we have $\bot \notin \mathcal{A}_{\mathsf{fin}}$. Finally, by (*), Property 2 of Lemma 12, and the fact that each $\mathcal{A}_{\mathsf{der}}[u]$ contains only one individual and no safe concepts, $\mathcal{A}_{\mathsf{fin}}$ satisfies the following properties (**):

1. For each $B(s) \in \mathcal{A}_{\mathsf{fin}}$ such that $s$ is $\Gamma$-relevant or new, we have $B \notin \mathsf{safe}(\mathcal{R}_v, \Gamma)$.

2. For each $R(s, t) \in \mathcal{A}_{\mathsf{fin}}$, the following properties hold:

   (a) if $t$ is a new individual and $s \neq t$, then $R \in \Gamma_h$, and

(b) if $s$ is new and $t$ is old, then $t$ is $\Gamma$-relevant and $R \in \Gamma_h$.

To complete the proof of this lemma, we show that no derivation rule of the hypertableau algorithm is applicable to $\mathcal{A}_{\mathsf{fin}}$ and $\mathcal{R}_v \cup \mathcal{R}_h$.

($\exists$-rule) Consider an arbitrary $\exists R.C(s) \in \mathcal{A}_{\mathsf{fin}}$. If $\exists R.C(s) \in \mathcal{A}_{\mathsf{der}}$, since the $\exists$-rule is not applicable to $\mathcal{A}_{\mathsf{der}}$, we have $\{R(s,t), C(t)\} \subseteq \mathcal{A}_{\mathsf{der}} \subseteq \mathcal{A}_{\mathsf{fin}}$. If $\exists R.C(s) \in \mathcal{A}_{\mathsf{der}}[u]$ for some individual $u$ that is new in $\mathcal{A}_{\mathsf{der}}$, by Definition 14 the $\exists$-rule is not applicable to $\mathcal{A}_{\mathsf{der}}[u]$, so $\{R(s,t), C(t)\} \subseteq \mathcal{A}_{\mathsf{der}}[u] \subseteq \mathcal{A}_{\mathsf{fin}}$. Either way, the $\exists$-rule is not applicable to $\exists R.C(s)$ in $\mathcal{A}_{\mathsf{fin}}$.

($Hyp$-rule) Assume that the $Hyp$-rule is applicable to an $\mathcal{EL}$-rule $\rho \in \mathcal{R}_v \cup \mathcal{R}_h$ of the form (8), deriving $C(s)$. Then, individuals $t_1, \ldots, t_m$ in $\mathcal{A}_{\mathsf{fin}}$ exist such that $A_i(s) \in \mathcal{A}_{\mathsf{fin}}$ for each $1 \leq i \leq k$ and $\{R_i(s,t_i), B_{i,1}(t_i), \ldots, B_{i,m_i}(t_i)\} \subseteq \mathcal{A}_{\mathsf{fin}}$ for each $1 \leq i \leq m$. Then we have the following possibilities:

- $\varrho \in \mathcal{R}_h$. For each new individual $u$ and each assertion $\alpha \in \mathcal{A}_{\mathsf{der}}[u] \setminus \mathcal{A}_{\mathsf{der}}$, by Definition 14 either $\mathsf{sig}(\alpha) \in \Gamma_v \setminus \Gamma$ or $\alpha$ is of the form $\exists R.C$. Thus, $A_i(s) \in \mathcal{A}_{\mathsf{der}}$ for each $1 \leq i \leq k$ and $\{R_i(s,t_i), B_{i,1}(t_i), \ldots, B_{i,m_i}(t_i)\} \subseteq \mathcal{A}_{\mathsf{der}}$ for each $1 \leq i \leq m$, so the $Hyp$-rule is applicable to $\varrho$ and $\mathcal{A}_{\mathsf{der}}$, which is a contradiction.

- $\varrho \in \mathcal{R}_v$. We first show the following property (***): if $s$ or some $t_i$ is new, then $s = t_j$ for each $t_j$. We have the following cases.

  - Assume that $s$ is new and consider an arbitrary $1 \leq i \leq m$. Clearly, $R_i \in \Gamma_v$; furthermore, if $R_i \in \Gamma$, since $\varrho$ is $\mathcal{EL}$-safe, the body of $\varrho$ contains an atom that is matched to some $B_{ij}(t_i) \in \mathcal{A}_{\mathsf{fin}}$ such that $B_{ij} \in \mathsf{safe}(\mathcal{R}_v, \Gamma)$. Assume now that $t_i \neq s$. If $t_i$ is new, then $R_i \in \Gamma$ by Statement (2a) of (**); furthermore, if $t_i$ is old, then $R_i \in \Gamma$ and $t_i$ is $\Gamma$-relevant by Statement (2b) of (**); consequently, $R_i \in \Gamma$ and $t_i$ is either new or $\Gamma$-relevant. But then, by Statement (1) of (** and Property 3 of Definition 14, then $B_{ij}(t_i) \notin \mathcal{A}_{\mathsf{fin}}$, which is a contradiction. Hence, we conclude that $s = t_i$.

  - Assume that $t_i$ is new for some $1 \leq i \leq m$. If $t_i \neq s$, by Statement (2a) of (**) then $R_i \in \Gamma$. Since $\varrho$ is $\mathcal{EL}$-safe, the body of $\varrho$ then contains an atom that is matched to some $B_{ij}(t_i) \in \mathcal{A}_{\mathsf{fin}}$ such that $B_{ij} \in \mathsf{safe}(\mathcal{R}_v, \Gamma)$. Statement (1) of (**) then implies $B_{ij}(t_i) \notin \mathcal{A}_{\mathsf{fin}}$, which is a contradiction. Hence, we conclude that $s = t_i$; by the previous case then $s = t_j$ for each $1 \leq j \leq m$.

  Let $\mathcal{A}' = \mathcal{A}^{\mathsf{e}}$ if $s$ is old, and $\mathcal{A}' = \mathcal{A}_{\mathsf{der}}[s]$ otherwise. A straightforward consequence of (***) is that $A_i(s) \in \mathcal{A}'$ for each $1 \leq i \leq k$ and $\{R_i(s,t_i), B_{i,1}(t_i), \ldots, B_{i,m_i}(t_i)\} \subseteq \mathcal{A}'$ for each $1 \leq i \leq m$. The $Hyp$-rule is not applicable to $\varrho$ and $\mathcal{A}'$, so $C(s) \in \mathcal{A}'$. Since $\mathcal{A}' \subseteq \mathcal{A}_{\mathsf{fin}}$, we have $C(s) \in \mathcal{A}_{\mathsf{fin}}$, which contradicts the assumption that the $Hyp$-rule is applicable to $\varrho$ and $\mathcal{A}_{\mathsf{fin}}$.

This completes the proof of this lemma. □

Finally, we prove that the $\mathcal{EL}\ \Omega^{\mathsf{e}}$-algorithm is an optimal import-by-query algorithm.

**Theorem 13.** *The $\mathcal{EL}$ $\Omega^{\mathbf{e}}$-algorithm is an import-by-query algorithm based on ABox entailment oracles for the class of inputs $\mathcal{C}[\Gamma^{\mathcal{C}}, \mathcal{R}_v^{\mathcal{C}} \cup \mathcal{A}_v^{\mathcal{C}}, \mathcal{T}_h^{\mathcal{C}}]$ from Definition 11. The algorithm can be implemented such that it runs in* PTime *in $N$ with a polynomial number (in $N$) of calls to $\Omega_{\mathcal{T}_h,\Gamma}^{\mathbf{e}}$, where $N = |\mathcal{R}_v \cup \mathcal{A}_v| + |\Gamma|$ for the input $\mathcal{R}_v$, $\mathcal{A}_v$, and $\Gamma$.*

*Proof.* That the $\mathcal{EL}$ $\Omega^{\mathbf{e}}$-algorithm is an import-by-query algorithm is a straightforward consequence of Lemmas 12 and 13. To estimate the algorithm's running time, note that each application of a derivation rule adds an assertion of the form $C(a)$ or $R(a, b)$ for $C \in \Gamma_v \cup \{\bot\}$, where $a$ and $b$ are individuals occurring either in $\mathcal{A}_v$ or are of the form $a_A$ with $A \in \mathsf{sig}(\mathcal{R}_v)$. Clearly, the maximal number of individuals occurring in an ABox in a derivation is polynomial in the size of $\mathcal{A}_v$, $\mathcal{R}_v$, and $\Gamma$, and so is the maximal number of assertions. Furthermore, no derivation rule removes assertions from an ABox, so the number of assertions in an ABox monotonically increases in the course of a derivation. Consequently, the number of rule applications is polynomial in the size of $\mathcal{A}_v$, $\mathcal{R}_v$, and $\Gamma$. In the same way as in the standard $\mathcal{EL}$ hypertableau algorithm (Motik & Horrocks, 2008), one can show that each derivation rule can be applied in polynomial time, which implies the claim of this theorem. $\square$